\documentclass[10pt,twocolumn]{article}

\usepackage[margin=5.2em]{geometry}
\usepackage{times}
\usepackage{epsfig}
\usepackage{graphicx}
\usepackage{amsmath}
\usepackage{amssymb}
\usepackage{amsmath}
\usepackage{amssymb}
\usepackage{mathtools}
\usepackage{amsthm}
\usepackage{amsmath,amssymb,amsfonts}
\usepackage{graphicx}
\usepackage{textcomp}
\usepackage{xcolor}
\usepackage{times}
\usepackage{epsfig}
\usepackage{graphicx}
\usepackage{siunitx}
\usepackage{booktabs}
\usepackage{mathtools}
\usepackage{tabularx}
\usepackage{algorithm}
\usepackage{algorithmic}

\usepackage[pagebackref=true,breaklinks=true,colorlinks,bookmarks=false]{hyperref}

\newcommand{\mc}{\mathcal}
\newcommand{\lamcyc}{\lambda_\text{rec}}
\newcommand{\thetadepth}{{\theta_\mc{Y}}}
\newcommand{\thetargb}{{\theta_\mc{X}}}
\newcommand{\omegdepth}{{\omega_\mc{Y}}}
\newcommand{\omegrgb}{{\omega_\mc{X}}}

\newcommand{\wass}{\mathcal W}
\newcommand{\norm}[1]{\left\lVert#1\right\rVert}
\newcommand{\cspace}{[0,255]^{d_1\times d_2\times 3}}
\newcommand{\dspace}{\mathbb{R}^{d_1\times d_2\times 1}}
\newcommand{\fex}{\phi_\mc{X}}
\newcommand{\fey}{\phi_\mc{Y}}
\newcommand{\outdom}{\mc{Y}}

\usepackage{cleveref}
\usepackage{authblk}

\begin{document}
	
	\title{Unsupervised Single-shot Depth Estimation using Perceptual Reconstruction}
	
	\author[1]{Christoph Angermann}
	\author[1]{Matthias Schwab}
	\author[1]{Markus Haltmeier}
	\author[2]{Christian Laubichler}
	\author[3]{Steinbj\"orn J\'{o}nsson}
	
	\affil[1]{Department of Mathematics, University of Innsbruck, Technikerstraße 13, \hspace{10cm} 6020 Innsbruck, Austria
		\hspace{10cm}
		\url{applied-math.uibk.ac.at}}
	
	\affil[2]{LEC GmbH, Inffeldgasse 19, 8010 Graz, Austria
		\hspace{10cm} \url{www.lec.at}}
	
	\affil[3]{INNIO Jenbacher GmbH \& Co OG, Achenseestrasse 1-3, 6200 Jenbach, Austria\hspace{10cm}\url{www.innio.com/en}}

	\maketitle
	\thispagestyle{empty}
	
\begin{abstract}
	Real-time estimation of actual object depth is an essential module for various autonomous system tasks such as 3D reconstruction, scene understanding and condition assessment. During the last decade of machine learning, extensive deployment of deep learning methods to computer vision tasks has yielded approaches that succeed in achieving realistic depth synthesis out of a simple RGB modality. Most of these models are based on paired RGB-depth data and/or the availability of video sequences and stereo images. The lack of sequences, stereo data and RGB-depth pairs makes depth estimation a fully unsupervised single-image transfer problem that has barely been explored so far.
	This study builds on recent advances in the field of generative neural networks in order to establish fully unsupervised single-shot depth estimation. Two generators for RGB-to-depth and depth-to-RGB transfer are implemented and simultaneously optimized using the Wasserstein-1 distance, a novel perceptual reconstruction term and hand-crafted image filters. We comprehensively evaluate the models using industrial surface depth data as well as the Texas 3D Face Recognition Database, the CelebAMask-HQ database of human portraits and the SURREAL dataset that records body depth. For each evaluation dataset the proposed method shows a significant increase in depth accuracy compared to state-of-the-art single-image transfer methods.
\end{abstract}

\section{Introduction}

Real-time depth inference of a given object is an essential computer vision task which can be applied in various robotic tasks such as simultaneous localization and mapping \cite{nyu,kitti,zhao2020} as well as autonomous quality inspection in industrial applications \cite{angermann2021wear,laubichler21}. As the popularity of VR applications has continued to grow, instant depth estimation has also become an integral part of modeling complex 3D information out of single 2D images of human faces \cite{arslan2019,khan2020} or body parts \cite{vlasic2008,surreal,tang2019}.
Depth information about an object can be directly obtained from sensors for optical distance measurement. Time-of-Flight (ToF) cameras, LIDAR or stereo imaging systems are often used in practice and were also employed to generate paired RGB-depth data from some well-known depth databases \cite{vlasic2008,bosphorus,texas,kitti,nyu,ionescu2014,tang2019}. Since these sensors are typically costly and time-consuming devices that are also sensitive to external influences, their applicability to fast full-image depth generation on small on-site devices is limited. These limitations have motivated depth synthesis out of a simpler modality in terms of acquisition effort, namely an RGB image. This development has initiated a completely new field of research in computer vision. 

An important contribution was made by Eigen et al. \cite{eigen2014}, who proposed deep convolutional neural networks (DCNNs) for monocular depth synthesis of indoor and outdoor scenes. Basically, monocular single-image depth estimation out of RGB images can be seen as a modality transfer in which observed data of one modality is mapped to desired properties of another, potentially more complex, modality. 
Although DCNNs are a promising approach that succeed on such transfer tasks, they are commonly based on large amounts of training data, and generation and acquisition can be a demanding task. In the supervised setting in particular, DCNNs make use of paired training data during network parameter optimization, i.e., the network is provided with a single-view RGB and corresponding per-pixel depth \cite{eigen2014,arslan2019,tang2019,kwak2020}. Since large scale dense depth profiles are not abundant in many applications, supervised approaches are not feasible for these objects.
One possible way to remedy these shortcomings of supervised methods is to consider self-supervised approaches based on monocular video clips in which a supervisory depth counterpart is extracted from pose changes between adjacent frames.
These models can be trained on RGB sequences in a self-supervised manner, where a depth network and a pose estimation network are simultaneously optimized via sophisticated view-synthesis losses \cite{zhou2017, godard2019,zhao2020,jafarian2021}. Obviously, these methods require non-static scenes or a moving camera position (e.g., moving humans \cite{jafarian2021}, autonomous driving \cite{kitti}).



\begin{figure}[htb!]
	\centering
	\includegraphics[width=0.28\columnwidth]{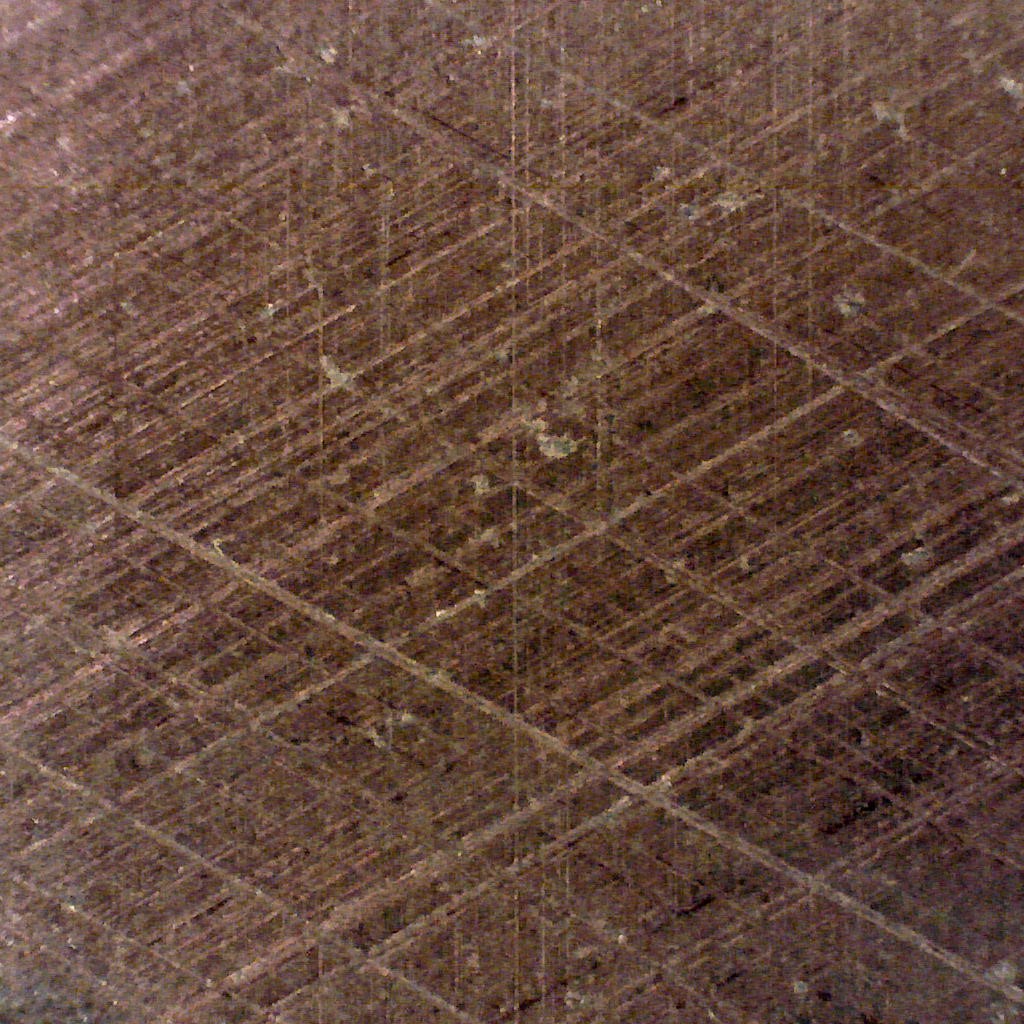}\hfill
	\includegraphics[width=0.28\columnwidth]{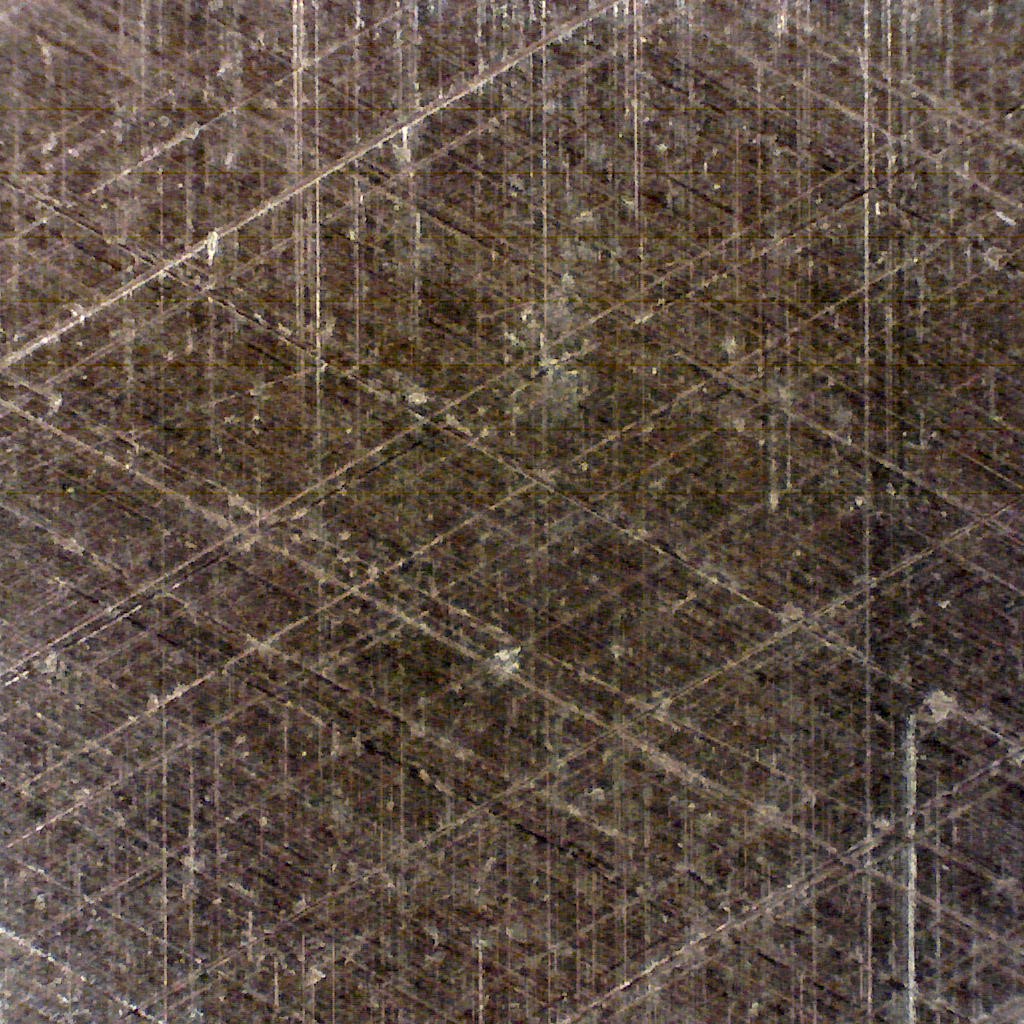}\hfill 
	\includegraphics[width=0.28\columnwidth]{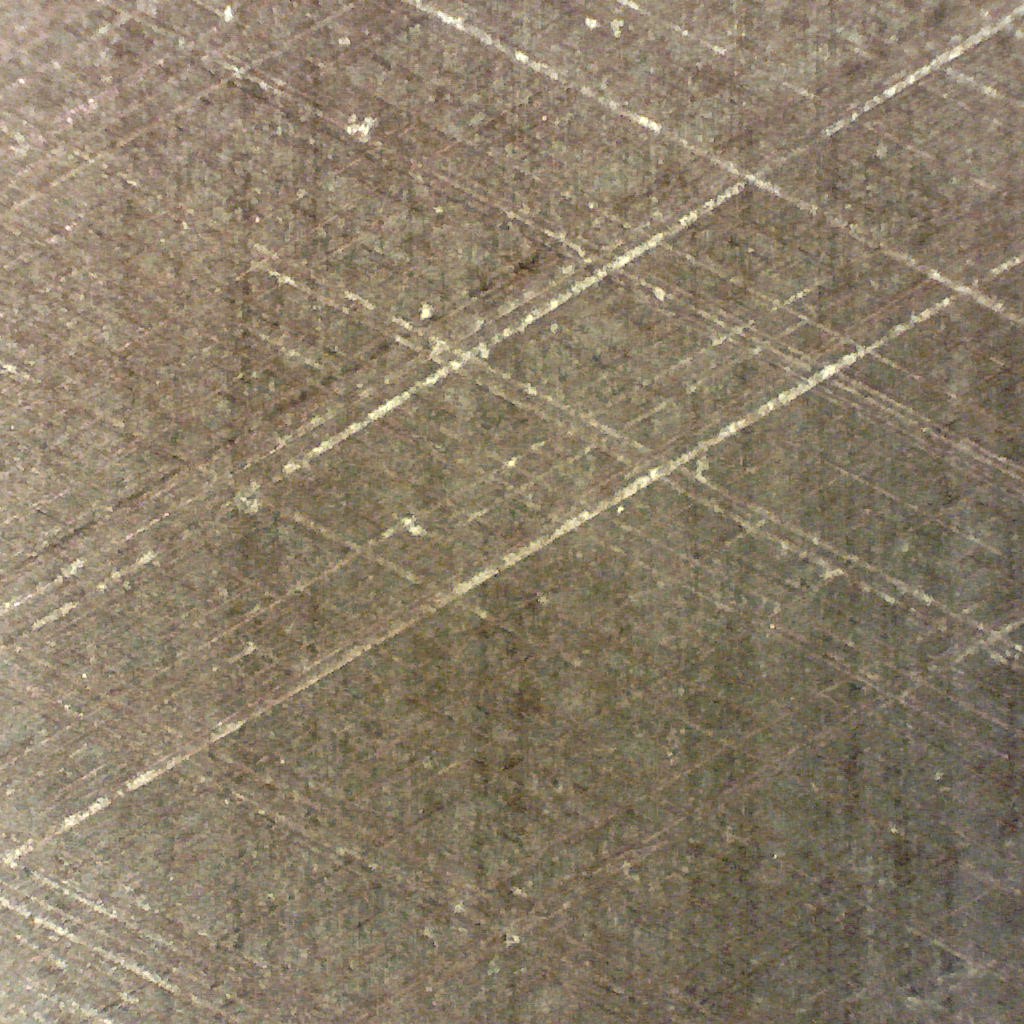}\\\vspace{.5em}
	\includegraphics[width=0.28\columnwidth]{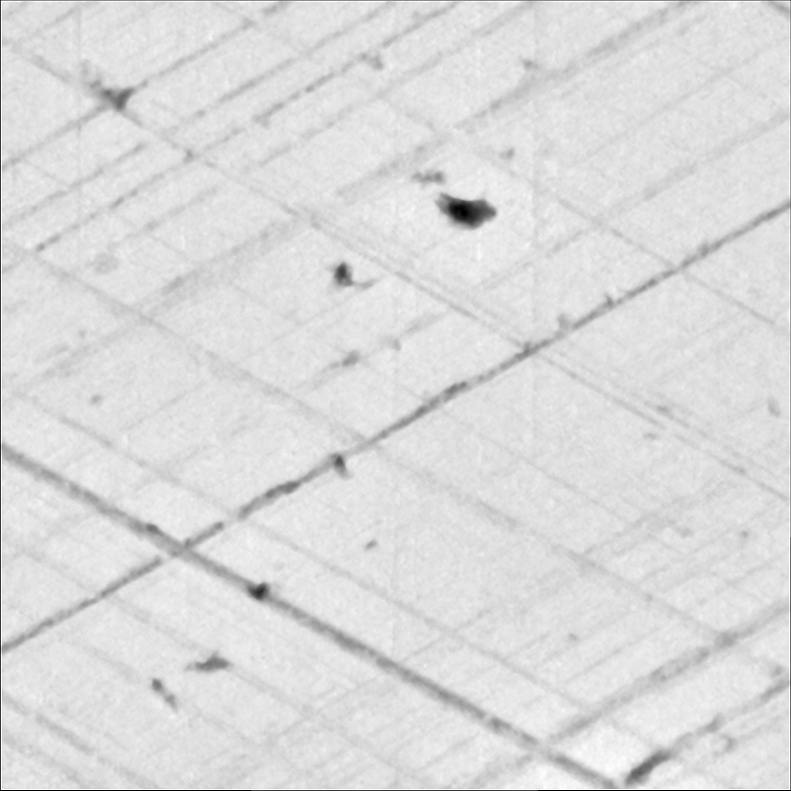}\hfill
	\includegraphics[width=0.28\columnwidth]{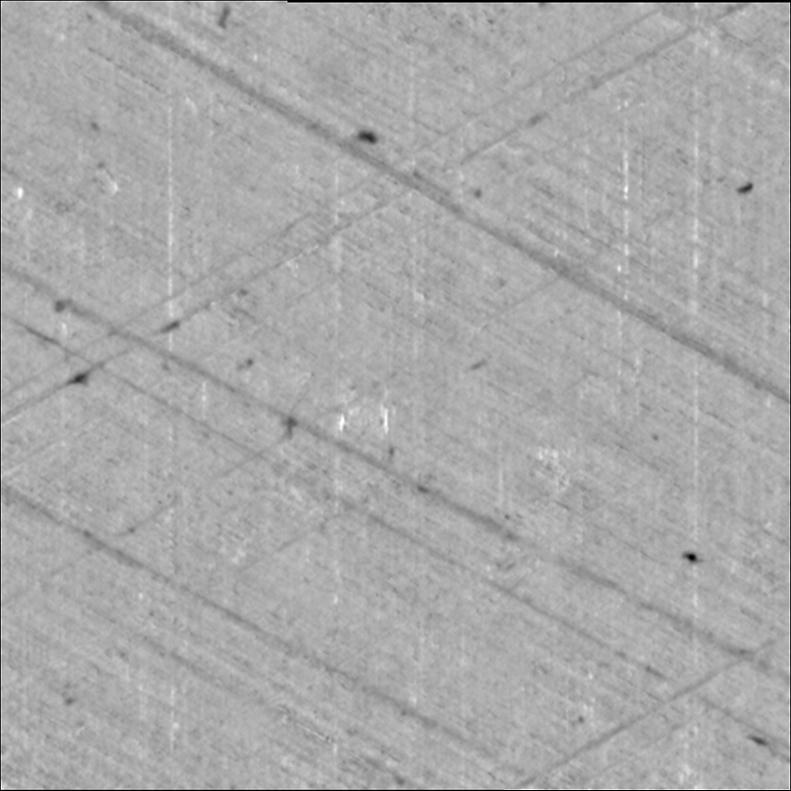}\hfill
	\includegraphics[width=0.28\columnwidth]{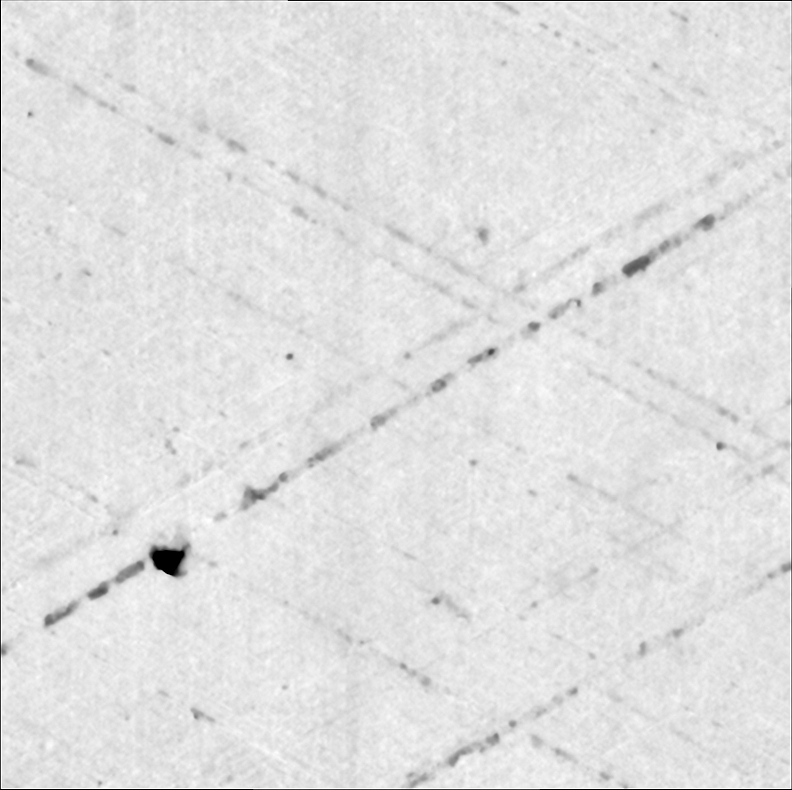}

	\caption{Top: RGB measurements of the inner surface of three cylinder liners with a spatial range of $\SI[product-units=power]{4.2x4.2}{\mm}$, recorded by a handheld microscope. Bottom: Depth profile of the same cylinder with a spatial range of $\SI[product-units=power]{1.9x1.9}{\mm}$, measured with a confocal microscope. The pixels of the modalities are not aligned.}
	\label{fig:measurements}
\end{figure}

{A very recent example for a scenario, where neither video sequences, stereo pairs nor paired data are available, is non-destructive evaluation of interal combustion engines for stationary power generation \cite{angermann2021wear,laubichler21}. Within this application, surface depth information has to be extracted from RGB image data.}
With current standards, cylinder condition can be assessed from a depth profile on a micrometer scale of the measured area (cf. Figure \ref{fig:measurements}). However, microscopic depth sensing of cylinder liner surface areas is a time-consuming and resource-intensive task which  consists of disassembling the liner, removing it from the engine, cutting it into segments and measuring them with a highly expensive and stationary confocal microscope \cite{angermann2021wear}. With a handheld microscope, however, single RGB records of the liner's inner surface can be generated from which depth profiles may be synthesized.
Since depth data is generated on a quite small scale ($\SI[product-units=power]{1.9x1.9}{\mm}$) and is comparatively high resolved, it is hardly possible to generate RGB data with accurately aligned pixel positions. This results in a fully unsupervised approach required for reasonable depth synthesis of this static scene.

%

{The main objective of this study is to propose a general method for depth estimation out of scenes for which neither paired data, video sequences, nor stereo pairs are available. Therefore, we consider the depth estimation problem as an intermodal transfer task of single images.}
Several recent advances in unpaired modality transfer are based on generative adversarial models (GAN) \cite{goodfellow2014}, cycle-consistency \cite{zhu2017} and probabilistic distance measures \cite{arjovsky2017,gulrajani2017}. The method proposed in this paper builds on established model architectures and training strategies in deep learning which are beneficially combined for unpaired single-view depth synthesis. Introduction of a novel perceptual reconstruction term in combination with appropriate hand-crafted filters further improves accuracy and depth contours.

The method is comprehensively tested on the afore mentioned industrial application of surface depth estimation. Furthermore, the approach is applied to other, external, datasets to create realistic scenarios where perfectly aligned RGB-depth data of single images is not available in practice. {More precisely, we test the model on the Texas 3D Face Recognition database (Texas-3DFRD) \cite{texas}, the Bosphorus-3DFA \cite{bosphorus} and the CelebAMask-HQ \cite{lee2020} to show its plausibility for facial data in an unsupervised setting.}
The SURREAL dataset \cite{surreal} is used to test performance on RGB-D videos of human bodies, where RGB and depth frames are not perfectly aligned.
{For every evaluation experiment the depth accuracy of the proposed framework is compared to state-of-the-art methods in unsupervised single-image transfer. To be more precise, the methods used for comparison are standard cycleGAN \cite{zhu2017}, CUT \cite{park2020} that uses contrastive learning for one-sided transfer and gcGAN \cite{huan2019} that utilizes geometric constraints between modalities. For facial data, we additionally compare to Wu et al. \cite{wu2020}, a very recent work where in addition to the depth profile also the albedo image, the illumination source and a symmetry confidence map is predicted in an unsupervised manner.}\\


\textbf{{Contributions:}}
\begin{itemize}
	
	\item This study finds a solution to the industrial problem of single-shot surface depth estimation where no paired data, no video sequences and no stereo pairs are available. 
	\item In this work depth estimation is considered as a single-image modality transfer; the proposed method shows superior performance over state-of-the-art works, quantitatively and qualitatively.
	\item Application to the completely different tasks of unsupervised face and human body depth synthesis indicates the universality of the approach.

\end{itemize}


\section{Related Work}

{The following section summarizes the most important milestones in the development of generative adversarial networks, highlights important work on single-image depth estimation as well as depth synthesis via GANs. In the supplementary, background is provided on some 3D databases that have been critical to the development of deep learning-based models for depth estimation.}

%


\subsection{Generative Adversarial Networks}

A standard GAN \cite{goodfellow2014} consists of a generator network $G\colon\mc Z\to\mc X$ mapping from a low-dimensional latent space $\mc Z$ to image space $\mc X$, where parameters of the generator are adapted so that the distribution of generated examples assimilates the distribution of a given data set. To be able to assess any similarity between arbitrary high-dimensional image distributions, a discriminator $f\colon\mc X\to [0,1]$ is trained simultaneously to distinguish between generator distribution and real data distribution. In a two-player min-max game, generator parameters are then updated to fool a steadily improving discriminator.
%
Usage of the initially proposed discriminator approach can cause the vanishing gradient problem  and  does not provide any information on the real distance between the generator and the real distribution. This issue
has been discussed thoroughly in \cite{arjovsky2017}, where the problem is bypassed by replacing the discriminator with a critic network that approximates the Wasserstein-1 distance \cite{villani2008} between the real distribution and the generator distribution. 

While the quintessence of GANs is to draw synthetic instances following a given data distribution, 
cycle-consistent GANs \cite{zhu2017} allow one-to-one mappings between two image domains $\mc X$ and $\mc Y$.
In essence, two generator networks $G_{\mc Y}\colon\mc X\to\mc Y,G_{\mc X}\colon\mc Y\to\mc X$ and corresponding discriminator networks $f_{\mc Y}\colon\mc Y\to[0,1],f_\mc X\colon \mc X \to [0,1]$ are trained simultaneously to enable generation of synthetic instances for both image domains (e.g., synthesizing winter landscapes from summer scenes and vice versa).
To ensure one-to-one correspondence, a cycle-consistency term is added to the two adversarial loss functionals. Although cycle-consistent GANs had initially been constructed for style transfer purposes, they were also very well received in the area of modality transfer in biomedical applications \cite{han2017,hiasa2018,lei2019}. 
Since optimization and fine-tuning of GANs often turns out to be extremely demanding and time-intensive, much research has emphasized stabilization of the training process through the development of stable network architectures such as DCGAN \cite{radford2015} or PatchGAN \cite{isola2017}. 

 \subsection{Monocular Depth Estimation}
Deep learning based methods achieve state-of-the-art results on depth synthesis task by training a DCNN on
a large-scale and extensive data set \cite{kitti,nyu}. Most of RGB-based models are supervised, i.e.  they require corresponding depth data that is pixel-wise aligned. One of the first DCNN approaches by Eigen et al. \cite{eigen2014} included sequential deployment of a coarse-scale stack and a refinement  module and was benchmarked on the KITTI \cite{kitti} and the NYU Depth v2 data set \cite{nyu}. Using a encoder-decoder structure in combination with an adversarial loss term helped to increase visual quality of the dense depth estimates \cite{jung2017}. Later methods also considered deep residual networks \cite{laina2016} or deep ordinal regression networks \cite{fu2018} in order to significantly increase performance on these data sets, where commonly considered performance measures are the root mean squared error (RMSE) or the $\delta_1$ accuracy \cite{zhao2020}.  Since a lot of research focused on further performance increase at the expense of model complexity and runtime, Wofk et al. \cite{wofk2019} used a lightweight network architecture \cite{howard2017} and achieved comparable results.

\subsection{Depth Estimation using GAN}
Use of left-right consistency and a GAN architecture results in excellent unsupervised depth estimation based on stereo images
\cite{pilzer2018,zhao2019}.
In \cite{kundu2018} and \cite{zheng2018}, a GAN has been trained to perform unpaired depth synthesis out of single monocular images. To this end, GANs were employed in the context of domain adaptation using an additional synthesized  data set of the same application with paired samples. This approach may not be regarded as a fully unsupervised method and requires availability or construction of a synthetic dataset. Arslan et Seke \cite{arslan2019} consider a conditional GAN (CGAN) \cite{isola2017} for solving single-image face depth synthesis. Nevertheless, CGANs rely on paired data since the adversarial part estimates the plausibility of an input-output pair. Another interesting approach was tried in \cite{kwak2020}, where indoor depth and segmentation were estimated simultaneously using cycle-consistent GANs.  The cycle-consistency loss helped them to maintain the characteristics of the RGB input during depth synthesis while the simultaneous segmentation resolved the fading problem in which depth information is hidden by larger features. However, the proposed discriminator network and reconstruction term in the generator loss function are based on paired RGB and depth/segmentation data, which is not available  for the aforementioned industrial application of surface depth synthesis.


\section{Method}
\begin{figure*}[htb!]
	\centering
	\includegraphics[width=0.75\textwidth]{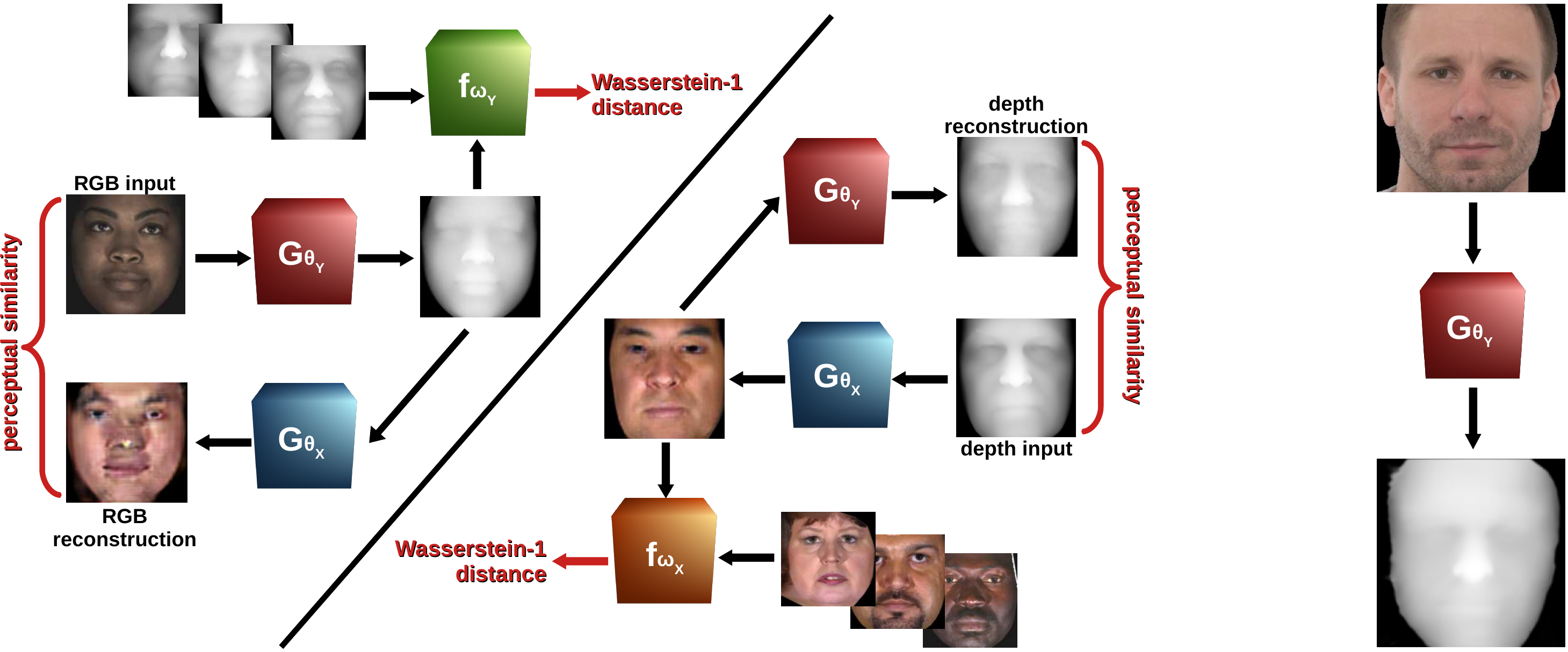}
	\caption{Illustration of the proposed framework: The left part describes the domains in which the RGB-to-depth generator $G_\thetadepth$ and the contrary depth-to-RGB generator $G_\thetargb$ operate. Both generators are updated via the probabilistic Wasserstein-1 distance, estimated by $f_\omegdepth$ in the input and $f_\omegrgb$ in the target domain. Perceptual similarity is compared between each generator input and its reconstruction. The right plot indicates that during inference, only $G_\thetadepth$ has to be deployed to synthesize new depth profiles. RGB images and ground truth depth images were taken from the Texas-3DFRD \cite{texas}.}
	\label{fig:cyclegan}
\end{figure*}
This section proposes an approach to monocular single-image depth synthesis with unpaired data and discusses the introduced framework and training strategy in detail.

\subsection{Setting and GAN Architecture}

The underlying structure of the proposed modality synthesis are two GANs linked with a reconstruction term (cf. Figure \ref{fig:cyclegan}). To be more exact, let $\mc  X \subset \cspace$ and $\mc Y \subset \dspace$ denote the domain of RGB and depth images, respectively, where the number of image pixels $d_1\cdot d_2$ is the same in both domains. Furthermore, let $X\coloneqq\{x_1,\ldots,x_M\}$ be the set of $M$ given RGB images and $Y\coloneqq\{y_1,\ldots,y_N\}$ the set of $N$ available but unaligned depth profiles. $P_\mc{X}$ and $P_\mc{Y}$ denote the distributions of the images in both domains.
The proposed model includes a generator function $G_{\thetadepth}\colon\mc X\to \mc Y$, which aims to map an input RGB image to a corresponding depth counterpart in the target domain. A generator function for image transfer may be approximated by a DCNN, which is parameterized by a weight vector $\thetadepth$ consisting of several convolution kernels. 
By adjusting $\thetadepth$, the distribution of generator outputs $P_{\thetadepth}$ may be brought closer to the real data distribution in the depth domain $P_\mc{Y}$. Note we do not know what $P_{\thetadepth}$ and $P_\mc{Y}$ actually look like, we only have access to unpaired training samples $G_{\thetadepth}(x)\sim P_{\thetadepth},\ x\in X$ and $y\sim P_\mc{Y},\ y\in Y$. An adversarial approach is deployed to ensure assimilation of both high-dimensional distributions in the GAN setting. The distance between the generator distribution and the real distribution is estimated by an additional DCNN $f_\omegdepth\colon\mc Y \to \mathbb{R}$, which is parameterized by weight vector $\omegdepth$ and is trained simultaneously with the generator network since $P_\thetadepth$ changes after each update to the generator weights $\thetadepth$. This ensures that $G_\thetadepth$ can be pitted against a steadily improving loss network $f_\omegdepth$ \cite{goodfellow2014}. 


This research work has chosen a network critic based on the Wasserstein-1 distance \cite{villani2008,arjovsky2017}.
The Wasserstein-1 distance (earth mover distance) between two distributions $P_1$ and $P_2$ is defined as
$ \wass_1(P_1, P_2) \coloneqq \inf_{J\in\mathcal J(P_1,P_2)}\mathbb{E}_{(x,y)\sim J}\norm{x-y}$,
where the infimum is taken over the set of all joint probability distributions that have marginal distributions $P_1$ and $P_2$. 
Since the exact computation of the infimum is highly intractable, the Kantorovich-Rubinstein duality \cite{villani2008} is used
\begin{align}\label{eq:wgan}
\wass_1(P_1,P_2)
=\sup_{\norm{f}_L\leq 1}\left[\underset{y\sim P_1}{\mathbb E}f(y)- \underset{y\sim P_{2}}{\mathbb{E}}f(y)\right],
\end{align}
where $\norm{\cdot}_L\leq C$ denotes that a function is $C$-Lipschitz. Equation (\ref{eq:wgan}) indicates that a good approximation to $\wass_1(P_\outdom,P_\thetadepth)$ is found by maximizing the distance ${\mathbb E}_{y\sim P_\outdom}f_\omegdepth(y)- {\mathbb{E}}_{y\sim P_\thetadepth}f_\omegdepth(y)$ over the set of DCNN weights $\{\omegdepth\mid f_\omegdepth\colon\outdom\to\mathbb{R}\ \text{1-Lipschitz}\}$, where the Lipschitz continuity of $f_\omegdepth$ can be enhanced via a gradient penalty \cite{gulrajani2017}. Given training batches $\mathbf{y}=\{y_n\}_{n=1}^b,\ y_n \overset{\mathrm{iid}}{\sim} P_\mc{Y}$ and $\mathbf{x}=\{x_n\}_{n=1}^b,\ x_n\overset{\mathrm{iid}}{\sim} P_\mc{X}$, this yields the following  empirical risk for critic $f_\omegdepth$:
\begin{equation}
\small
\label{eq:critic}
\begin{split}
\mc R_\text{cri}(\omegdepth,\thetadepth,p,\mathbf{y},\mathbf{x})\coloneqq &\frac{1}{b}\sum_{n=1}^{b}\bigg[ f_\omegdepth(G_\thetadepth(x_n))-f_\omegdepth(y_n)\\
+&p\cdot \left(\Big(\norm{\nabla_{\tilde y_n}f_\omegdepth(\tilde y_n)}_2-1\Big)_+ \right)^2\bigg],
\end{split}
\end{equation}
where $p$ denotes the influence of the gradient penalty, $( \cdot) _+\coloneqq \max(\{0,\cdot\})$  and $\tilde y_n \coloneqq \epsilon_n\cdot G_\thetadepth(x_n)+ (1-\epsilon_n)\cdot y_n$ for $\epsilon_n\overset{\mathrm{iid}}{\sim} \mc U[0,1]$. The goal of the RGB-to-depth generator $G_\thetadepth$ is to minimize the distance. Since only the first term of the functional in (\ref{eq:critic}) depends on the generator weights $\thetadepth$, the adversarial empirical risk for generator $G_\thetadepth$ simplifies as follows:
\begin{align}
\label{eq:adv}
\mc R_\text{adv}(\thetadepth,\omegdepth,\mathbf{x})\coloneqq -\frac{1}{b}\sum_{n=1}^{b}f_\omegdepth(G_\thetadepth(x_n)).
\end{align}






\subsection{Perceptual Reconstruction}
\label{sec:loss}
In the context of depth synthesis, it is not sufficient to ensure that the output samples lie in the depth domain. Care must be taken that synthetic depth profiles do not become irrelevant to the input. A reconstruction constraint forces generator input and output to share same spatial structure by taking into account the similarity between the input and the reconstruction of the synthesized depth profile. Obviously, calculation of a reconstruction error requires an opposite generator function  $G_\thetargb\colon\mc Y\to \mc X$ to assimilate real RGB distribution $P_\mc{X}$ as well as the corresponding distance network $f_\omegrgb\colon\mc X \to \mathbb{R}$. Both have to be optimized simultaneously to the RGB-to-depth direction. The reconstruction error is commonly evaluated by assessing similarity between $x$ and $G_\thetargb(G_\thetadepth(x))$ as well as similarity between $y$ and $G_\thetadepth(G_\thetargb(y))$ for $x\in \mc X$ and $y\in \mc Y$. In the setting of style transfer and cycle-consistent GANs \cite{zhu2017}, a pixelwise distance function on image space is considered, where the mean absolute error (MAE) or the mean squared error (MSE) are common choices.

The use of a contrary generator $G_\thetargb$ can be viewed as a type of regularization since it prevents mode collapse, i.e., generator outputs remain dependent on the inputs.
Deployment of the cycle-consistency approach \cite{zhu2017}, where reconstruction error is measured in image space, assumes no information loss during the modality transition. This corresponds to the applications of summer-to-winter landscape  or photograph-to-Monet painting transition. Determining $G_\thetadepth$ and $G_\thetargb$ is an ill-posed problem since a single depth profile may be generated by an infinite number of distinct RGB images and vice versa \cite{bhoi2019}. For example, during RGB-to-depth transition of human faces, information on image brightness, light source or the subject's skin color is lost. As a consequence, the contrary depth-to-RGB generator needed for regularization has to synthesize the lost properties of the image. Both generators $G_\thetadepth$ and $G_\thetargb$ may be penalized if the skin color or the brightness of the reconstruction is changed even though $G_\thetargb$ did exactly what we expected it to do, i.e., synthesize a face that is related to the input's depth profile.

Adapting the idea of \cite{dosovitskiy2016}, we propose a perceptual reconstruction loss, i.e., instead of computing a reconstruction error in image space, we consider certain image features of the reconstruction. Typical perceptual similarity metrics extract features by propagating the images (to be compared) through an auxiliary network that is usually pretrained on a large image classification task \cite{deng2009,dosovitskiy2016,heusel2017}. Nevertheless, we expect our feature extractor to be perfectly tailored to our data and not determined by an additional network pretrained on a very general classification task \cite{deng2009} that may not even cover our type of data. Therefore, we enforce the reconstruction consistency on the image space by using the MAE loss on feature vectors extracted by $\fex(\cdot)\coloneqq f_\omegrgb^l(\cdot)$, which corresponds to the $l$-th layer of the RGB critic (cf. Algorithm \ref{alg1}). Analogously, we define the feature extractor on depth space by $\fey(\cdot)\coloneqq f_\omegdepth^l(\cdot)$, which corresponds to the $l$-th layer of the depth critic. Although we are aware that feature extractor weights are adjusted with each update of critic weights $\omegrgb,\omegdepth$, we assume that, at least at a later stage of training, $\fex$ and $\fey$ have learned good and stable features on the image and depth domain. This yields the following empirical reconstruction risk:
\begin{align}
\begin{split}
\mc R_\text{rec} &(\thetargb,\thetadepth,\fex,\fey,\mathbf{x},\mathbf{y})\coloneqq\\
&\frac{1}{b}\sum_{n=1}^b\text{MAE}\big[\fex\big(G_\thetargb\left(G_\thetadepth(x_n)\right)\big),\fex(x_n) \big]\\
+\
&\frac{1}{b}\sum_{n=1}^b\text{MAE}\big[\fey\big(G_\thetadepth\left(G_\thetargb(y_n)\right)\big),\fey(y_n) \big].
\end{split}
\end{align}
In our implementation, we set $l\coloneqq L-2$  for a critic with $L$ layers, i.e., we use the second-to-last layer of the critic.\\

\begin{figure}[htb!]
	\centering 
	\includegraphics[width=.92\columnwidth]{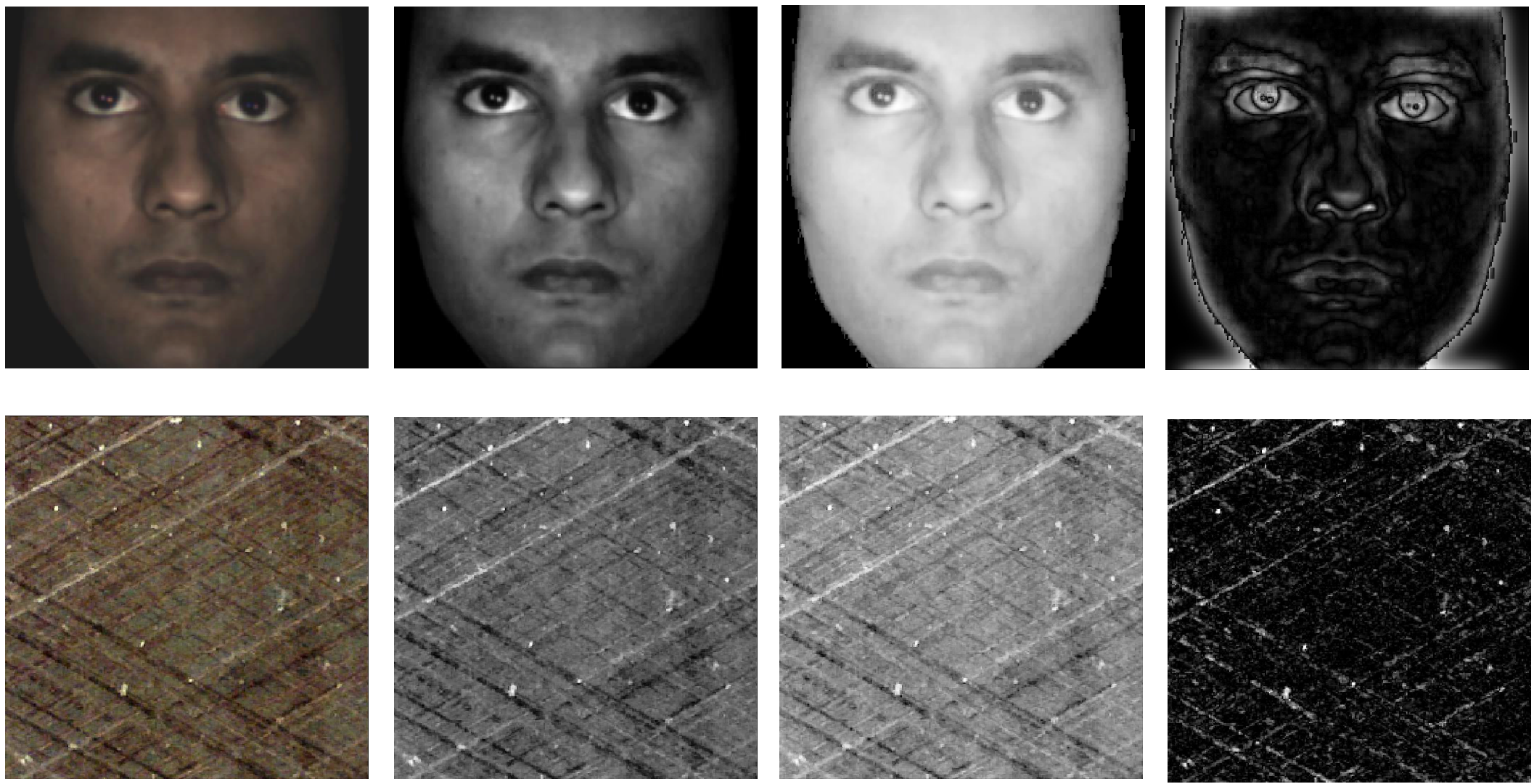}
	
	\caption{The first column visualizes the RGB samples and the second column the grayscale versions. The third column contains the gamma corrected counterparts, where the contrast in lower gray levels is enhanced for dark images in particular. The last column illustrates the application of the high-pass filter.}
	\label{fig:hpf}
\end{figure}

A good reconstruction term must still be found for the start of training when the critic features are not yet sufficiently reliable. At first, it is desirable to guide the framework to preserve structural similarity during RGB-to-depth and depth-to-RGB transition. Therefore, we propose to compare the input and its reconstruction in the image space while automatically removing the brightness, illumination and color of the RGB images beforehand. This can be ensured by applying the following steps:

\vspace{2em}
\begin{enumerate}
	\item Convert the image to grayscale by applying the function $g\colon\cspace\to \mathbb R^{d_1\times d_2},\quad x\mapsto \frac{0.299}{255}\cdot x_{(,,0)} +\frac{0.587}{255}\cdot x_{(,,1)}+\frac{0.144}{255}\cdot x_{(,,2)}$, where $({,,i})$ denotes the $i$-th color channel for $i=0,1,2$.
	\item Enhance the brightness of the grayscale image using an automated gamma correction based on the image brightness \cite{babakhani2015}, i.e. take the grayscale image $x_\text{gr}$ to the power of $\Gamma(x_\text{gr})\coloneqq {-0.3\cdot 2.303}/{\ln \overline{x_\text{gr}}}$, where $ \overline{x_\text{gr}}$ denotes the average of the gray values.
	\item Convolve the enhanced image with a high-pass filter $h$ in order to dim the lighting source and color information (cf. Figure \ref{fig:hpf}). The high-pass filter may be applied in Fourier domain, i.e., the 2D Fourier transform is multiplied by a Gaussian high-pass filter matrix $H^\sigma$ defined by $H^\sigma_{i,j}\coloneqq 1-\exp\big(\norm{(i,j)-(\frac{d_1}{2},\frac{d_2}{2})}_2^2 / (2\sigma^2)\big)$ for $i=1,\ldots,d_1$ and $ j=1,\ldots,d_2$. In our implementation, $\sigma=4$ yielded  satisfactory results for all tasks.
\end{enumerate}

This yields the updated empirical reconstruction risk:
\begin{align}
\small
\label{eq:rec}
\begin{split}
\mc R_\text{rec} (\thetargb,&\thetadepth,\fex,\fey,\gamma,\mathbf{x},\mathbf{y})\coloneqq\\
\gamma\cdot&\frac{1}{b}\sum_{n=1}^b\text{MAE}\big[\fex\big(G_\thetargb\left(G_\thetadepth(x_n)\right)\big),\fex(x_n) \big]\\
+\hspace{2.4em} 
\gamma\cdot&\frac{1}{b}\sum_{n=1}^b\text{MAE}\big[\fey\big(G_\thetadepth\left(G_\thetargb(y_n)\right)\big),\fey(y_n) \big]\\
+
(1-\gamma)\cdot&\frac{1}{b}\sum_{n=1}^b\text{MAE}\big[\psi\big(G_\thetadepth\left(G_\thetargb(x_n)\right)\big),\psi(x_n) \big]\\
+
(1-\gamma)\cdot&\frac{1}{b}\sum_{n=1}^b\text{MAE}\big[G_\thetadepth\left(G_\thetargb(y_n)\right),y_n \big],
\end{split}
\end{align}
where $\psi(\cdot)\coloneqq h*g(\cdot)^{\Gamma(g(\cdot))}$ and $\gamma$ is gradually increased from 0 to 1 during training to control feature extractor reliability. In the far right column in Figure \ref{fig:hpf}, we may observe the strong effect of operator $\psi$. For the face sample, the face shape and the positions of the nose and the eyes are very clear, at the same time the low image brightness and the exposure direction are resolved. The main edges of the cylinder liner surfaces are clearly identifiable whereas the different brown levels and illumination inconsistencies of the input are no longer visible. 

Using the previously discussed risk functions $\mc R_\text{cri}$ (\ref{eq:critic}), $\mc R_\text{adv}$ (\ref{eq:adv}) and $\mc R_\text{rec}$ (\ref{eq:rec}), Algorithm \ref{alg1} summarizes the proposed architecture for fully unsupervised single-view depth estimation.
Implementation of the proposed framework is publicly available on \url{https://github.com/anger-man/unsupervised-depth-estimation}.\\

\begin{algorithm}[htb]
	\scriptsize
	\caption{Proposed Framework}
	\label{alg1}
	\begin{algorithmic}
		\REQUIRE $\alpha_f$ critic learning rate; $\alpha_G$ generator learning rate; $p$ gradient penalty; $n_f$ number of critic iterations; $n_G$ number of generator updates; $b$ minibatch size; $\lamcyc$ reconstruction loss weight
		\REQUIRE {$\omegdepth,\omegrgb$ initial critic weights; $\thetadepth,\thetargb$ initial generator weights; $\gamma =0$}
		
		\FOR{$k = 1,\ldots,n_G$}
		\FOR{$i = 1,\ldots,n_f$}
		\STATE{Sample $\mathbf{x}=\{x_n\}_{n=1}^b\subset X$ and $\mathbf{y}=\{y_n\}_{n=1}^b\subset Y$}
		
		\STATE{$\{\tilde y_n\}_{n=1}^b\leftarrow  \left\{\epsilon_n\cdot G_{\thetadepth}(x_n)+(1-\epsilon_n)\cdot y_n,\ \epsilon_n\sim \mc U [0,1]\right\}_{n=1}^b$}
		\STATE{$\{\tilde x_n\}_{n=1}^b\leftarrow  \left\{\epsilon_n\cdot G_{\thetargb}(y_n)+(1-\epsilon_n)\cdot x_n,\ \epsilon_n\sim \mc U [0,1]\right\}_{n=1}^b$}
		
		\STATE{$\partial_{\mc Y}\leftarrow \nabla_{\omegdepth}\mc R_\text{cri}(\omegdepth,\thetadepth,p,\mathbf{y},\mathbf{x})$}
		
		\STATE{$\partial_{\mc X}\leftarrow \nabla_{\omegrgb}\mc R_\text{cri}(\omegrgb,\thetargb,p,\mathbf{x},\mathbf{y})$}
		
		
		%
		%

		\STATE{$\omegdepth\leftarrow\text{Adam}(\omegdepth,\partial_\mc{Y}, \alpha_f,\beta_1=0, \beta_2 = 0.9 )$}
		\STATE{$\omegrgb\leftarrow\text{Adam}(\omegrgb,\partial_\mc{X}, \alpha_f,\beta_1=0, \beta_2 = 0.9 )$}
		\ENDFOR
		\STATE{Sample $\mathbf{x}=\{x_n\}_{n=1}^b\subset X$ and $\mathbf{y}=\{y_n\}_{n=1}^b\subset Y$; set $\fey,\fex$ to $l$-th layer of $f_\omegdepth,f_\omegrgb$}
		
		\STATE{$
			\partial_\mc{Y} \leftarrow\qquad\  \nabla_{\thetadepth}\mc R_\text{adv}(\thetadepth,\omegdepth,\mathbf{x})+$}
		\STATE{$\qquad\quad	\lamcyc \cdot \nabla_{\thetadepth}\mc R_\text{rec} (\thetargb,\thetadepth,\fex,\fey,\gamma,\mathbf{x},\mathbf{y})$}
		\STATE{$
			\partial_\mc{X} \leftarrow\qquad\  \nabla_{\thetargb}\mc R_\text{adv}(\thetargb,\omegrgb,\mathbf{y})+$}
		\STATE{$\qquad \quad 	\lamcyc \cdot \nabla_{\thetargb}\mc R_\text{rec} (\thetargb,\thetadepth,\fex,\fey,\gamma,\mathbf{x},\mathbf{y})$}
		
		\STATE{$\thetadepth\leftarrow\text{Adam}(\thetadepth,\partial_D,\alpha_G,\beta_1=0,\beta_2=0.9)$}
		\STATE{$\thetargb\leftarrow\text{Adam}(\thetargb,\partial_C,\alpha_G,\beta_1=0,\beta_2=0.9)$}
		\STATE{$\gamma \leftarrow \frac{k}{n_G}$}
		
		\ENDFOR
		
	\end{algorithmic}
\end{algorithm}

{As critical as the loss function design of an unsupervised method is the choice of an appropriate architecture for the critic and the generator network. A decoder for the critic is built following the PacthGAN critic that was initially proposed in \cite{isola2017} with nearly \num{15.7e6} parameters. The PatchGAN architecture is empirically proven to perform quite stably over a variety of different generative task and is part of many state-of-the-art architectures for image generation \cite{zhu2017,park2020,fu2019}. The generator is a ResNet18 \cite{he2015} with a depth-specific upsampling part taken from \cite{godard2019} (\num{19.8e6} parameters). Detailed information on critic and generator implementations is provided in the supplementary.}

%

\section{Experiments and Discussion}

The framework proposed in Algorithm \ref{alg1} is implemented with the publicly TensorFlow framework \cite{tensorflow}. The applications are inner surface depth estimation of cylinder liners, face depth estimation based on the Texas-3DFRD \cite{texas} and body depth synthesis using the SURREAL dataset \cite{surreal}. 
{In this section we benchmark the proposed framework on each dataset and separately present the results, followed by a discussion at the end. As discussed in the introduction, the methods used for comparison are a standard cycleGAN \cite{zhu2017}, gcGAN \cite{fu2019} and CUT \cite{park2020}. For CUT we use the publicly available github repository \footnote{\scriptsize \url{github.com/taesungp/contrastive-unpaired-translation}}. For cycleGAN we remove the novel perceptual loss and handcrafted image filters from our method and replace them with the standard cycleGAN loss. For gcGAN we use the critic and generator implementations of our method, remove the contrary generator and employ up-down-flip as the geometric constraint.}

{
 In our implementation, we set the number of generator updates $n_G$ to 10k, the minibatch size $b$ to 8 and the penalty term $p$ to 100. The number of critic iterations $n_f$ is initially established to be 24 to ensure a good approximation of the Wasserstein-1 distance in the beginning. After 1000 generator updates, it is halved to speed up training. Furthermore, we set $\alpha_f$ to \num{5e-5} and $\alpha_G$ to \num{1e-4}. The influence of the reconstruction term $\lamcyc$ is found for each dataset and method individually by a parameter grid search.}

\subsection{Surface Depth}\label{sec:surface}
This study uses the same database initially proposed in \cite{angermann2021wear} for depth estimation of inner cylinder liner surfaces of large internal combustion engines. 
Depth measurements cover a spatial region of \SI[product-units=power]{1.9 x 1.9}{\mm}, have a dimension of approximately \num{4000x4000} pixels and are acquired using a resource-intensive logistic chain as discussed in the introduction. The profiles denote relative depth with respect to the core area of the surface on a \SI{}{\micro\metre} scale.The RGB data is taken from the same cylinder surfaces with a simple handheld microscope. The RGB measurements cover a region of \SI[product-units=power]{4.2 x 4.2}{\mm} and have a resolution of nearly \num{1024x1024} pixels. Measurement positions are not registered to the depth data. 592 random samples are obtained from each image domain. The RGB and depth data is then augmented separately to nearly 7000 samples via random cropping, flipping and gamma correction \cite{babakhani2015}. To make computation feasible with an \textit{NVIDIA GeForce RTX 2080} GPU, each sample is resized to a dimension of \num{256x256} pixels. 
In order to assess the visual quality between two completely unaligned domains, we also generated depth profiles of 211 additional surface areas and registered them with great effort using shear transformations and a mutual information criterion. These evaluation samples are not included in the training database. During optimization, RGB images and depth profiles are scaled from $[0,255]$ to $[-1,1]$ and from $[-5,5]$ to $[-1,1]$, respectively, whereas evaluation metrics (RMSE and MAE) are calculated on the original depth scale in \SI{}{\micro\meter}.

\begin{table}[thb!]
	\centering
	\scriptsize
	\caption{Unsup. surface depth estimation: The reported metrics are RMSE and MAE of the ground truth and the synthesized depth and are evaluated on unseen data (smaller is better).}
	
	\begin{tabularx}{.99\columnwidth}{l | c |c| c | c }
		\toprule
		\textbf{Method} &\textbf{two-sided}&$\mathbf{\lamcyc}$ &\textbf{RMSE $\pm$ std ( \SI{}{\micro\metre})}&\textbf{MAE $\pm$ std ( \SI{}{\micro\metre})} \\ \midrule
		Proposed & $\checkmark$ &10& $0.751\pm 0.195$& $0.533\pm 0.144$ \\ \midrule
		gcGAN & $\mathbf{x}$&1 & $0.777\pm 0.196$& $0.555\pm 0.145$ \\ \midrule
		cycleGAN & $\checkmark$& 2& $0.833\pm 0.175$& $0.600\pm 0.132$ \\ \midrule
		CUT &$\mathbf{x}$&10 & $1.434\pm 0.402$& $1.074\pm 0.326$
		\\ \bottomrule
	\end{tabularx}
	\label{tab:dt4}
\end{table}


\begin{figure}[htb]
	\centering
	\includegraphics[width=0.82\columnwidth]{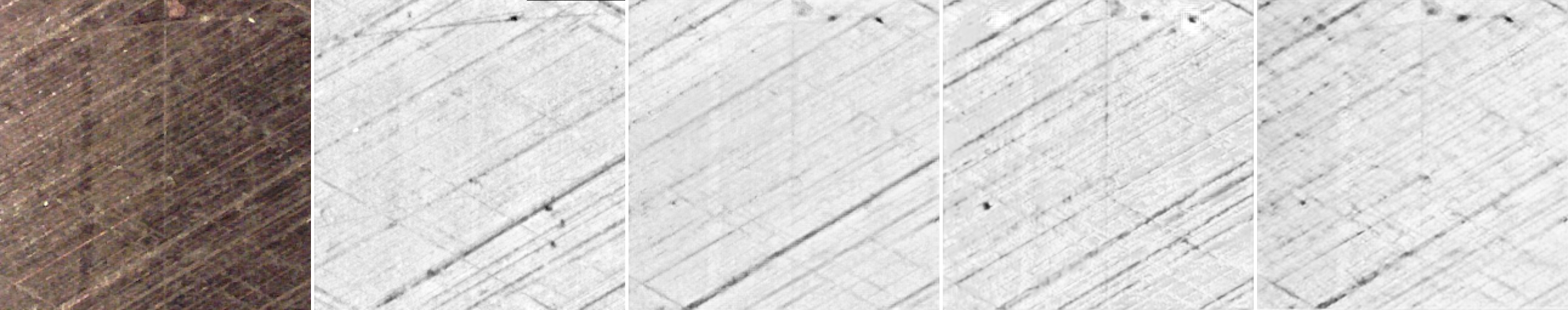}\\ \vspace{.2em}
	\includegraphics[width=0.82\columnwidth]{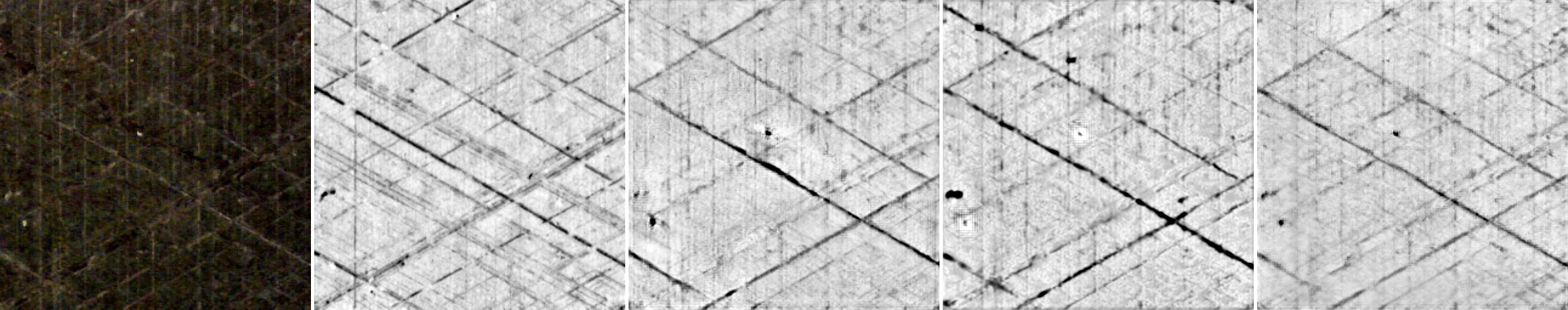}\\ \vspace{.2em}
	\includegraphics[width=0.82\columnwidth]{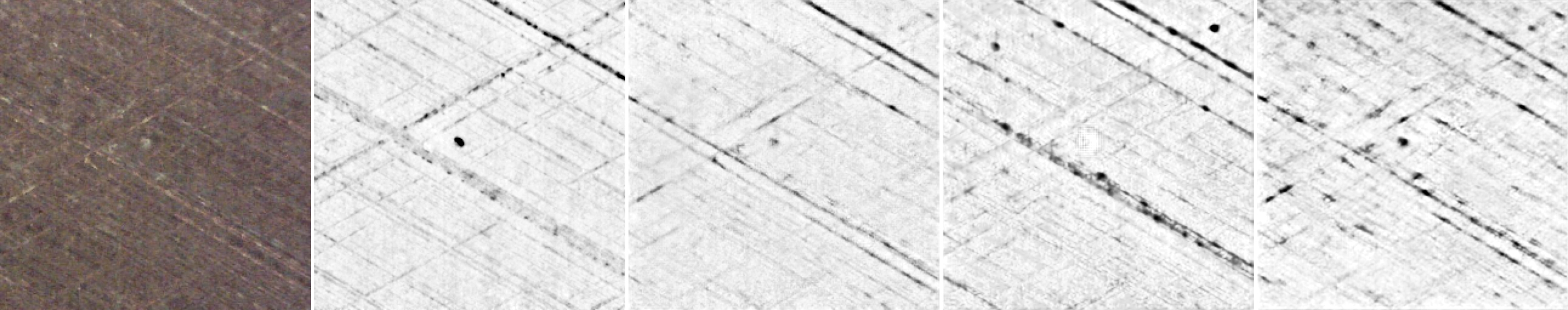}\\ \vspace{.2em}
	\caption{From left to right: Surface RGB input, ground truth and profiles predicted by our method, gcGAN and cycleGAN.}
	\label{fig:dt4}
\end{figure}

\begin{figure}[htb!]
	\centering
	\includegraphics[width=0.22\columnwidth]{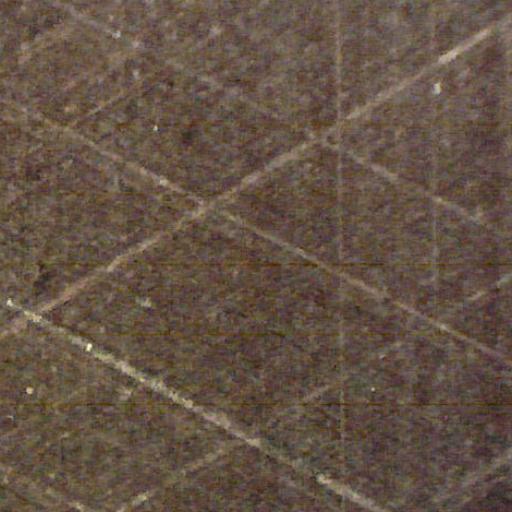}\hfill
	\includegraphics[width=0.35\columnwidth]{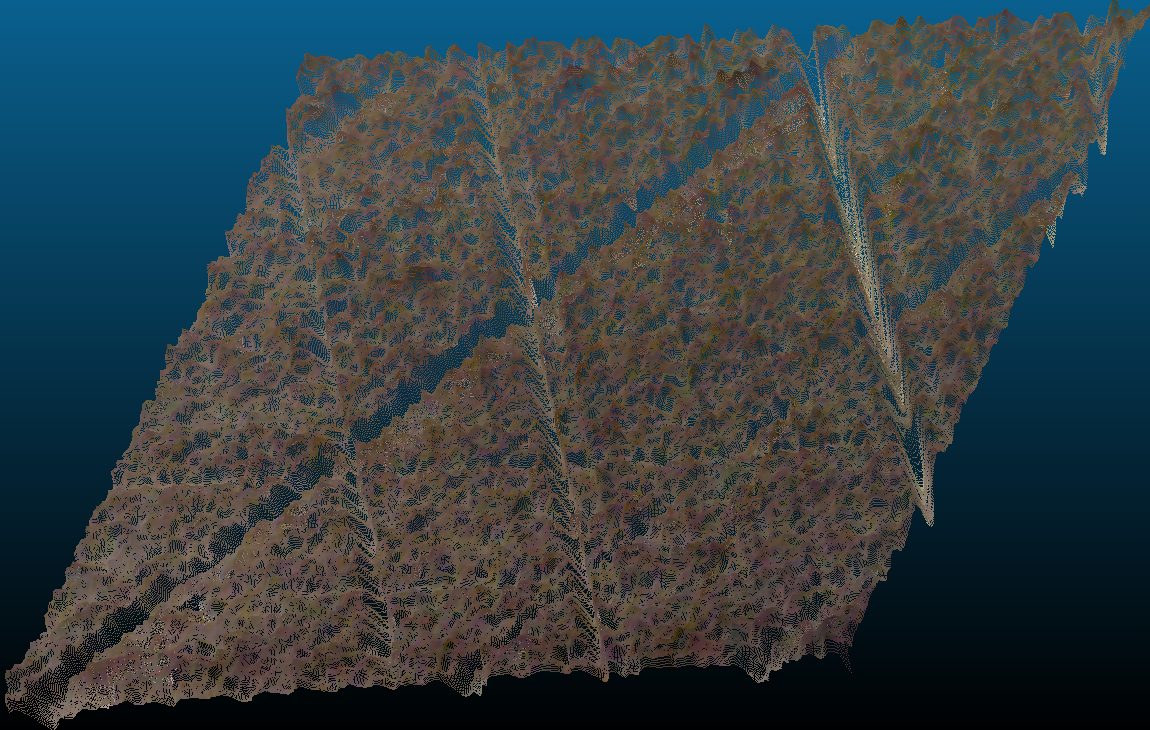} \hfil
	\includegraphics[width=0.35\columnwidth]{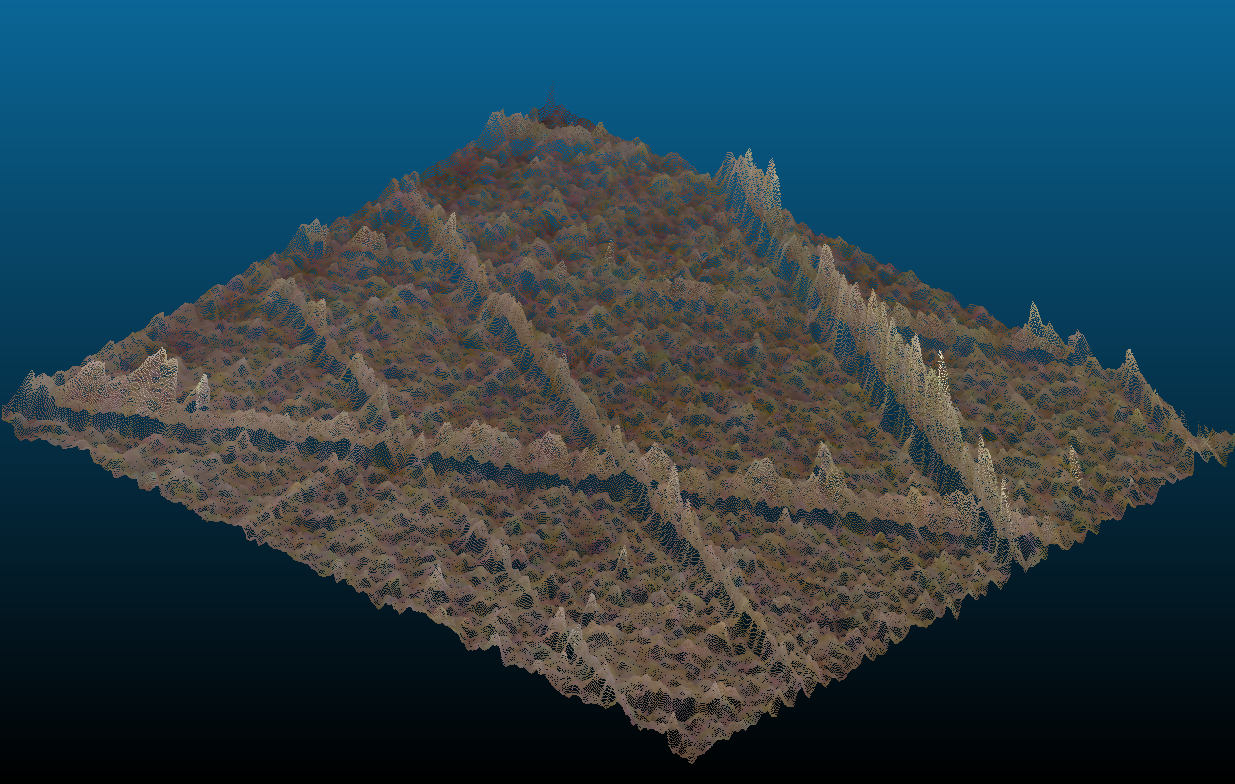}  
	\caption{An instant 3D model generated by our proposed framework provides valuable information on the liner surface condition.}
	\label{fig:dt42}
\end{figure}

\subsection{Face Depth}\label{sec:face}

The Texas-3DFRD \cite{texas} consists of 118 individuals and a variety of facial expressions and corresponding depth profiles are available for each of them. Depth pixels represent absolute depth and their values are in $[0,1]$ where 1 represents the near clipping plane while 0 denotes the background. We randomly select 16 individuals as evaluation data and use the remaining samples as training data. For unsupervised training, we randomly select \SI{50}{\percent} of the training individuals for the input domain and use the depth images of the remaining \SI{50}{\percent} for the target domain. We resize all RGB frames and depth profiles to a dimension of \num{256x256} pixels. Data is augmented via flipping, histogram equalization and Gaussian blur to nearly $6300$ samples per modality. During optimization, RGB images are scaled from $[0,255]$ to $[-1,1]$ and depth profiles are scaled from $[0,1]$ to $[-1,1]$, whereas the evaluation metrics RMSE and MAE are computed on the original depth scale. 
%
\begin{table}[thb!]
	\centering
	\scriptsize
	\caption{Unsup. face depth estimation: The reported metrics are RMSE and MAE of the ground truth and the synthesized depth and are evaluated on unseen data (smaller is better).}
	
	\begin{tabularx}{.93\columnwidth}{l | c |c| c | c }
		\toprule
		\textbf{Method} &\textbf{two-sided}&$\mathbf{\lamcyc}$ &\textbf{RMSE $\pm$ std}&\textbf{MAE $\pm$ std} \\ \midrule
		Proposed & $\checkmark$ &10& $0.068\pm 0.027$& $0.051\pm 0.023$ \\ \midrule
		gcGAN & $\mathbf{x}$&0.3 & $0.078\pm 0.039$& $0.058\pm 0.034$ \\ \midrule
		cycleGAN & $\checkmark$&1 & $0.105\pm 0.049$& $0.073\pm 0.033$ \\ \midrule
		CUT &$\mathbf{x}$&10 & $0.094\pm 0.039$& $0.081\pm 0.042$
		\\ \bottomrule
	\end{tabularx}
	\label{tab:tex}
\end{table}

\vspace{3em}
More experiments on unsupervised facial depth synthesis on the Bosphorus-3DFA \cite{bosphorus}, the CelebAMask-HQ \cite{lee2020} and qualitative comparison to Wu et al. \cite{wu2020} are presented in the supplementary.
\begin{figure}[htb]
	\centering
	\includegraphics[width=0.82\columnwidth]{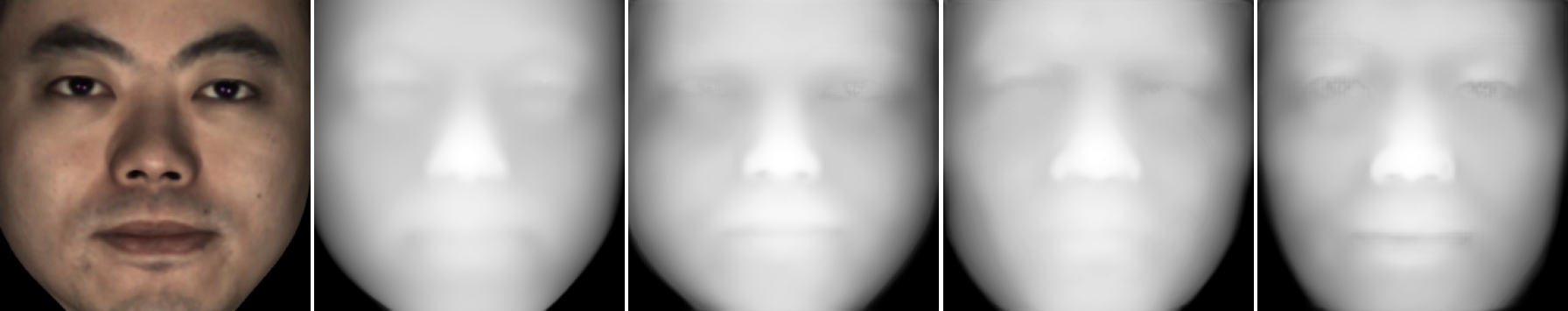}\hspace{-.12em}
	\includegraphics[width=0.1625\columnwidth]{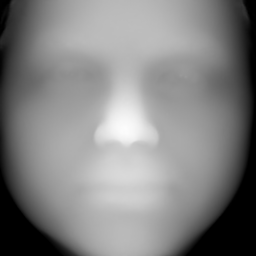}\\ \vspace{.2em}
	\includegraphics[width=0.82\columnwidth]{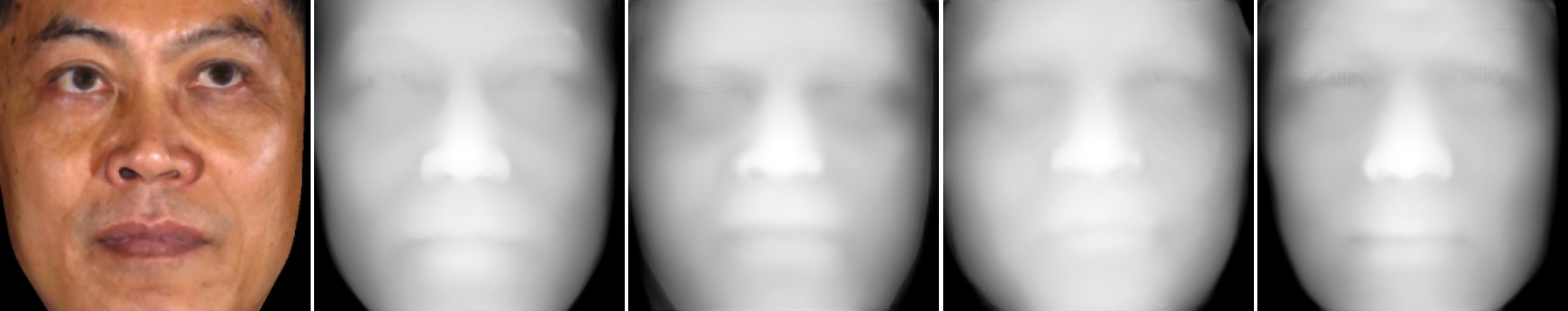}\hspace{-.12em}
	\includegraphics[width=0.1625\columnwidth]{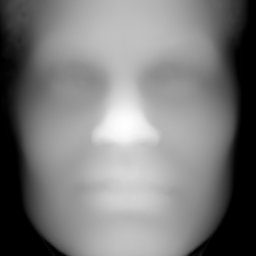}\\ \vspace{.2em}
	\includegraphics[width=0.82\columnwidth]{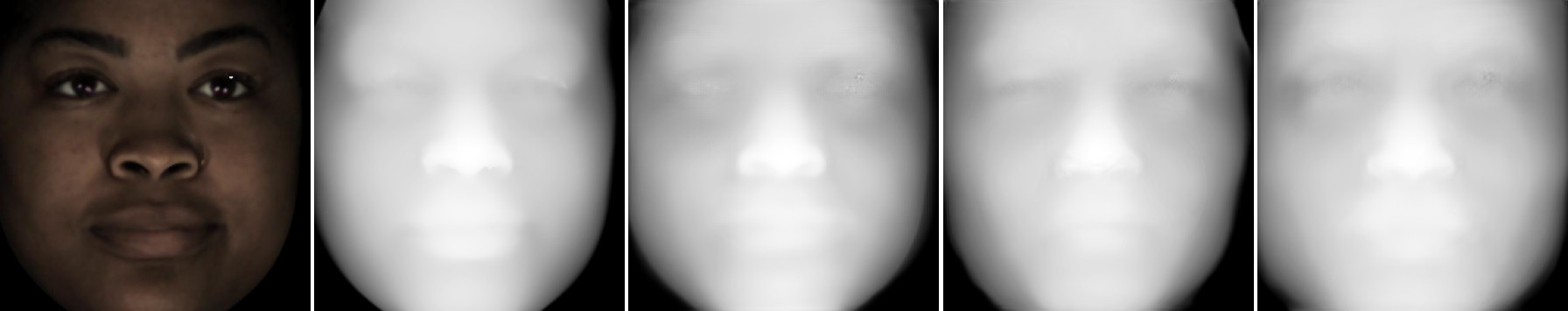}\hspace{-.12em}
	\includegraphics[width=0.1625\columnwidth]{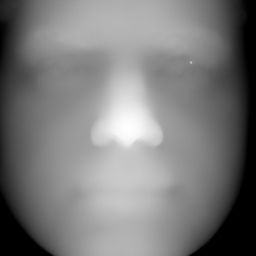}\\ \vspace{.2em}
	\caption{From left to right: Face RGB input, ground truth and profiles predicted by our method, gcGAN, cycleGAN and CUT.}
	\label{fig:tex}
\end{figure}
\begin{figure}[htb!]
	\centering
	\includegraphics[width=0.246\columnwidth]{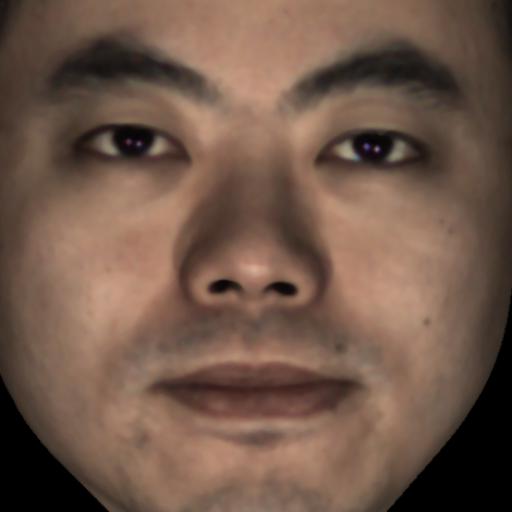}\hfill
	\includegraphics[width=0.22\columnwidth]{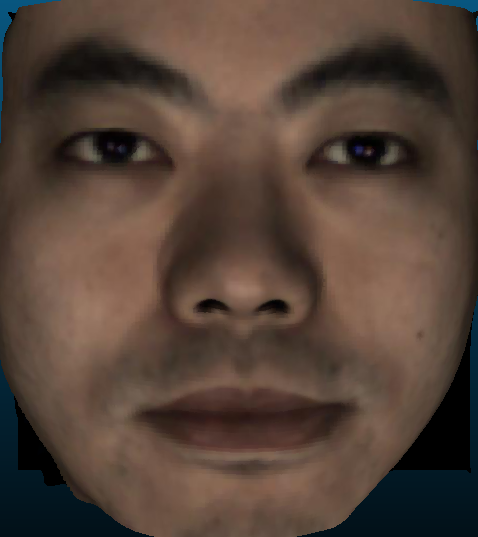} \hfil
	\includegraphics[width=0.22\columnwidth]{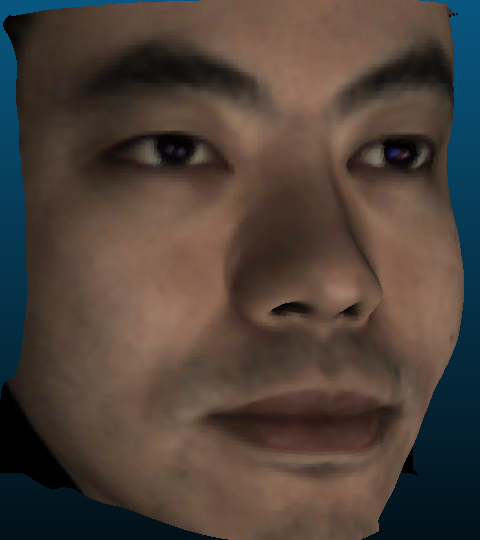} \hfil
	\includegraphics[width=0.22\columnwidth]{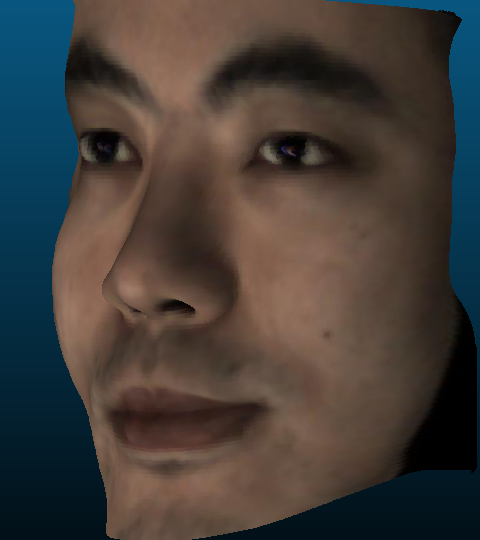}  
	
	\caption{An example of viewpoint augmentation using a 3D face model instantly generated by our proposed framework.}
	\label{fig:tex2}
\end{figure}


\subsection{Body Depth}
The SURREAL dataset \cite{surreal} consists of nearly 68k video clips that show 145 different synthetic subjects performing various actions. The clips consist of 100 RGB frames with perfectly aligned depth profiles that denote real-world camera distance. We use the same train/test split as Varol et al. \cite{surreal}, i.e., we remove nearly 12.5k clips and use the middle frame of each 100-frame clip for evaluation. For the remaining clips, an amount of 2500 clips is randomly selected for training. We choose 20 RGB and 20 depth frames per clip ensuring that RGB and depth frames are disjointed in order to mimic an application without any accurately aligned RGB-depth pairs. This results in approximately 50k samples per modality. We strictly follow the preprocessing pipeline of Varol et al. \cite{surreal}, cropping each frame to the human bounding box and resizing/padding images to a dimension of \num{256x256} pixels. 
In addition, for each image, we subtract the median of depth values to fit the depth images into the range $\pm0.4725$ meters, where values less or equal $-0.4725$ denote background. During optimization, RGB images are scaled from $[0,255]$ to $[-1,1]$ and depth profiles are scaled from $[-0.4725,0.4725]$ to $[-1,1]$, whereas evaluation metrics RMSE and MAE are computed on the original depth scale in meters. 

\begin{table}[thb!]
	\centering
	\scriptsize
	\caption{Unsup. body depth estimation: The reported metrics are RMSE and MAE of the ground truth and the synthesized depth and are evaluated on unseen data (smaller is better).}
	\begin{tabularx}{.965\columnwidth}{l | c |c| c | c }
		\toprule
		\textbf{Method} &\textbf{two-sided}&$\mathbf{\lamcyc}$ &\textbf{RMSE $\pm$ std (\SI{}{\meter})}&\textbf{MAE $\pm$ std (\SI{}{\meter})} \\ \midrule
		Proposed & $\checkmark$ &1& $0.080\pm 0.033$& $0.022\pm 0.020$ \\ \midrule
		gcGAN & $\mathbf{x}$&1 & $0.095\pm 0.036$& $0.030\pm 0.021$ \\ \midrule
		cycleGAN & $\checkmark$&1 & $0.091\pm 0.035$& $0.033\pm 0.019$ \\ \midrule
		CUT &$\mathbf{x}$&1 & $0.183\pm 0.021$& $0.071\pm 0.016$
		\\ \bottomrule
	\end{tabularx}
	\label{tab:sur}
\end{table}

\begin{figure}[htb]
	\centering
	\includegraphics[width=0.82\columnwidth]{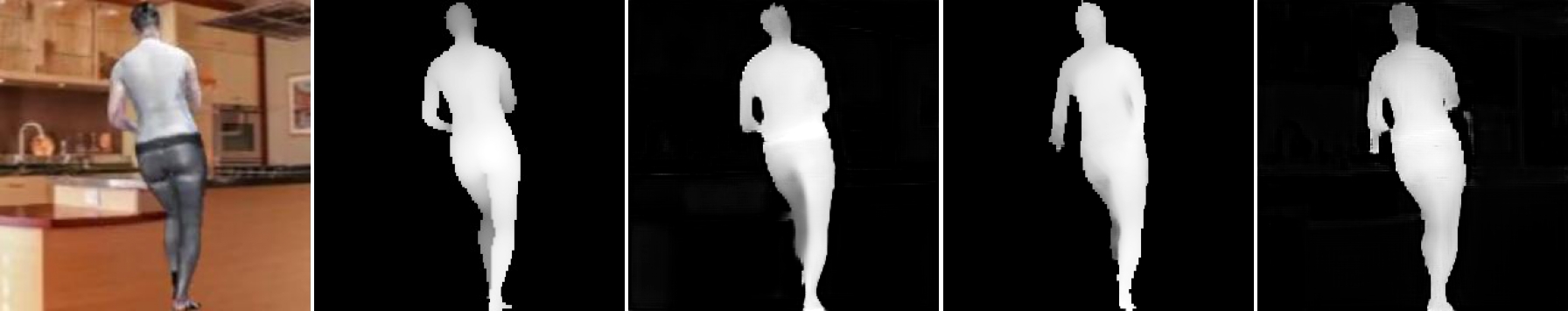}\hspace{-.12em}
	\includegraphics[width=0.1625\columnwidth]{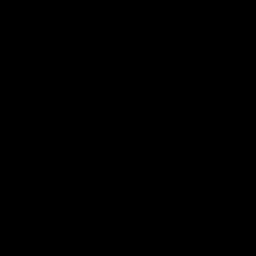}\\ \vspace{.2em}
	\includegraphics[width=0.82\columnwidth]{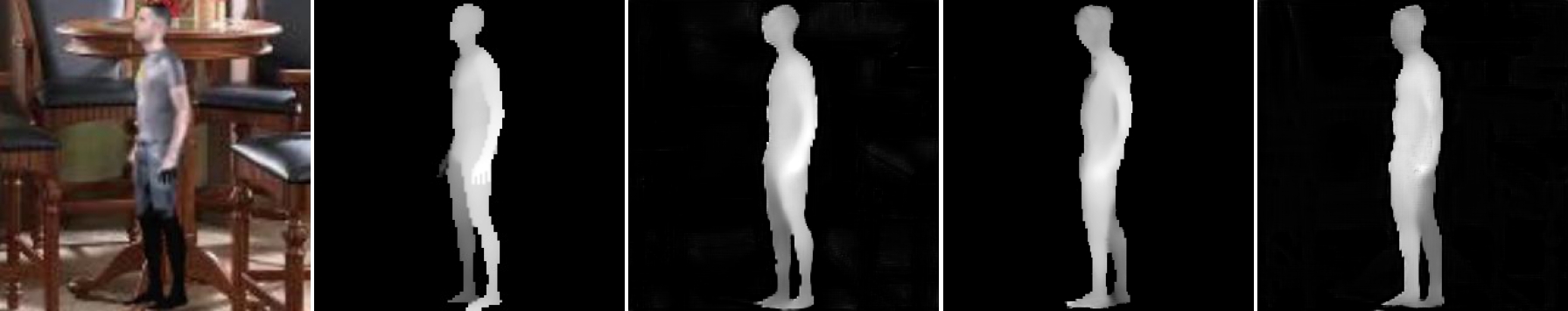}\hspace{-.12em}
	\includegraphics[width=0.1625\columnwidth]{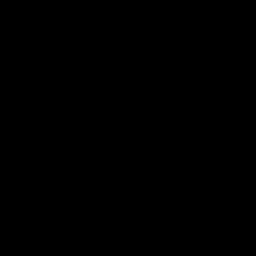}\\ \vspace{.2em}
	\includegraphics[width=0.82\columnwidth]{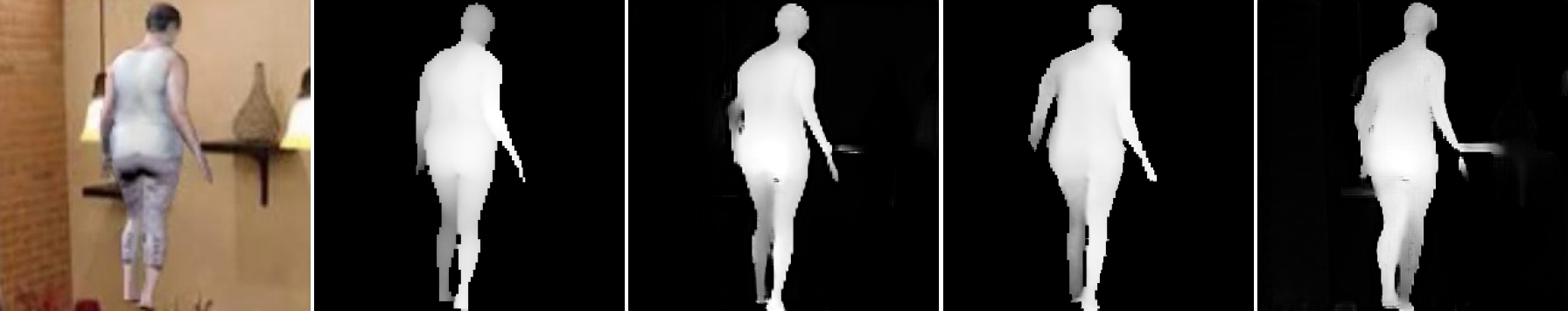}\hspace{-.12em}
	\includegraphics[width=0.1625\columnwidth]{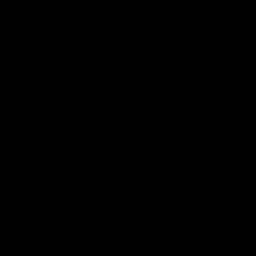}\\ \vspace{.2em}
	\caption{From left to right: Body RGB input, ground truth and profiles predicted by the proposed method, gcGAN, cycleGAN and CUT.}
	\label{fig:sur}
\end{figure}

\subsection{Discussion}
{Quantitative evaluation on unseen test data in \cref{tab:dt4,tab:tex,tab:sur} confirms superiority of the proposed method compared to other state-of-the-art modality transfer methods. Especially the CUT method is not suitable for the depth estimation of planar surfaces and human bodies. Obviously, usage of a novel perceptual reconstruction term in combination with handcrafted image filters is able to overcome the shortcomings of a standard cycle-consistency constraint as explained in Section \ref{sec:loss} and improves depth accuracy significantly. 
	Considering the industrial application, Figure \ref{fig:dt4} indicates that we have been able to synthesize realistic surface depth profiles with an  RMSE of \SI{0.751}{\micro\metre} compared to the registered ground truth.
	In Figure \ref{fig:tex} we observe that predictions coming from our method seem most similar to the ground truth, while the results of cycleGAN and CUT do not correctly reproduce the contours of the input. In Figure \ref{fig:sur} it can be seen that the CUT benchmark completely fails on the SURREAL dataset, which can possibly be attributed to the fact that here, in parallel to the depth estimation, the body must also be segmented.} \\

	Although the proposed method was initially motivated by cycleGAN \cite{zhu2017}, it is important to point out  that replacement of the
	standard cycle-consistency term with perceptual losses and
	usage of appropriate hand-crafted filters in image space is a novel idea that overcomes significant shortcomings of the
	standard cycleGAN architecture in depth estimation that are
	thoroughly discussed in the paper. For depth synthesis of surfaces, faces and human bodies, the RMSE decreases (compared to a standard cycleGAN) about \SI{9.8}{\percent}, \SI{35.2}{\percent} and \SI{12.1}{\percent}, respectively. The proposed method
	has been mainly developed to find a solution to the problem
	of depth synthesis of planar cylinder liner surfaces. The results confirm that the
	framework not only succeeds on the cylinder surface task
	but also significantly improves performance in the field of
	face and whole body depth synthesis compared to state-of-the-art modality transfer methods.
\section{Conclusion}
This paper proposes a framework for fully unsupervised single-shot depth estimation from monocular RGB images based on the Wasserstein-1 distance, a novel perceptual reconstruction loss and handcrafted image filters. The model is comprehensively evaluated on differing depth synthesis tasks without using pairwise RGB and depth data during training. 
The approach provides a reasonable solution for estimating the relative depth of cylinder liner surfaces when generation of paired data is technically not feasible. Moreover, the proposed algorithm also shows promising results when applied to the task of absolute depth estimation of human bodies and faces, thereby proving that it may be generalized to other real-life tasks.
However, one disadvantage of the perceptual reconstruction approach is that four neural networks must be fitted in parallel.
Future work will therefore include the development of one-sided depth synthesis models in an unsupervised manner as well as the application of our approach to other modality transfer tasks.


%


\appendix
\onecolumn

\section{3D Databases - An Overview}
\label{ap:data}
Single-shot depth estimation has become increasingly popular over the last decade of deep learning.  
The first deep learning solutions for depth synthesis were motivated by the development of autonomous driving and localization systems and therefore were initially designed to automatically determine the depth of indoor or outdoor scenes \cite{eigen2014,zhou2017,kundu2018,pilzer2018,godard2019,zhao2019,kwak2020}. Deep convolutional neural networks, trained on large-scale and extensive data sets such as KITTI \cite{kitti} or NYU Depth Dataset v2 \cite{nyu} achieved state-of-the-art results. 
The outdoor video clips of the KITTI dataset can be used for various subtasks in computer vision such as optical flow, object detection, semantic segmentation and depth \cite{zhao2020}. 
Each video sequence of the KITTI dataset consists of stereo image pairs with aligned depth images (LIDAR), which renders the database a common benchmark for unsupervised or self-supervised depth estimation tasks \cite{zhou2017,pilzer2018,godard2019}. The NYU Depth Dataset v2 focuses on monocular sequences of indoor environments, where depth counterparts are obtained with a high quality RGB-D camera. Therefore, this dataset is considered a primary benchmark in supervised monocular depth estimation \cite{eigen2014,kwak2020}.

With the advent of virtual and augmented reality applications, single-image pose estimation and  3D reconstruction of human bodies or body parts received a great amount of attention in the research field of computer vision \cite{jafarian2021}. 
3D information on human faces provides additional benefits for face recognition or detection systems \cite{arslan2019}. 
The Texas-3DFRD \cite{texas} and the Bosphorus-3DFA \cite{bosphorus} are known representatives of paired face RGB-depth data of high quality and include a variety of head poses and emotional expressions. Both databases provide facial landmarks for additional face expression analysis, but with approximately 100 different individuals each, the sets are rather small. A larger number of facial depth models can be derived from 3D synthetic data of human faces as in \cite{khan2020,wood2021}. Leveraging the task to whole body depth estimation is challenging due to the fact that RGB-depth pairs of real individuals are not abundant in many datasets. A small dataset of 25 video clips for detailed human depth estimation is proposed in \cite{tang2019} while a depth dataset of 10 sequences recorded from different viewpoints is published in \cite{vlasic2008}. The Human3.6M dataset \cite{ionescu2014} contains high-resolution depth data from 11 individuals acting in varying scenarios.
\cite{surreal} propose using the approximately 68k video clips of synthetic humans in the large-scale SURREAL dataset for supervised training of human body depth and segmentation models.


\section{Network Details}
\label{networks}
In the following,\textbf{ k} denotes the kernel size, \textbf{s} the stride, and \textbf{channels} the number of layer output channels. \textbf{Input} corresponds to the input of each layer. Network input and output are denoted by $\mc I$ and $\mc O$, respectively, where for a generator network the output channel size equals 1 (RGB-to-depth) or 3 (depth-to-RGB).

\begin{table}[h!]
	\scriptsize
	
	\begin{minipage}{.472\columnwidth}
		\centering
		\caption{\textbf{ResNet18 generator.} The encoder is quite similar to the illustrated architecture in \cite{he2015}. The decoder architecture is a slightly modified version of \cite{godard2019}.  For upsampling, nearest neighbor method is used. Convolution layers followed by an instance normalization are denoted by conv-norm.}
		\begin{tabularx}{.966\columnwidth}{|l | l | l| l | l | l| l| }
			\toprule
			\textbf{name} &\textbf{type}& \textbf{k} &\textbf{s} &\textbf{channels}  &\textbf{input} &\textbf{activation} \\ \midrule
			con1 & conv-norm &7  & 2 & 64 &$\mc I$ &ReLU  \\ \midrule
			max1& maxpool {3x3} &  & 2 & 64 &con1&  \\ \midrule
			res1 & res-block&3&1&64&max1&ReLU \\ \midrule
			res2 & res-block&3&1&64&res1&ReLU \\ \midrule
			res3 & res-block&3&2&128&res2&ReLU \\ \midrule
			res4 & res-block&3&1&128&res3&ReLU \\ \midrule
			res5 & res-block&3&2&256&res4&ReLU \\ \midrule
			res6 & res-block&3&1&256&res5&ReLU \\ \midrule
			res7 & res-block&3&2&512&res6&ReLU \\ \midrule
			res8 & res-block&3&1&512&res7&ReLU \\ \midrule
			
			ups1& upsampling &  & 2 & 512 &res8&  \\ \midrule
			con2& conv-norm&3&1&512&ups1&ELU \\ \midrule
			cct1 & concatenate&&&768&con2,res6& \\ \midrule
			con3& conv-norm&3&1&512&cct1&ELU \\ \midrule
			
			ups2& upsampling &  & 2 & 512 &con3&  \\ \midrule
			con4& conv-norm&3&1&256&ups2&ELU \\ \midrule
			cct2 & concatenate&&&384&con4,res4& \\ \midrule
			con5& conv-norm&3&1&256&cct2&ELU \\ \midrule
			
			ups3& upsampling &  & 2 & 256 &con5&  \\ \midrule
			con6& conv-norm&3&1&128&ups3&ELU \\ \midrule
			cct3 & concatenate&&&192&con6,res2& \\ \midrule
			con7& conv-norm&3&1&128&cct3&ELU \\ \midrule
			
			ups4& upsampling &  & 2 & 128 &con7&  \\ \midrule
			con8& conv-norm&3&1&64&ups4&ELU \\ \midrule
			cct4 & concatenate&&&128&con8,con1& \\ \midrule
			con9& conv-norm&3&1&64&cct4&ELU \\ \midrule
			
			ups5& upsampling &  & 2 & 64 &con9&  \\ \midrule
			con10& conv-norm&3&1&32&ups5&ELU \\ \midrule
			con11& conv-norm&3&1&32&con10&ELU \\ \midrule
			
			$\mc O$ & convolution& 3 & 1 & $3/1$ &con11&tanh \\ \midrule
		\end{tabularx}
		\label{tab:resnet}
	\end{minipage}\hfill
	\begin{minipage}{0.472\columnwidth}
		
		\centering
		\caption{\textbf{PatchGAN critic.} LReLU denotes the Leaky ReLU activation function with slope parameter $0.2$.}
		\begin{tabularx}{.85\columnwidth}{|l | l | l| l | l | l| l| }
			\toprule
			\textbf{name} &\textbf{type}& \textbf{k} &\textbf{s} &\textbf{chns}  &\textbf{input} &\textbf{activation} \\ \midrule
			con1 & convolution & 4 & 1 & 16 &$\mc I$&LReLU  \\ \midrule
			con2 & convolution & 4 & 1 & 16 &con1&LReLU \\ \midrule
			con3 & convolution & 4 & 2 & 32 &con2&LReLU  \\ \midrule
			con4 & convolution & 4 & 1 & 32 &con3&LReLU \\ \midrule
			con5 & convolution & 4 & 2 & 64 &con4&LReLU \\ \midrule
			con6 & convolution & 4 & 1 & 64 &con5&LReLU \\ \midrule
			con7 & convolution & 4 & 2 & 128 &con6&LReLU \\ \midrule
			con8 & convolution & 4 & 1 & 128 &con7&LReLU \\ \midrule
			con9 & convolution & 4 & 2 & 256 &con8&LReLU \\ \midrule
			con10 & convolution& 4 & 1 & 256 &con9&LReLU \\ \midrule
			con11 & convolution& 4 & 2 & 512 &con10&LReLU \\ \midrule
			con12 & convolution& 4 & 1 & 512 &con11&LReLU \\ \midrule
			$\mc O$ & convolution& 4 & 1 & 1 &con12&linear\\ \bottomrule
		\end{tabularx}
	\end{minipage}\\[4em]
	
	\begin{minipage}{0.472\columnwidth}
		\centering
		\caption{\textbf{Residual block.} A residual block (res-block) with kernel size $k$, stride $s$ and channel size $c$ is implemented as follows:}
		\begin{tabularx}{.93\columnwidth}{|l | l | l| l | l | l| l| }
			\toprule
			\textbf{name} &\textbf{type}& \textbf{k} &\textbf{s} &\textbf{channels}  &\textbf{input} &\textbf{activation} \\ \midrule
			con1 & conv-norm & $k$ & $s$ & $c$ &$\mc I$&ReLU  \\ \midrule
			con2 & conv-norm & $k$ & $s$ & $c$ & con1& \\ \midrule
			skip & conv-norm & 1 & $s$ & $c$ & $\mc I$& \\ \midrule
			add & addition & & & $c$ & con2,skip& \\ \midrule
			$\mc O$ & activation &&&$c$&add&ReLU\\ \bottomrule
		\end{tabularx}
	\end{minipage}

\end{table}



\newpage

\section{Facial Depth Estimation on Bosphorus-3DFA and CelebAMask-HQ}

\label{ap:tns}
Section 4.2 demonstrates the plausibility of our proposed framework for fully unsupervised facial depth estimation using  the small Texas-3DFRD \cite{texas}. Obviously, the shooting position of the portrayed faces is always constant. The data set consists exclusively of frontal views, the illumination direction is consistent and all images are individually cropped to the facial region. However, the goal of this section is to train a model that is capable of generating depth profiles from arbitrary portrait images that are at least sufficient for reasonable viewpoint augmentation. To accomplish this, we make use of the following two data sets: the Bosphorus Database for 3D Face Analysis (Bosphorus-3DFA) \cite{bosphorus} and the CelebAMask-HQ \cite{lee2020} that records face portraits.\\

\begin{figure}[htb]
	\centering
	\includegraphics[width=0.11\columnwidth]{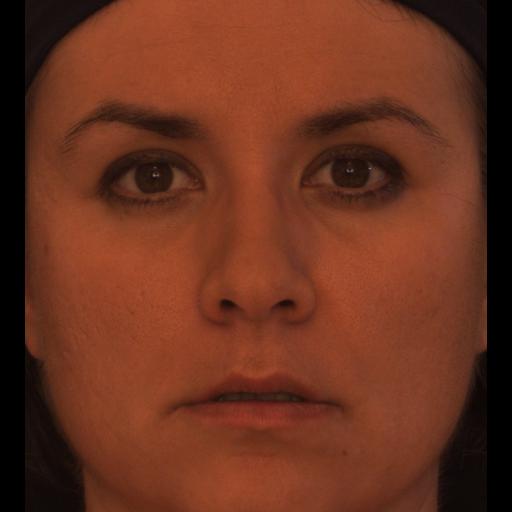}\hspace{.04em}
	\includegraphics[width=0.11\columnwidth]{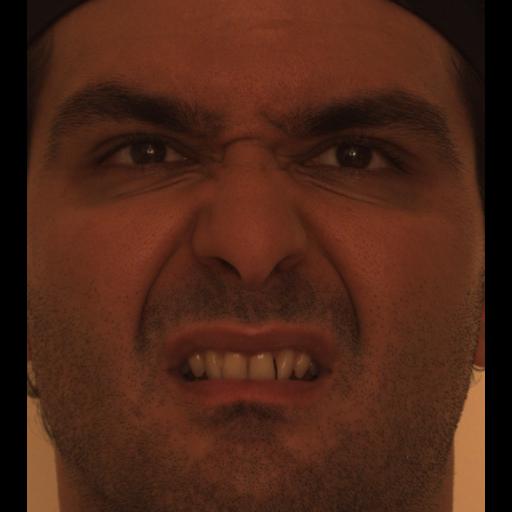}\hspace{.04em}
	\includegraphics[width=0.11\columnwidth]{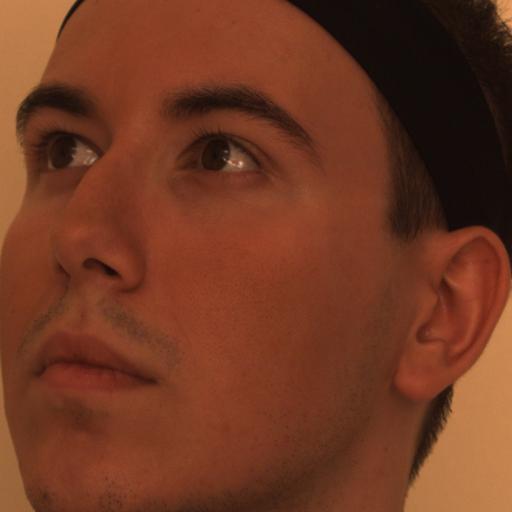}\hspace{.04em}
	\includegraphics[width=0.11\columnwidth]{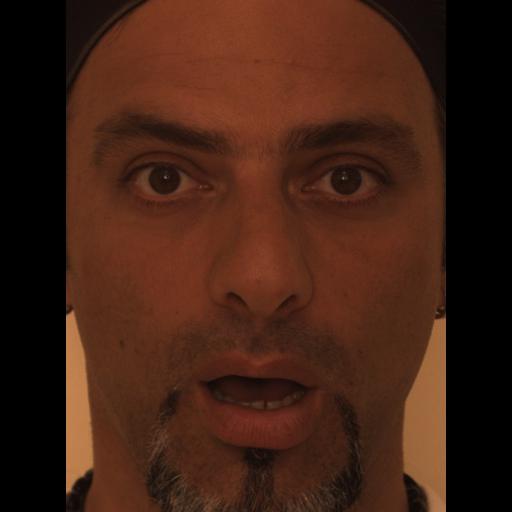}\hfill
	\includegraphics[width=0.11\columnwidth]{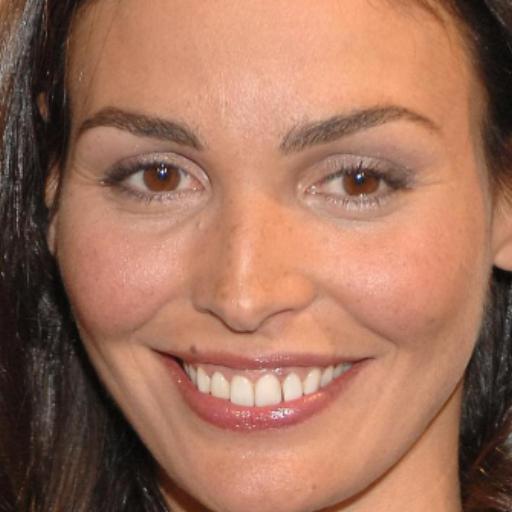}\hspace{.04em}
	\includegraphics[width=0.11\columnwidth]{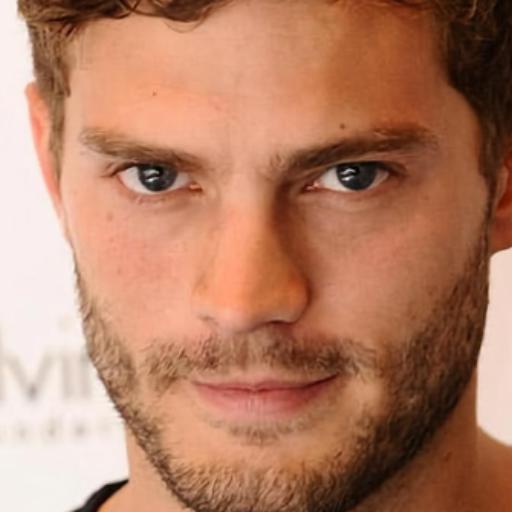}\hspace{.04em}
	\includegraphics[width=0.11\columnwidth]{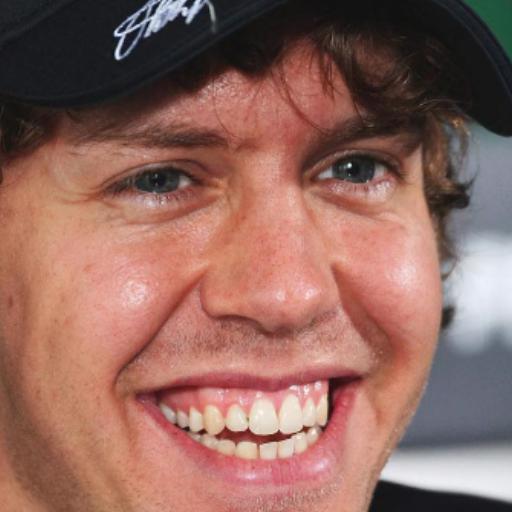}\hspace{.04em}
	\includegraphics[width=0.11\columnwidth]{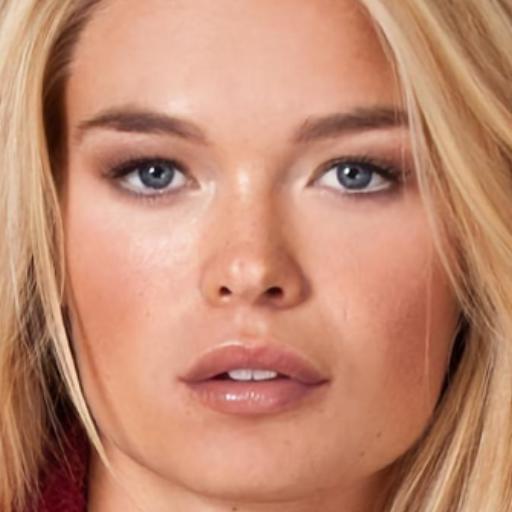}
	\caption{Left: RGB samples of the Bosphorus-3DFA \cite{bosphorus}. Right: Samples of the CelebAMask-HQ \cite{lee2020}.}
	\label{fig:ap1}
\end{figure}

The Bosphorus-3DFA consists of 105 individuals, where for each person, in contrast to the Texas-3DFRD, varying poses, different head rotations and occlusions (e.g. eyeglasses, long hair) are available. Pixel-aligned depth samples represent absolute depth and are preprocessed to the range $[0,1]$.
Analogously to Section 4.2, we resize all RGB frames and depth profiles to a dimension of \num{256x256} and conduct data augmentation via random cropping. This results into 11k samples per modality. Although this database now contains different positions and face expressions, the decisive disadvantage is that all images were taken with constant lighting and with the same background (cf. Figure \ref{fig:ap1}). Therefore, we add the CelebAMask-HQ to our experiment.\\

The CelebAMask-HQ is a large-scale facial portrait dataset with high-resolution face images of 30k celebrities selected from the CelebA dataset \cite{yang2015}. Each sample is provided with a segmentation mask of face attributes, and therefore this database is used to train and evaluate face analysis, face recognition and segmentation algorithms. In our opinion, this database is particularly well suited for depth prediction of arbitrary portraits, as it consists of RGB images with different exposures and different image backgrounds. Furthermore, all images are already cropped to a face-bounding box. We randomly select 10k RGB frames and resize them to a dimension of \num{256x256}.
The RGB images of the Bosphorus-3DFA and all samples of the CelebAMask-HQ are used as training data for the RGB domain, the depth profiles of the Bosphorus-3DFA are used for the depth domain. 
We conduct unsupervised training of our proposed framework as described in Algorithm 1. During optimization, RGB images are scaled from $[0,255]$ to $[-1,1]$ and depth profiles are scaled from $[0,1]$ to $[-1,1]$. \\

{We qualitatively benchmark our proposed method against Wu et al. \cite{wu2020}, where a method for fully unsupervised 3D modeling out of single images is introduced. To be more exact, a network is proposed that factors each input RGB into depth, albedo, viewpoint and illumination. In order to disentangle these different components without any supervision via paired data, stereo pairs or video sequences, Wu et al. make use of the fact that faces have in principle a symmetric structure. Thus, this proposed method for image disentanglement can also be applied to other object categories, provided that these have a symmetrical structure. The research of Wu et al. is one of the few works which has especially been developed for 3D modeling and where no supervision via paired RGB-depth data or availability of video sequences and stereo images is possible. The method has has been evaluated on several databases of cat and human faces, also including the CelebA. For visual comparison we make use of the publicly available demo version \footnote{\tiny \url{https://www.robots.ox.ac.uk/~vgg/blog/unsupervised-learning-of-probably-symmetric-deformable-3d-objects-from-images-in-the-wild.html?image=004_face&type=human}} provided by the authors.}

\begin{figure}[htb]
	\centering
	
	\includegraphics[width=0.107\columnwidth]{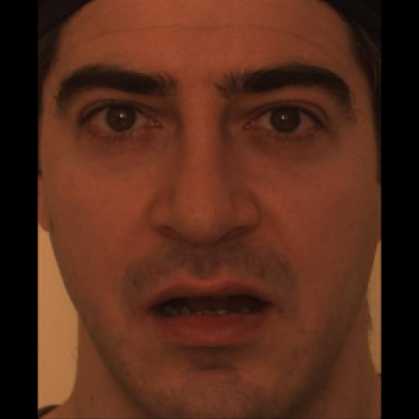}\hspace{.3em}
	\includegraphics[width=0.107\columnwidth]{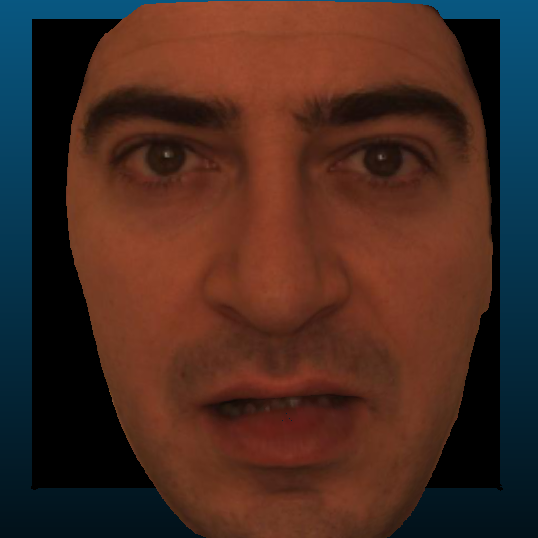}\hfill
	\includegraphics[width=0.107\columnwidth]{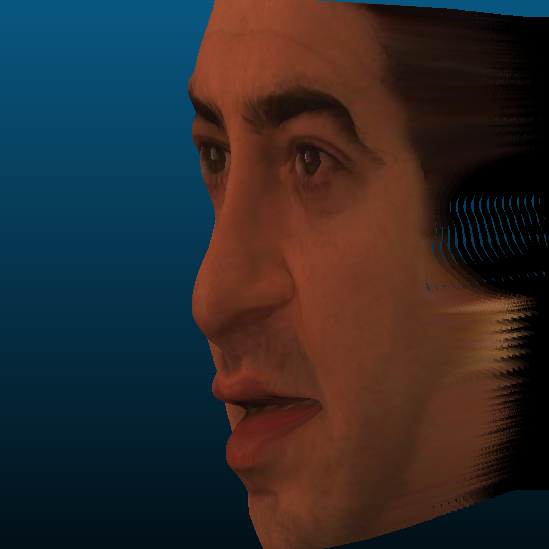}\hfill
	\includegraphics[width=0.107\columnwidth]{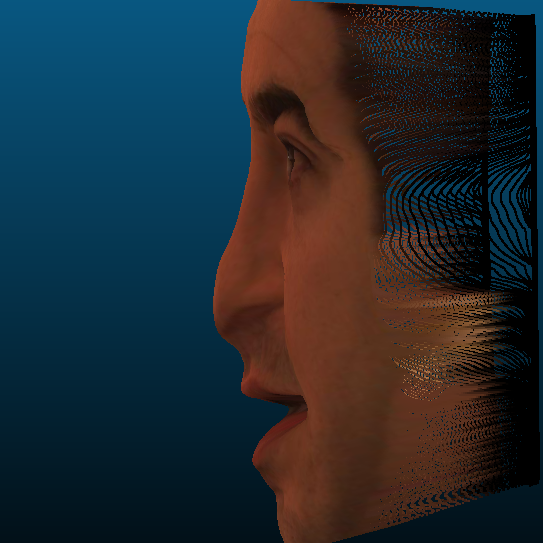}\hfill
	\includegraphics[width=0.107\columnwidth]{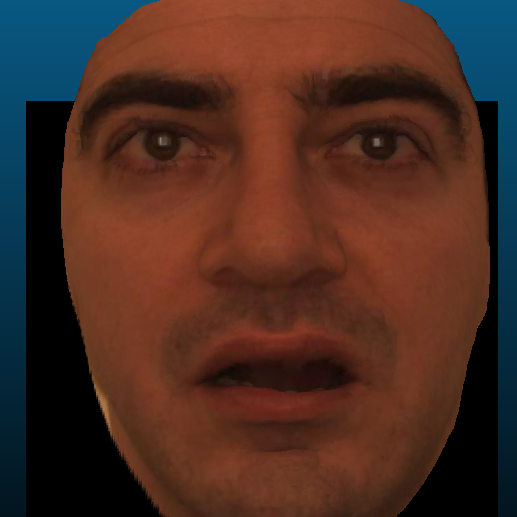}\hspace{.3em}
	\includegraphics[width=0.107\columnwidth]{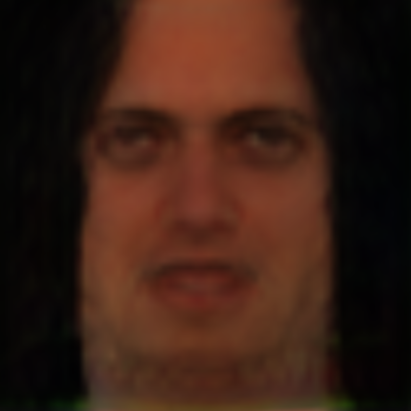}\hfill
	\includegraphics[width=0.107\columnwidth]{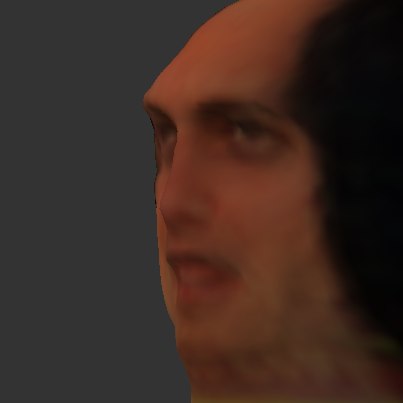}\hfill
	\includegraphics[width=0.107\columnwidth]{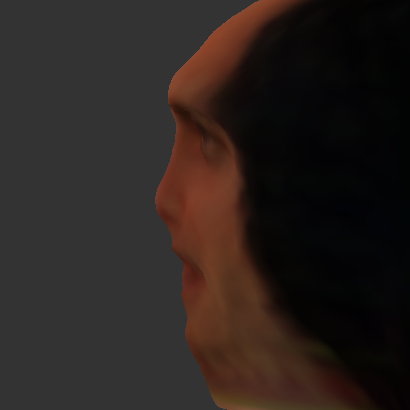}\hfill
	\includegraphics[width=0.107\columnwidth]{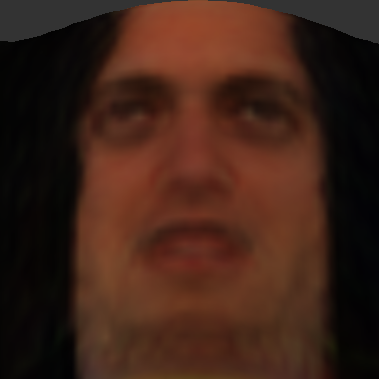}

	\includegraphics[width=0.107\columnwidth]{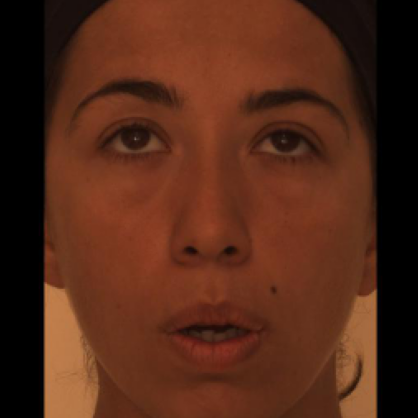}\hspace{.3em}
	\includegraphics[width=0.107\columnwidth]{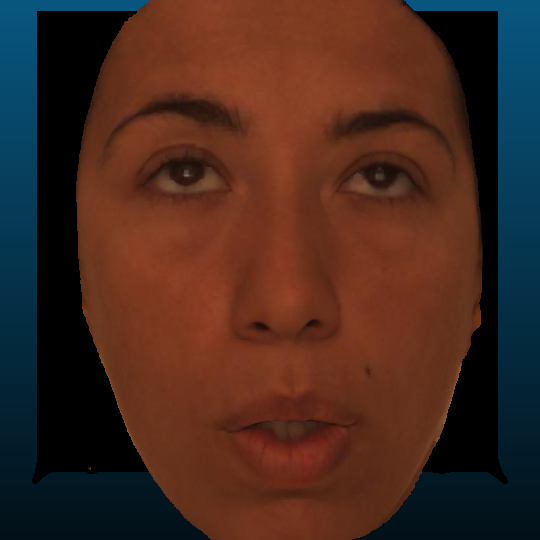}\hfill
	\includegraphics[width=0.107\columnwidth]{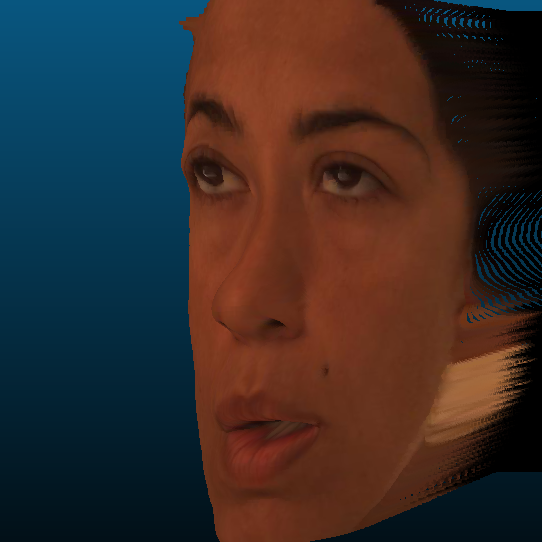}\hfill
	\includegraphics[width=0.107\columnwidth]{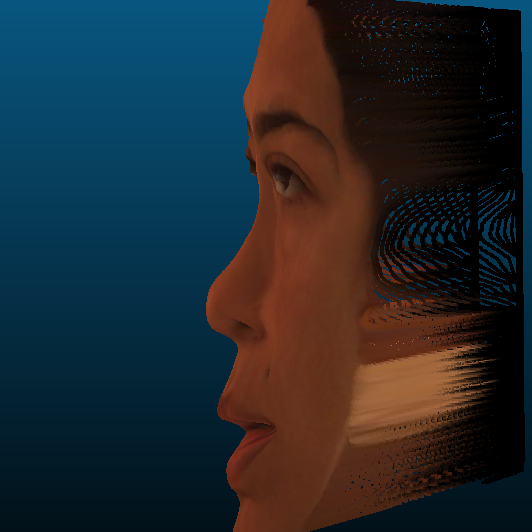}\hfill
	\includegraphics[width=0.107\columnwidth]{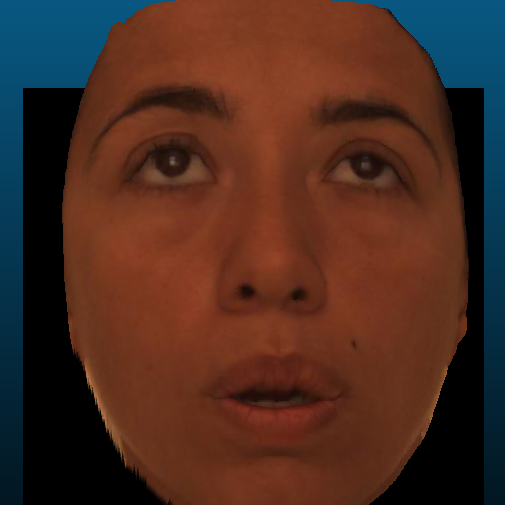}\hspace{.3em}
	\includegraphics[width=0.107\columnwidth]{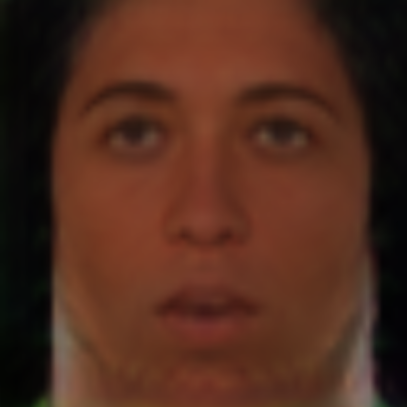}\hfill
	\includegraphics[width=0.107\columnwidth]{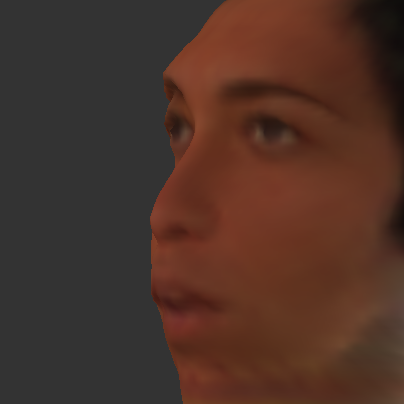}\hfill
	\includegraphics[width=0.107\columnwidth]{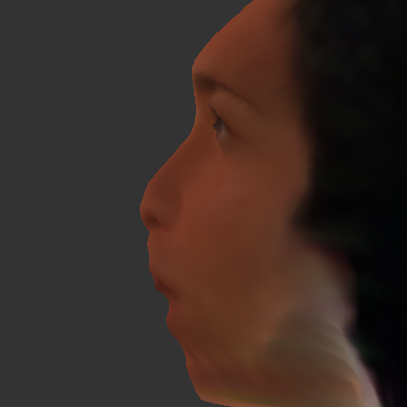}\hfill
	\includegraphics[width=0.107\columnwidth]{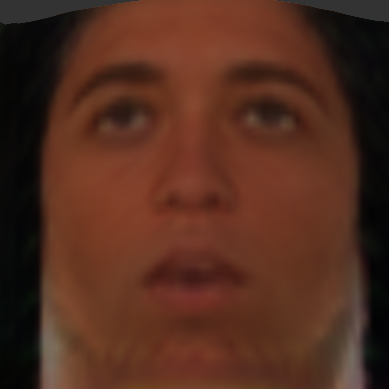}

	\includegraphics[width=0.107\columnwidth]{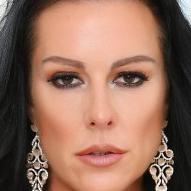}\hspace{.3em}
	\includegraphics[width=0.107\columnwidth]{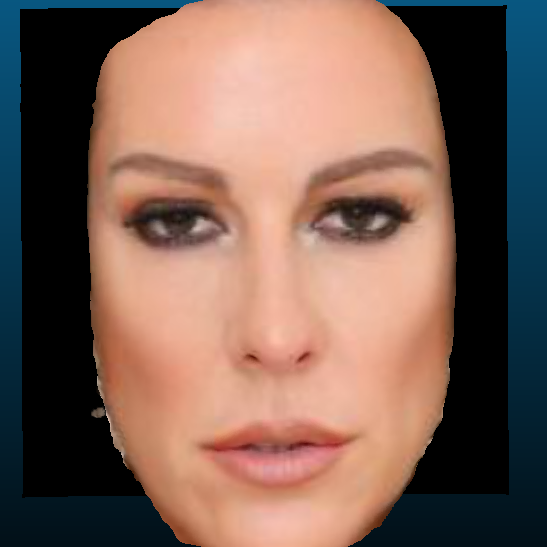}\hfill
	\includegraphics[width=0.107\columnwidth]{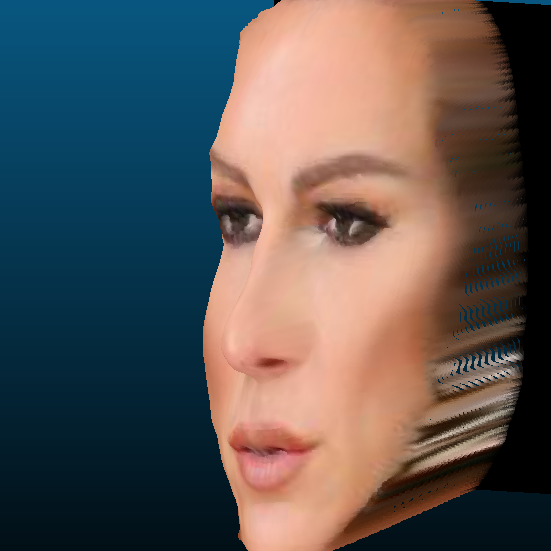}\hfill
	\includegraphics[width=0.107\columnwidth]{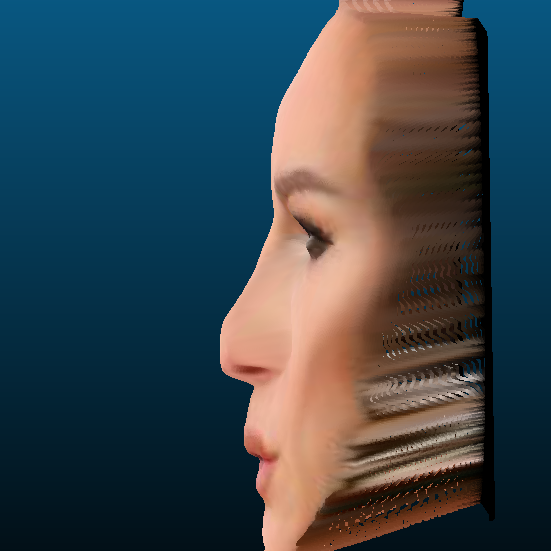}\hfill
	\includegraphics[width=0.107\columnwidth]{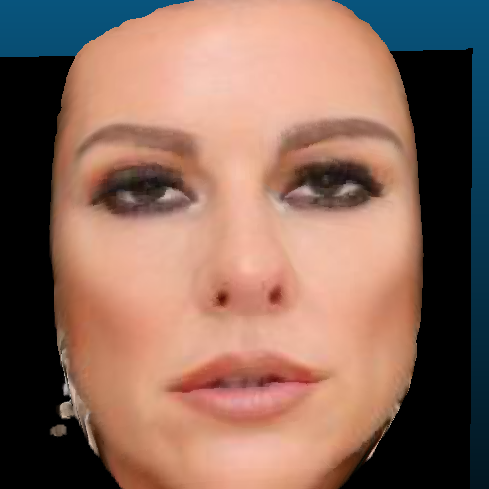}\hspace{.3em}
	\includegraphics[width=0.107\columnwidth]{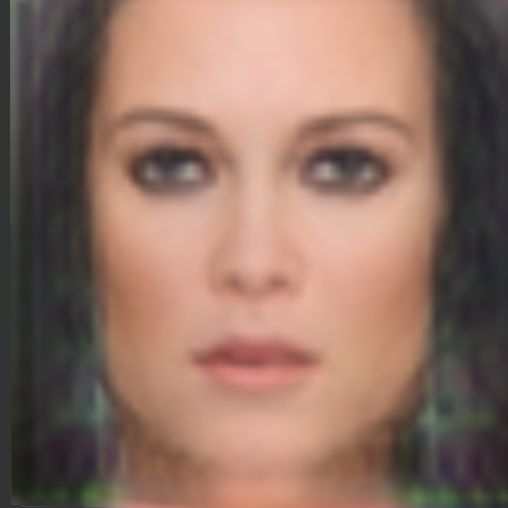}\hfill
	\includegraphics[width=0.107\columnwidth]{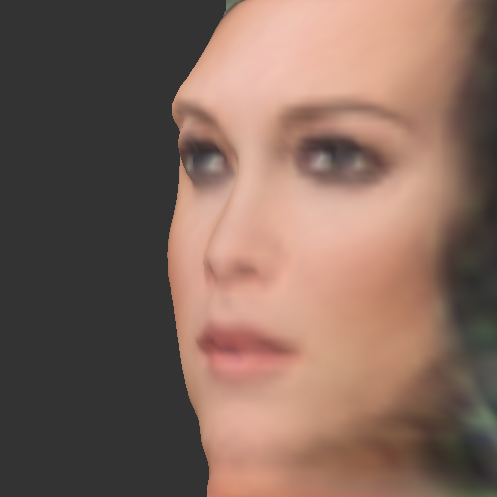}\hfill
	\includegraphics[width=0.107\columnwidth]{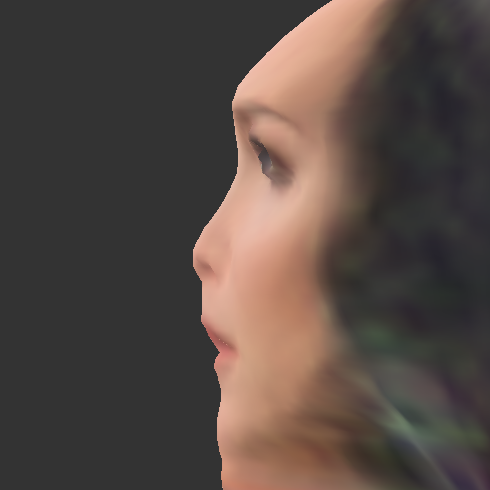}\hfill
	\includegraphics[width=0.107\columnwidth]{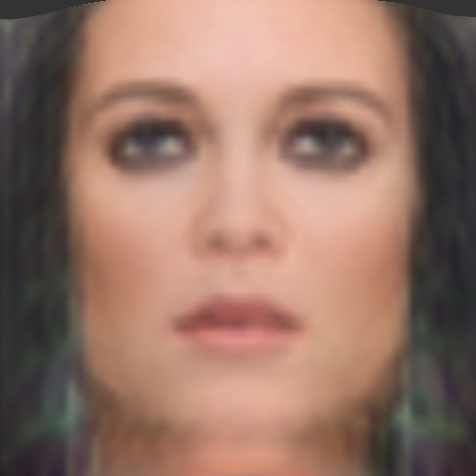}

	\includegraphics[width=0.107\columnwidth]{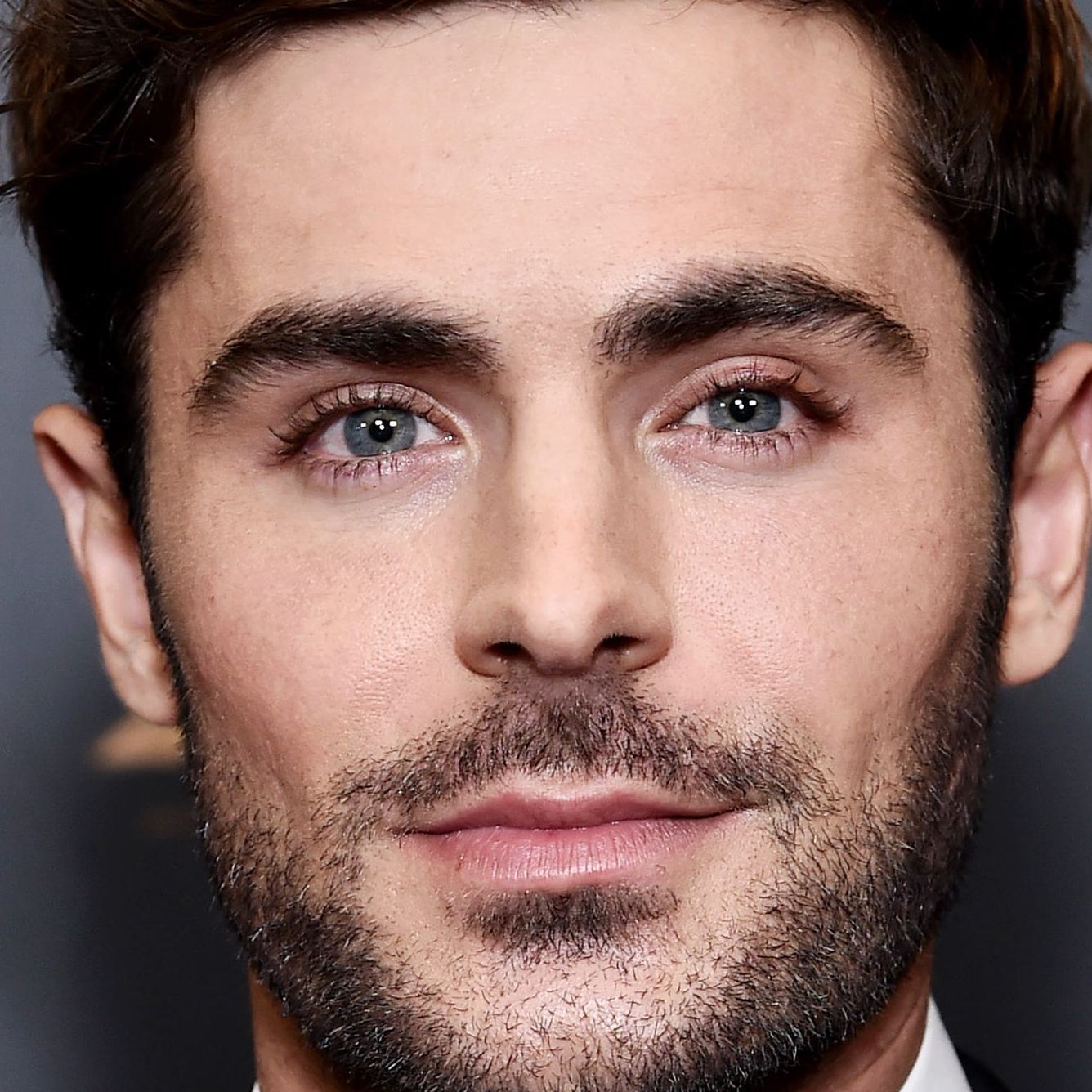}\hspace{.3em}
	\includegraphics[width=0.107\columnwidth]{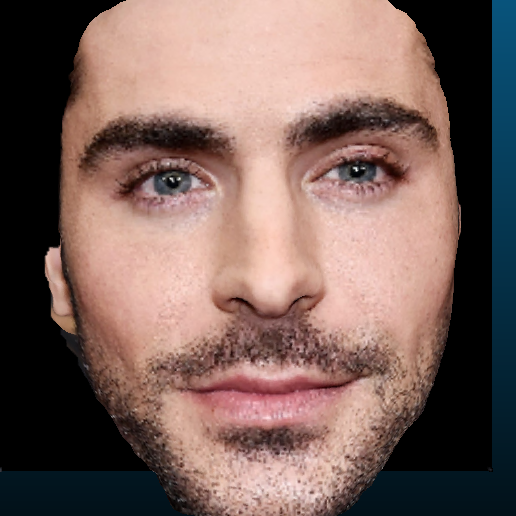}\hfill
	\includegraphics[width=0.107\columnwidth]{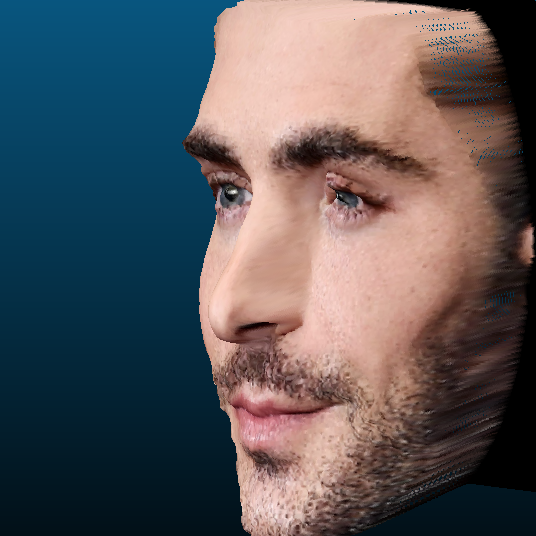}\hfill
	\includegraphics[width=0.107\columnwidth]{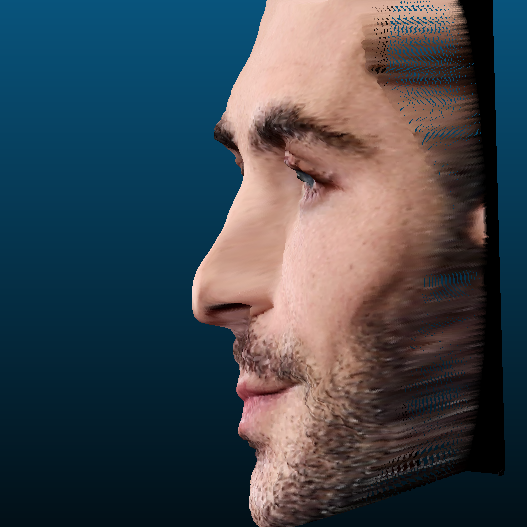}\hfill
	\includegraphics[width=0.107\columnwidth]{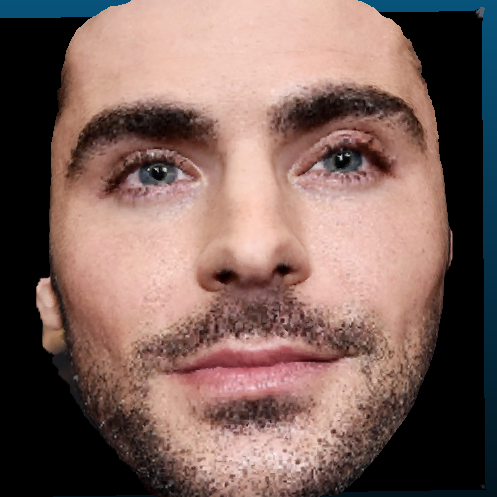}\hspace{.3em}
	\includegraphics[width=0.107\columnwidth]{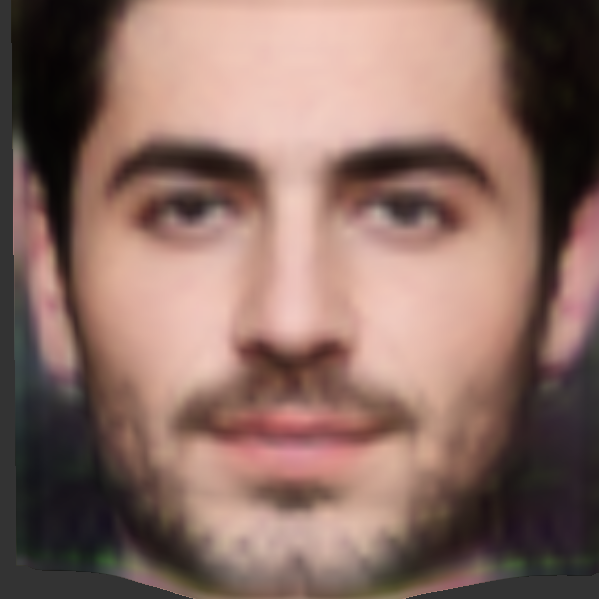}\hfill
	\includegraphics[width=0.107\columnwidth]{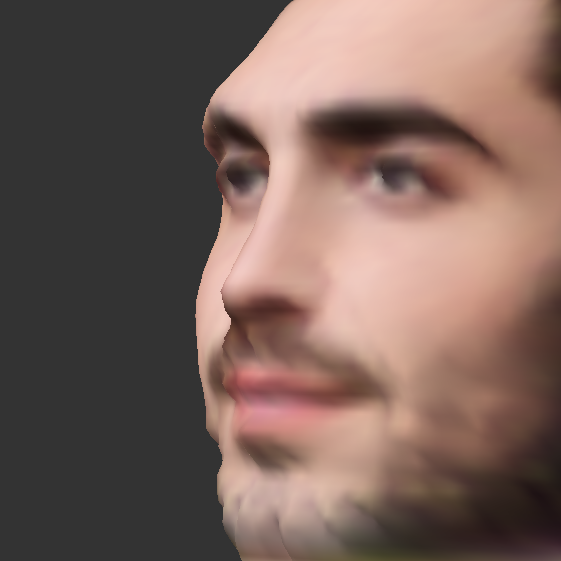}\hfill
	\includegraphics[width=0.107\columnwidth]{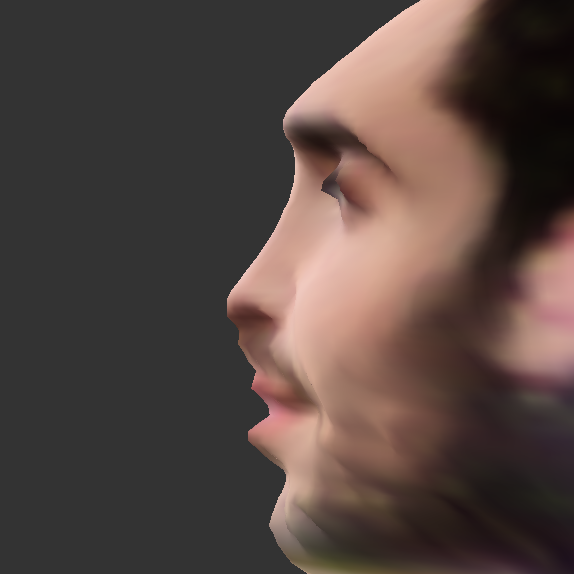}\hfill
	\includegraphics[width=0.107\columnwidth]{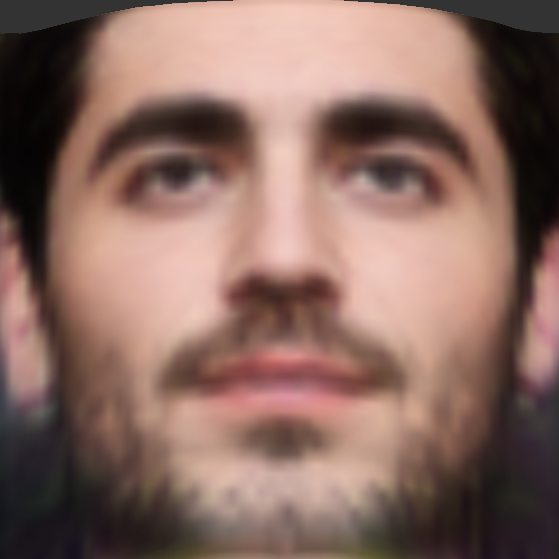}

	\includegraphics[width=0.107\columnwidth]{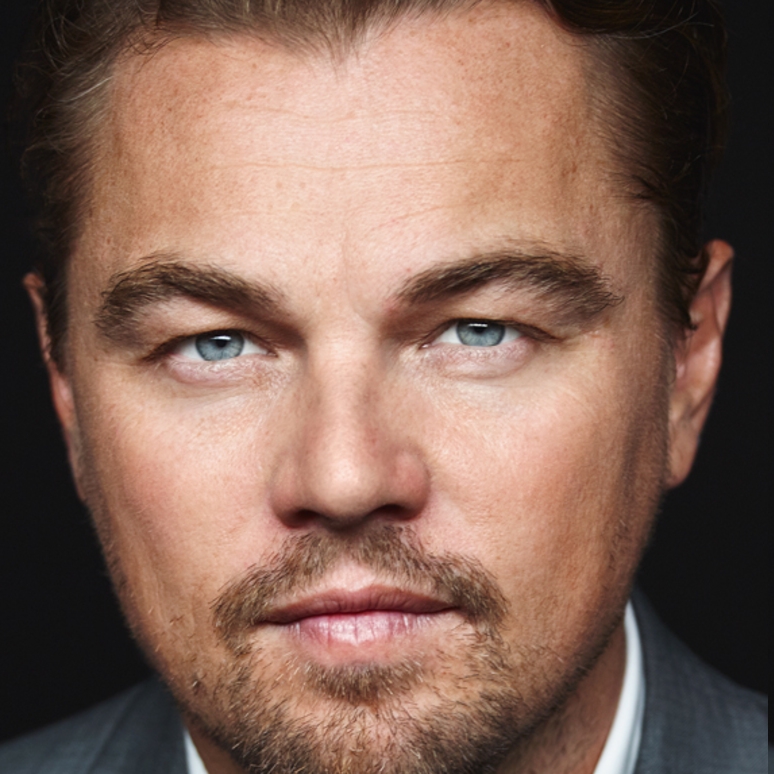}\hspace{.3em}
	\includegraphics[width=0.107\columnwidth]{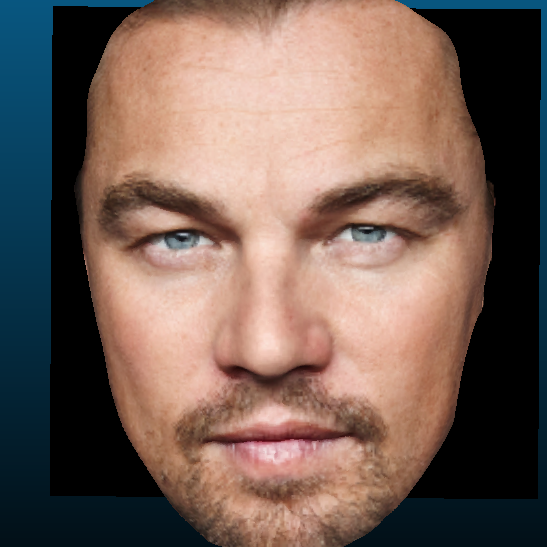}\hfill
	\includegraphics[width=0.107\columnwidth]{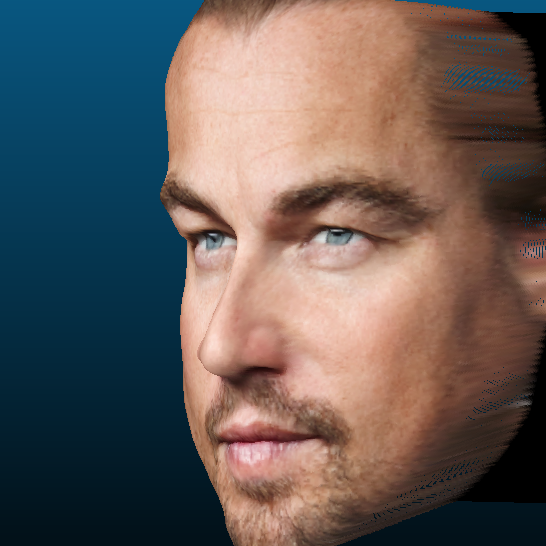}\hfill
	\includegraphics[width=0.107\columnwidth]{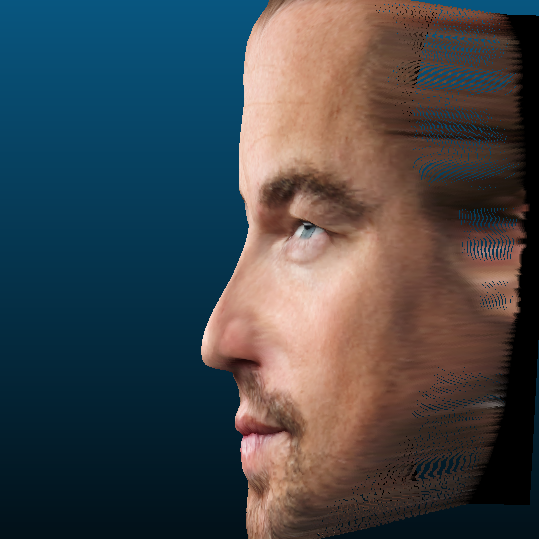}\hfill
	\includegraphics[width=0.107\columnwidth]{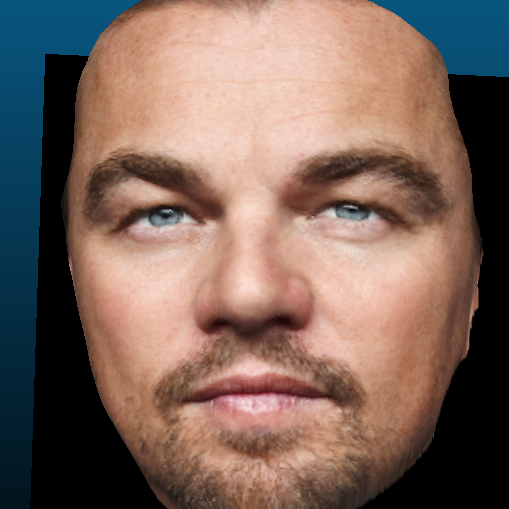}\hspace{.3em}
	\includegraphics[width=0.107\columnwidth]{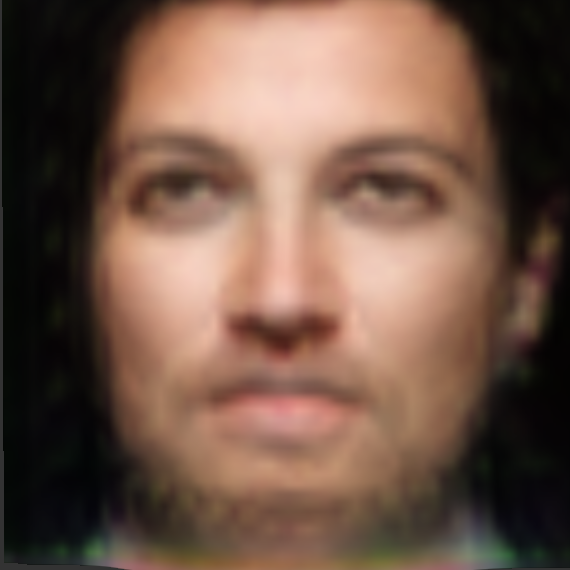}\hfill
	\includegraphics[width=0.107\columnwidth]{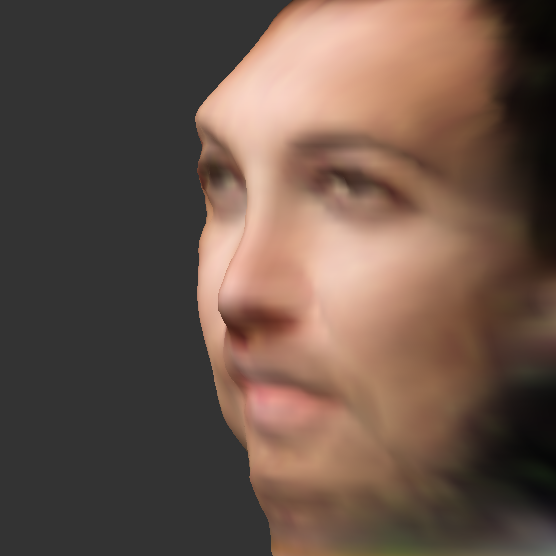}\hfill
	\includegraphics[width=0.107\columnwidth]{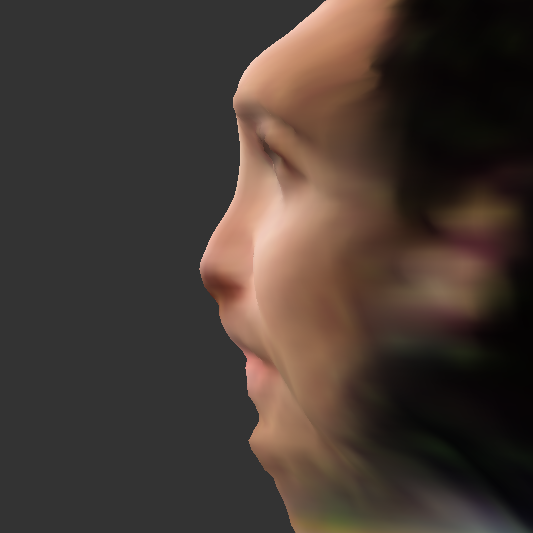}\hfill
	\includegraphics[width=0.107\columnwidth]{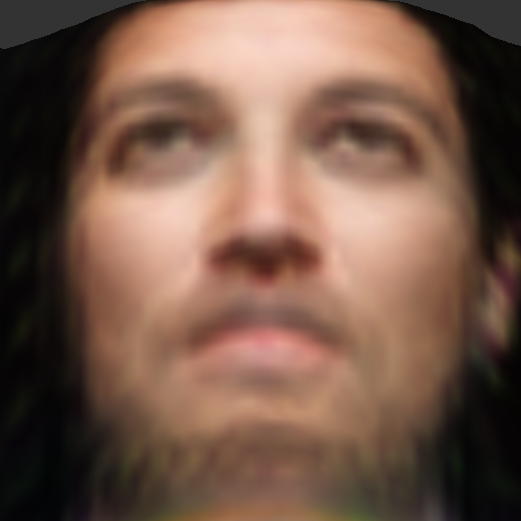}

	\includegraphics[width=0.107\columnwidth]{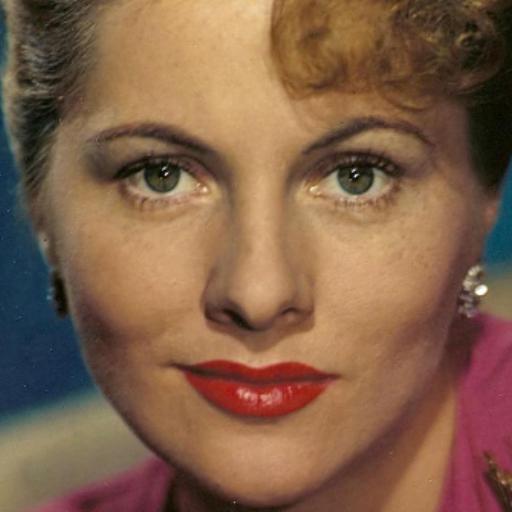}\hspace{.3em}
	\includegraphics[width=0.107\columnwidth]{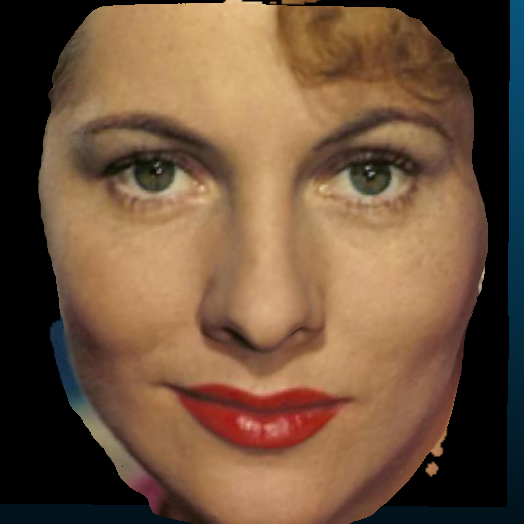}\hfill
	\includegraphics[width=0.107\columnwidth]{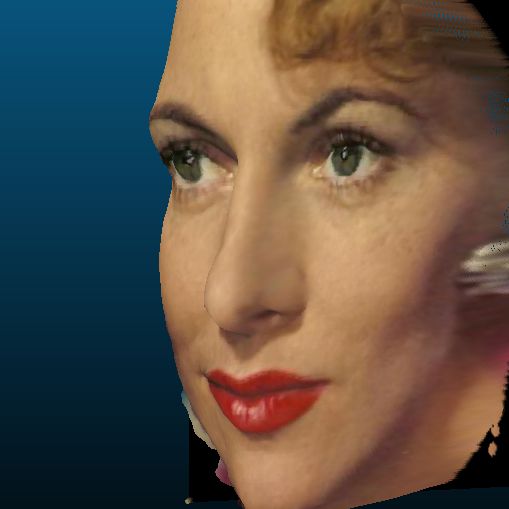}\hfill
	\includegraphics[width=0.107\columnwidth]{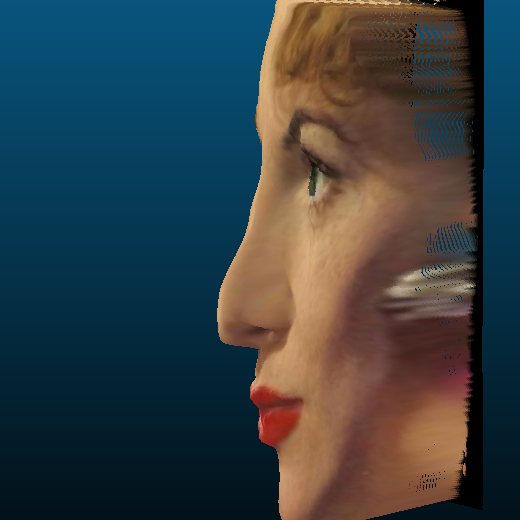}\hfill
	\includegraphics[width=0.107\columnwidth]{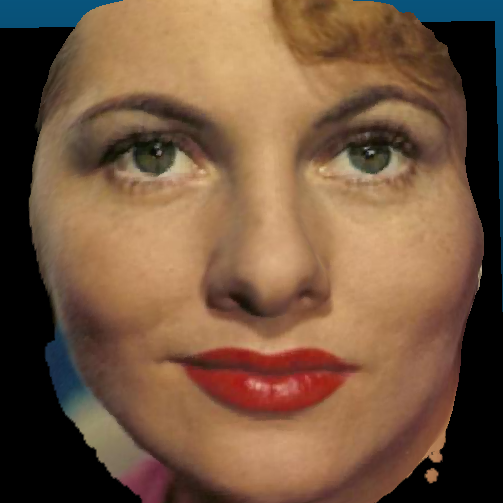}\hspace{.3em}
	\includegraphics[width=0.107\columnwidth]{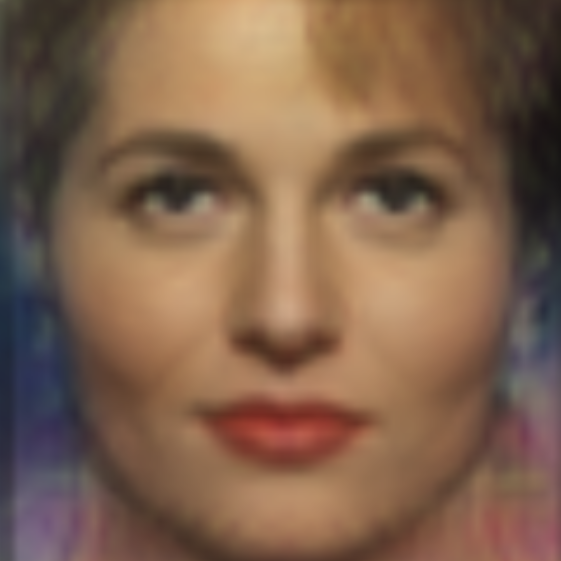}\hfill
	\includegraphics[width=0.107\columnwidth]{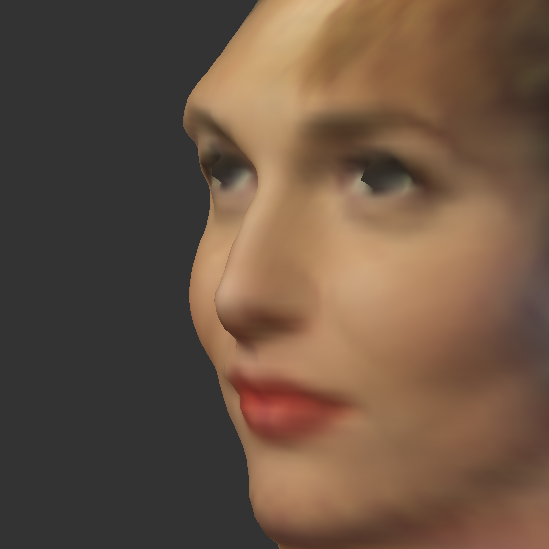}\hfill
	\includegraphics[width=0.107\columnwidth]{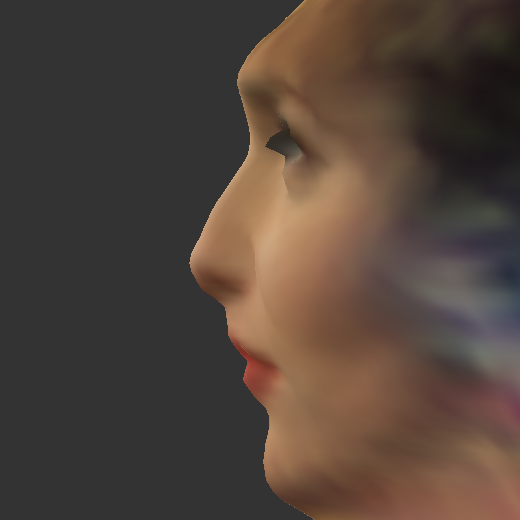}\hfill
	\includegraphics[width=0.107\columnwidth]{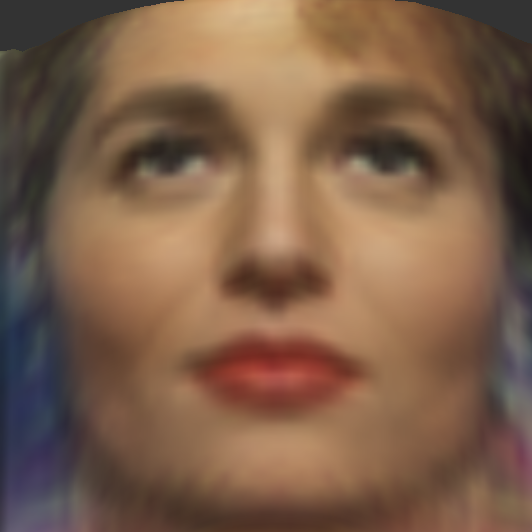}

	\includegraphics[width=0.107\columnwidth]{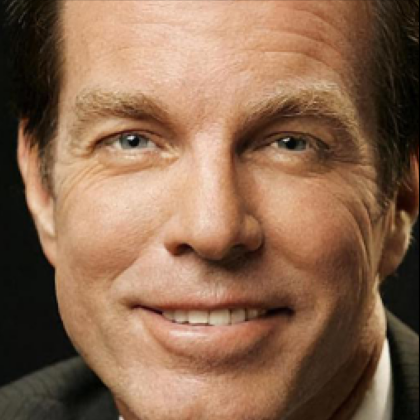}\hspace{.3em}
	\includegraphics[width=0.107\columnwidth]{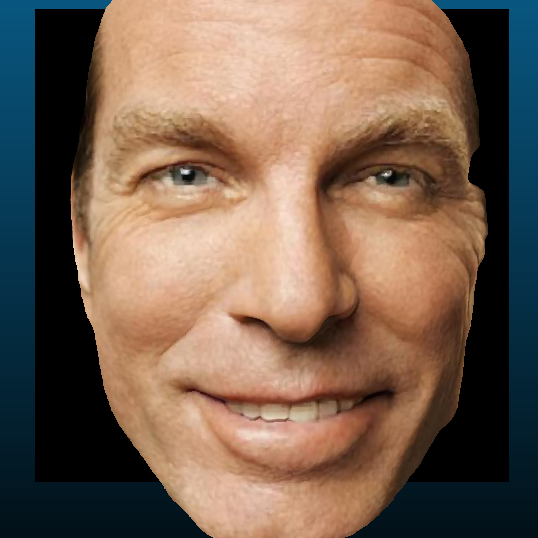}\hfill
	\includegraphics[width=0.107\columnwidth]{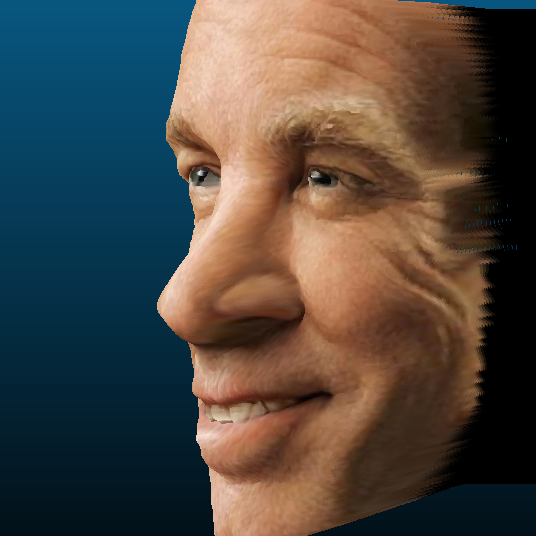}\hfill
	\includegraphics[width=0.107\columnwidth]{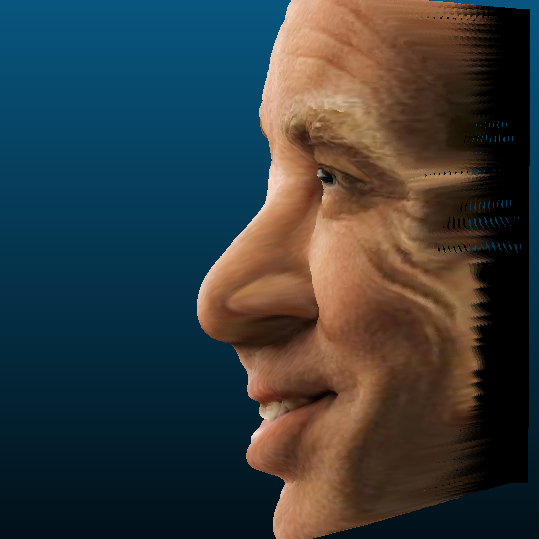}\hfill
	\includegraphics[width=0.107\columnwidth]{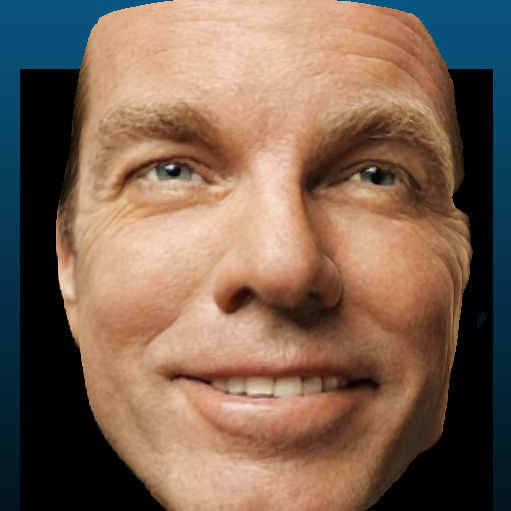}\hspace{.3em}
	\includegraphics[width=0.107\columnwidth]{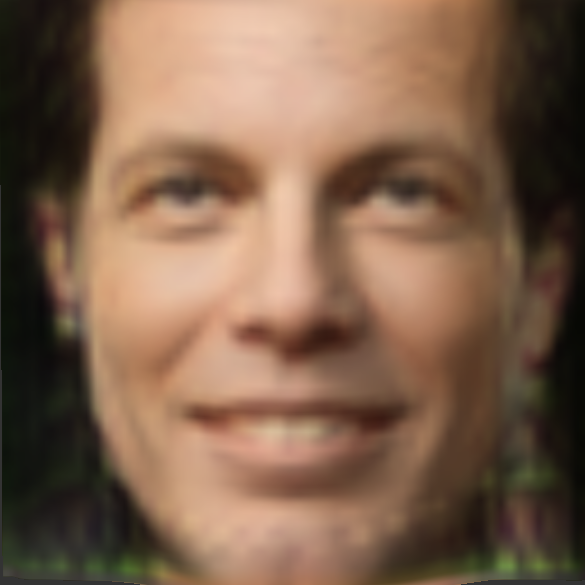}\hfill
	\includegraphics[width=0.107\columnwidth]{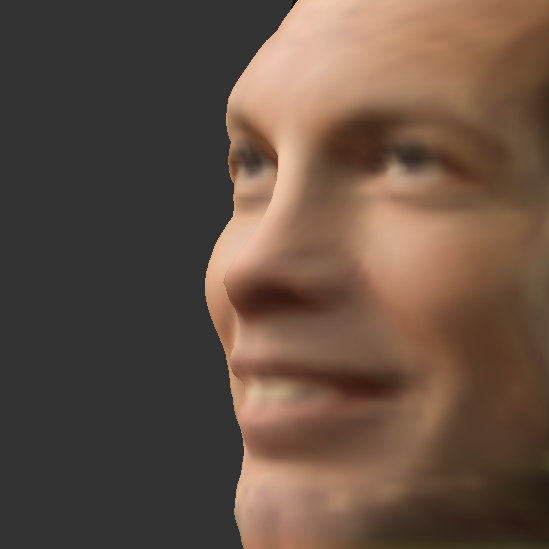}\hfill
	\includegraphics[width=0.107\columnwidth]{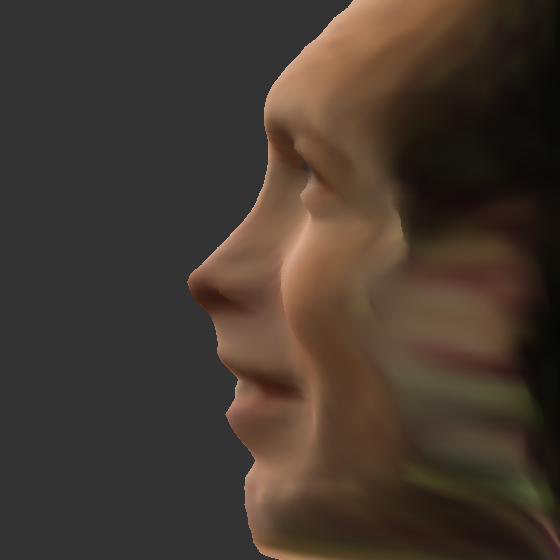}\hfill
	\includegraphics[width=0.107\columnwidth]{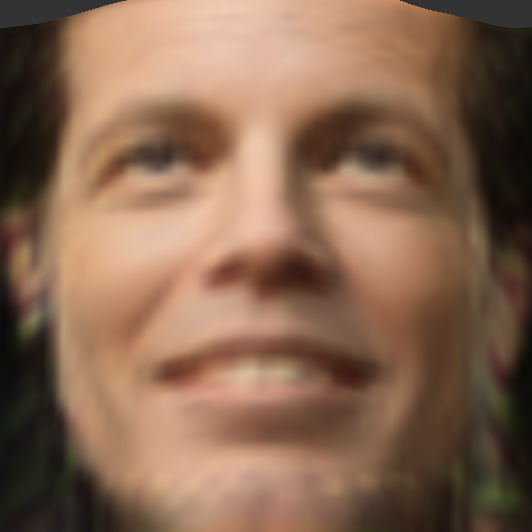}

	\includegraphics[width=0.107\columnwidth]{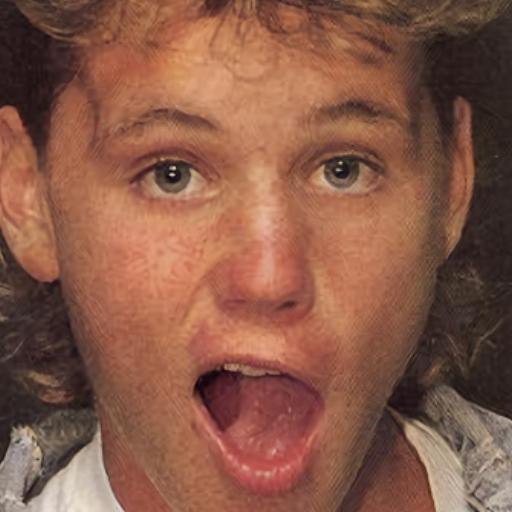}\hspace{.3em}
	\includegraphics[width=0.107\columnwidth]{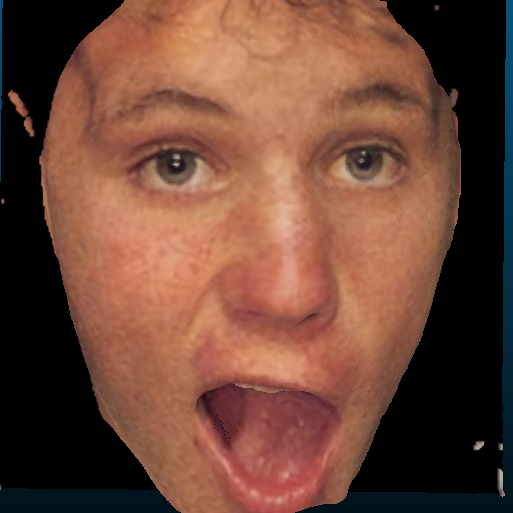}\hfill
	\includegraphics[width=0.107\columnwidth]{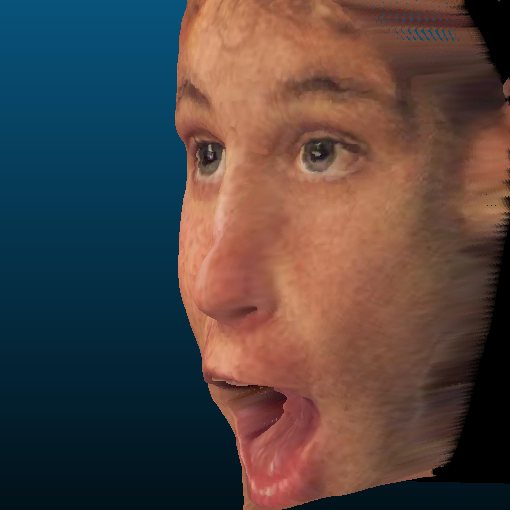}\hfill
	\includegraphics[width=0.107\columnwidth]{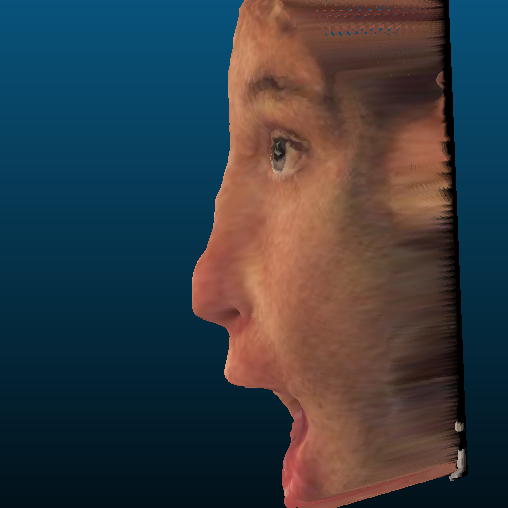}\hfill
	\includegraphics[width=0.107\columnwidth]{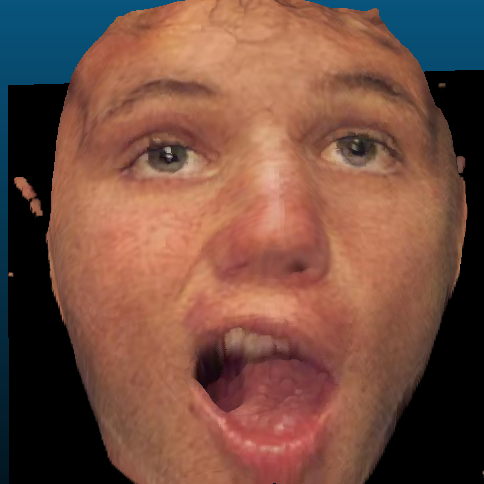}\hspace{.3em}
	\includegraphics[width=0.107\columnwidth]{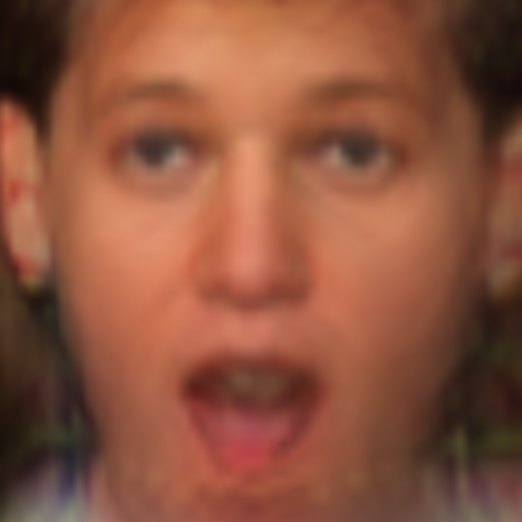}\hfill
	\includegraphics[width=0.107\columnwidth]{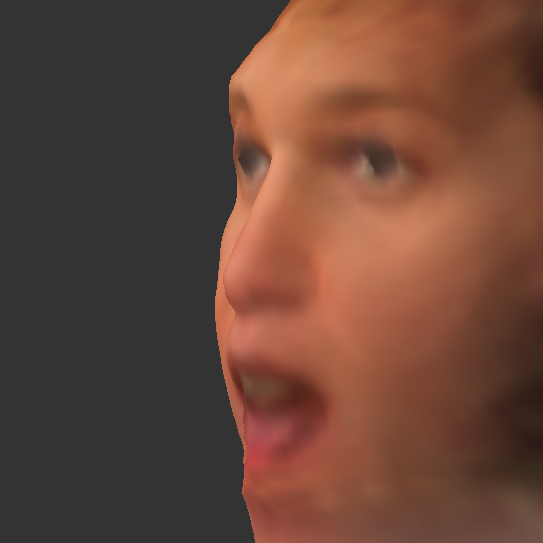}\hfill
	\includegraphics[width=0.107\columnwidth]{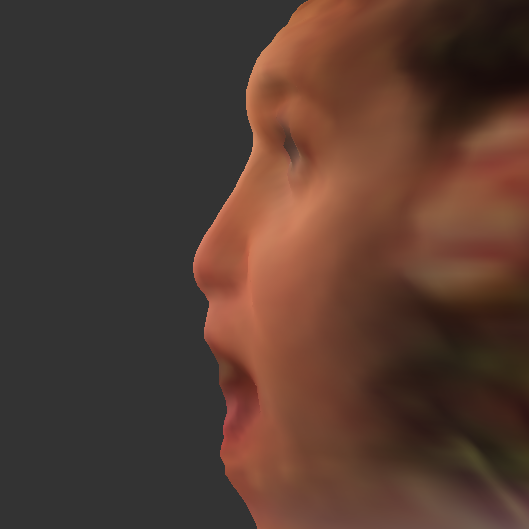}\hfill
	\includegraphics[width=0.107\columnwidth]{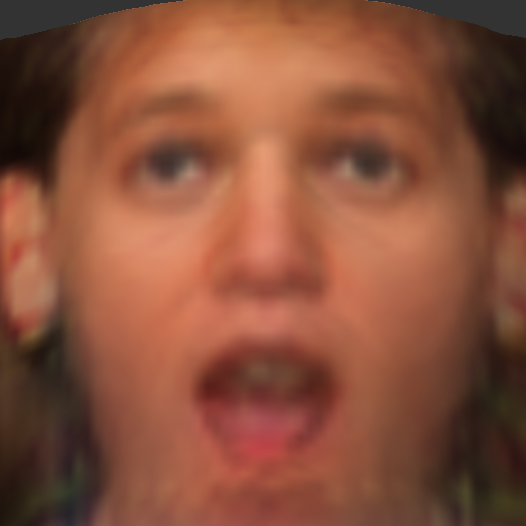}

	\includegraphics[width=0.107\columnwidth]{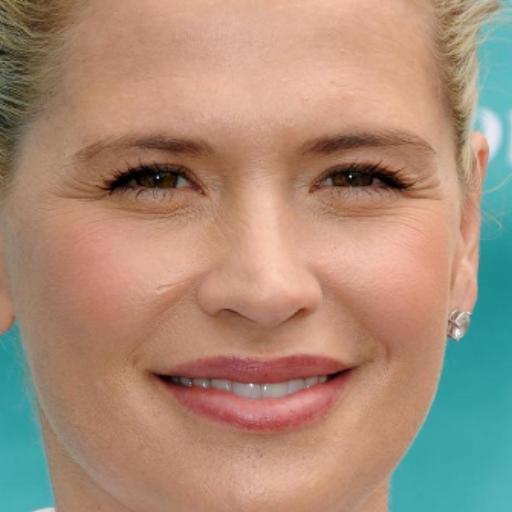}\hspace{.3em}
	\includegraphics[width=0.107\columnwidth]{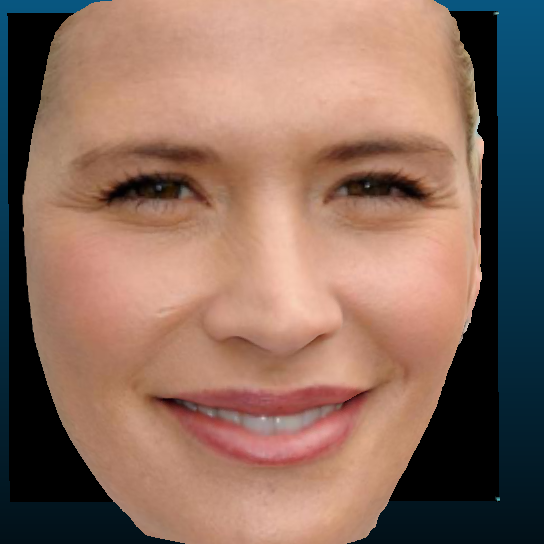}\hfill
	\includegraphics[width=0.107\columnwidth]{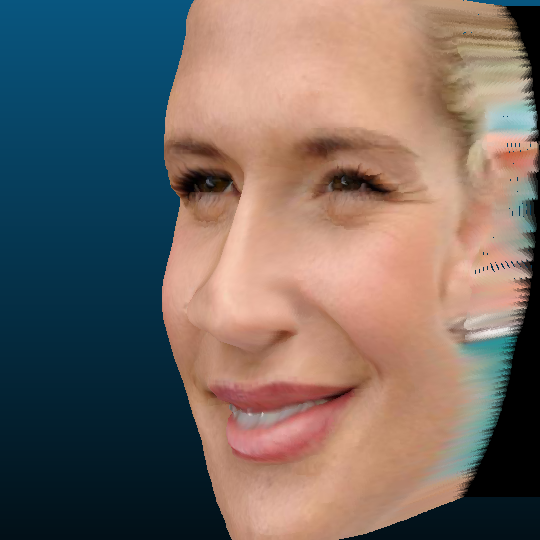}\hfill
	\includegraphics[width=0.107\columnwidth]{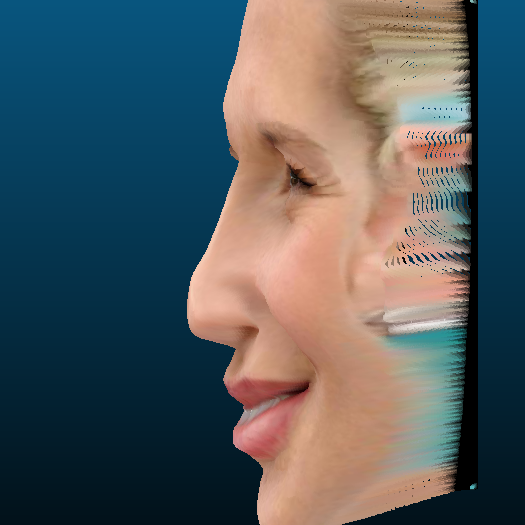}\hfill
	\includegraphics[width=0.107\columnwidth]{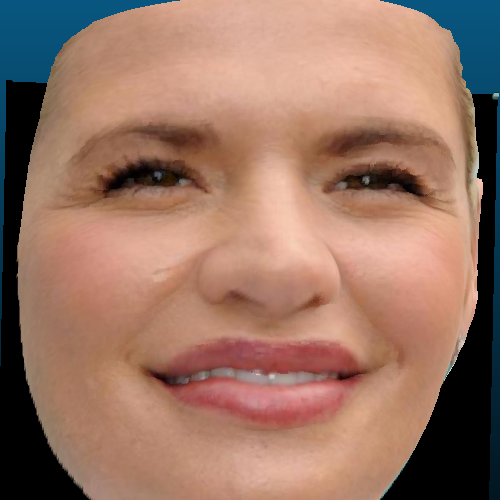}\hspace{.3em}
	\includegraphics[width=0.107\columnwidth]{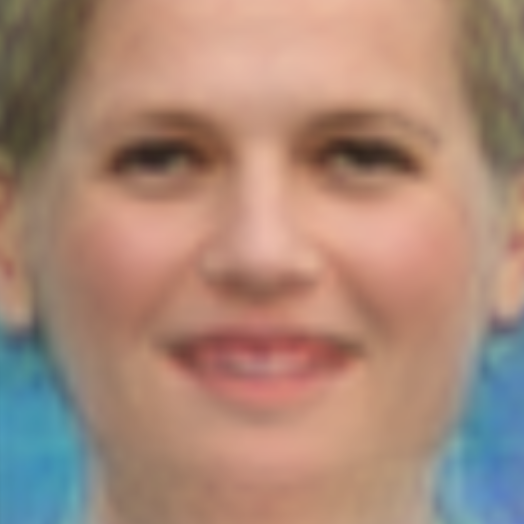}\hfill
	\includegraphics[width=0.107\columnwidth]{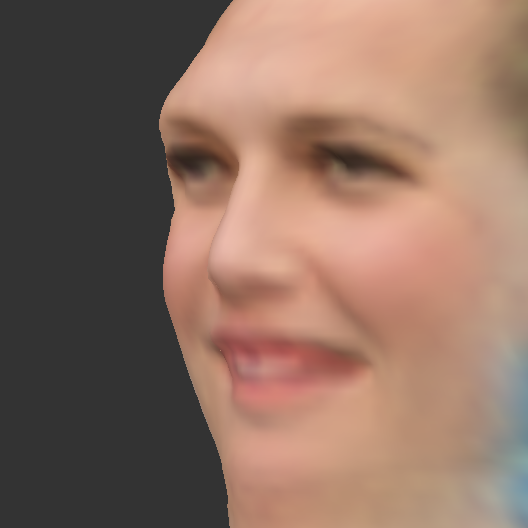}\hfill
	\includegraphics[width=0.107\columnwidth]{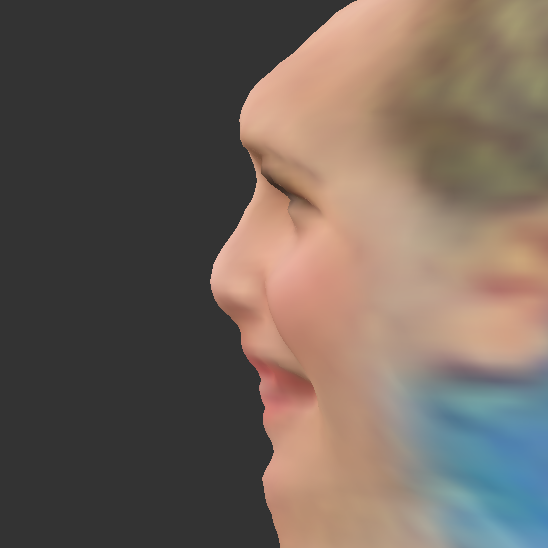}\hfill
	\includegraphics[width=0.107\columnwidth]{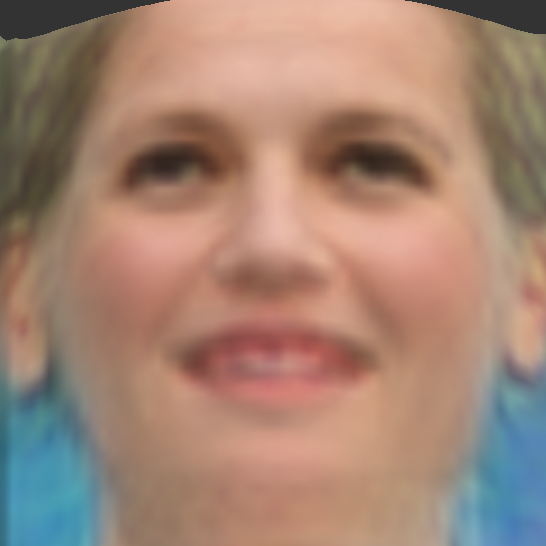}
	
	\includegraphics[width=0.107\columnwidth]{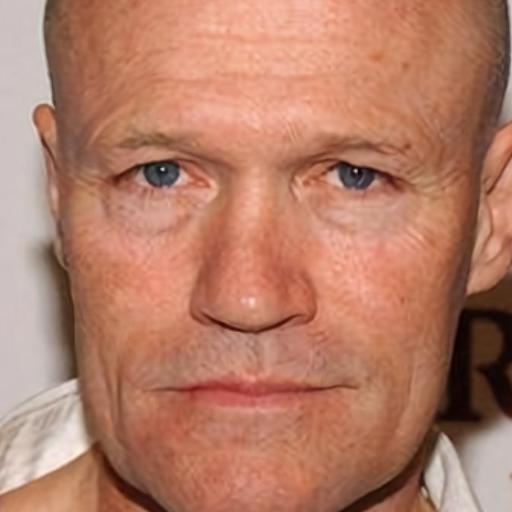}\hspace{.3em}
	\includegraphics[width=0.107\columnwidth]{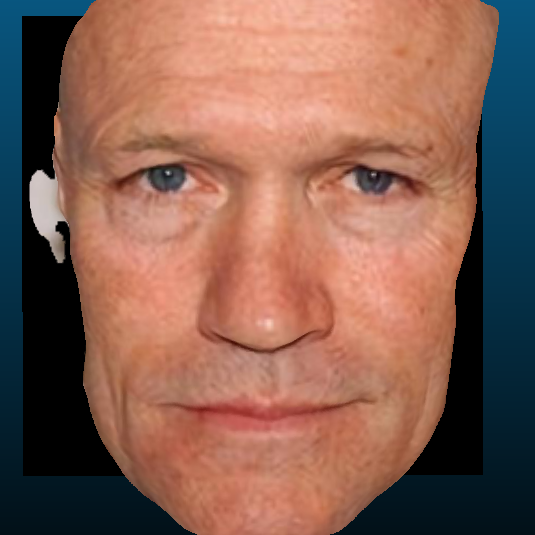}\hfill
	\includegraphics[width=0.107\columnwidth]{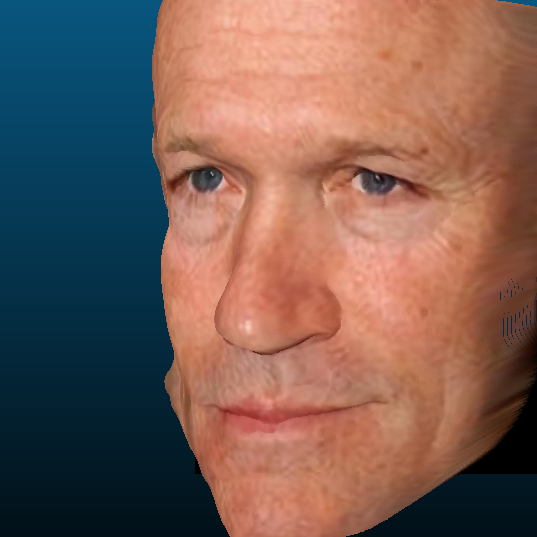}\hfill
	\includegraphics[width=0.107\columnwidth]{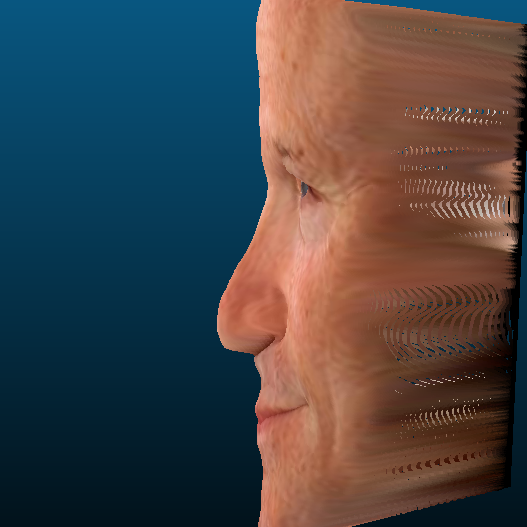}\hfill
	\includegraphics[width=0.107\columnwidth]{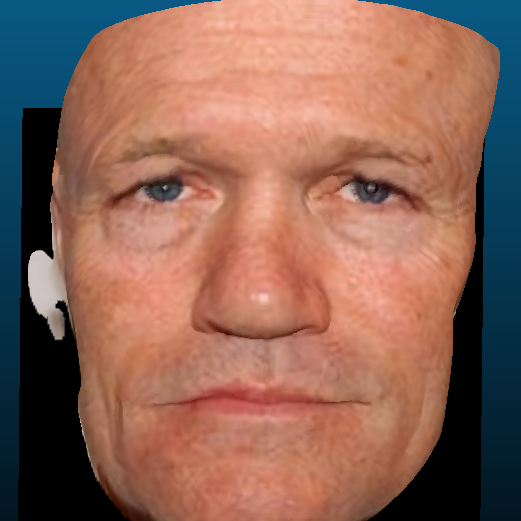}\hspace{.3em}
	\includegraphics[width=0.107\columnwidth]{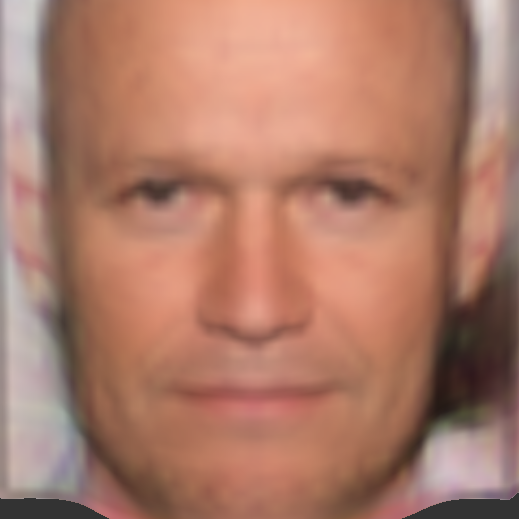}\hfill
	\includegraphics[width=0.107\columnwidth]{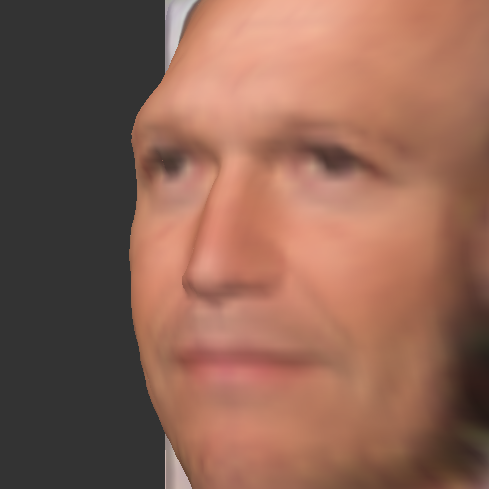}\hfill
	\includegraphics[width=0.107\columnwidth]{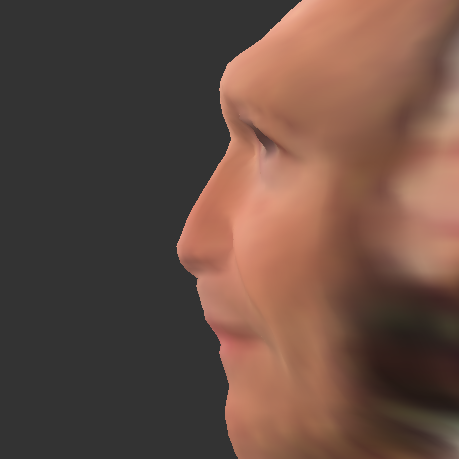}\hfill
	\includegraphics[width=0.107\columnwidth]{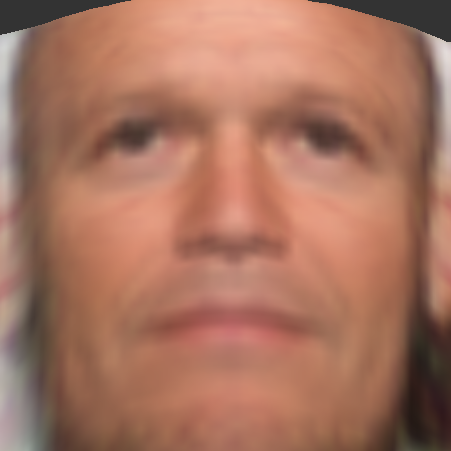}

	\includegraphics[width=0.107\columnwidth]{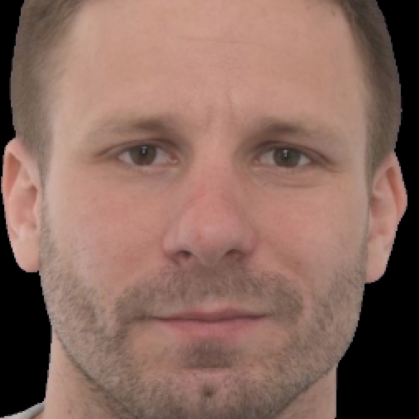}\hspace{.3em}
	\includegraphics[width=0.107\columnwidth]{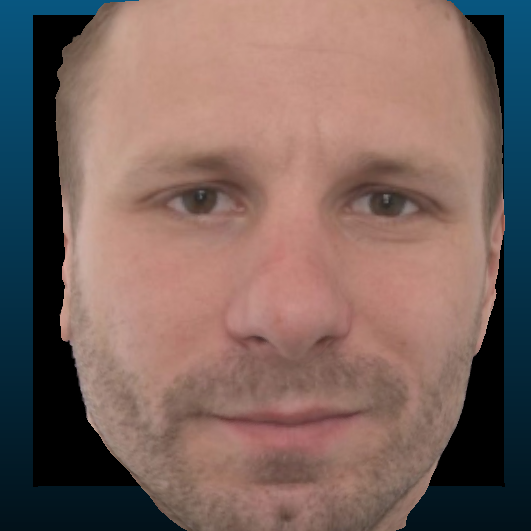}\hfill
	\includegraphics[width=0.107\columnwidth]{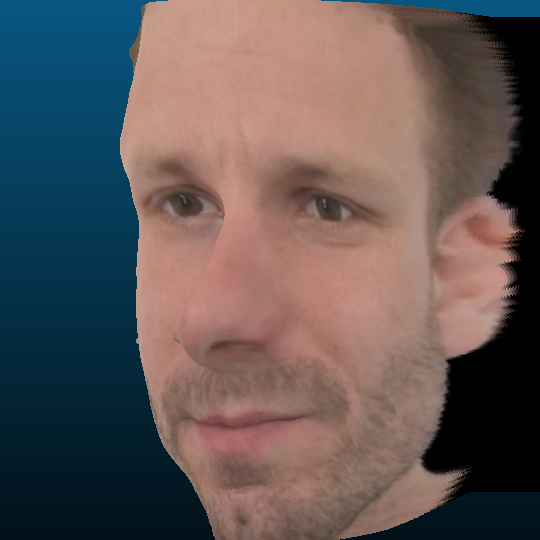}\hfill
	\includegraphics[width=0.107\columnwidth]{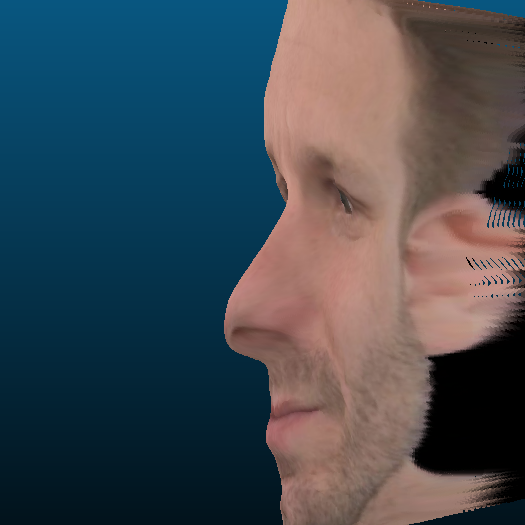}\hfill
	\includegraphics[width=0.107\columnwidth]{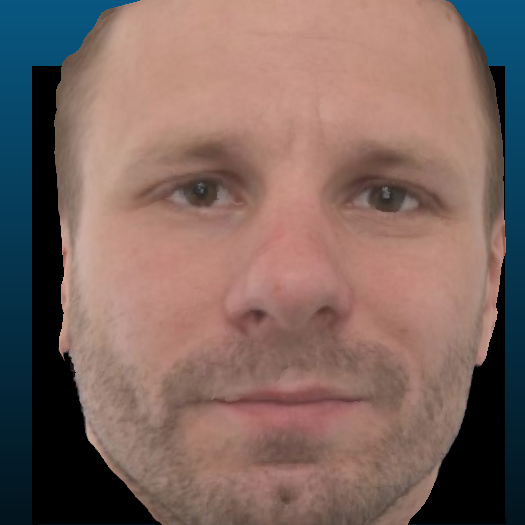}\hspace{.3em}
	\includegraphics[width=0.107\columnwidth]{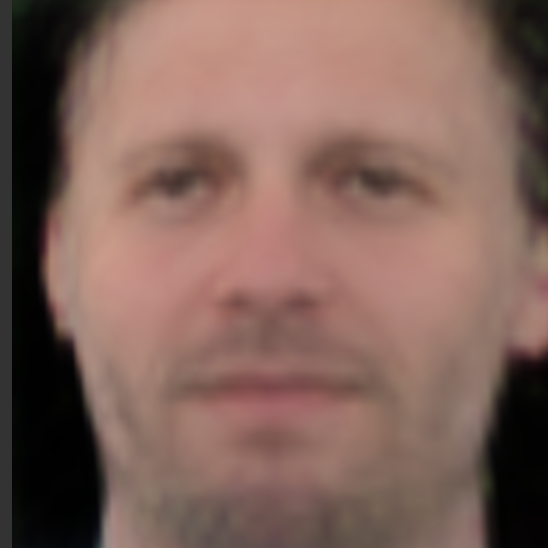}\hfill
	\includegraphics[width=0.107\columnwidth]{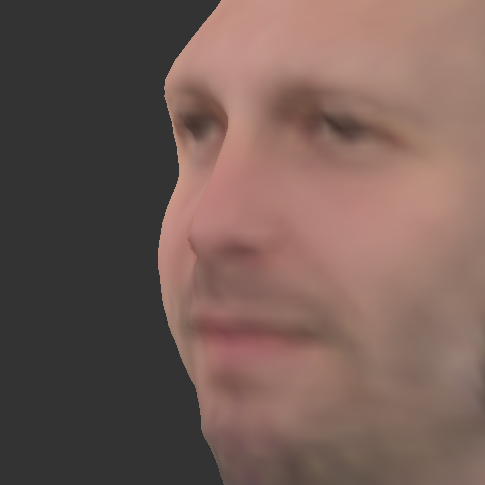}\hfill
	\includegraphics[width=0.107\columnwidth]{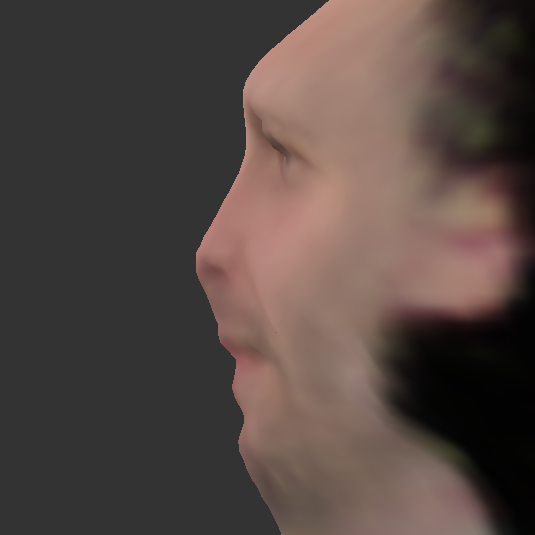}\hfill
	\includegraphics[width=0.107\columnwidth]{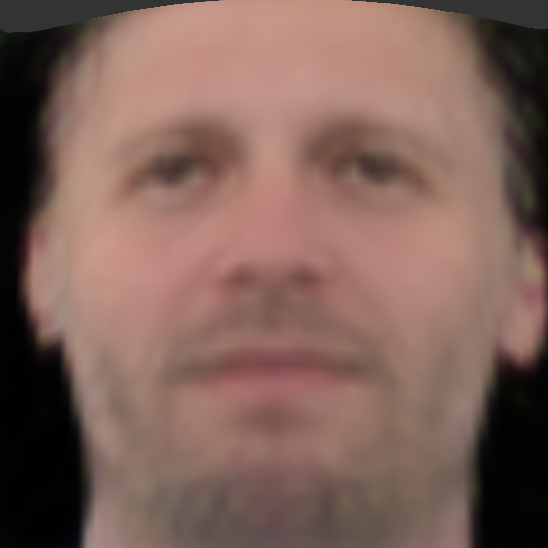}

	\includegraphics[width=0.107\columnwidth]{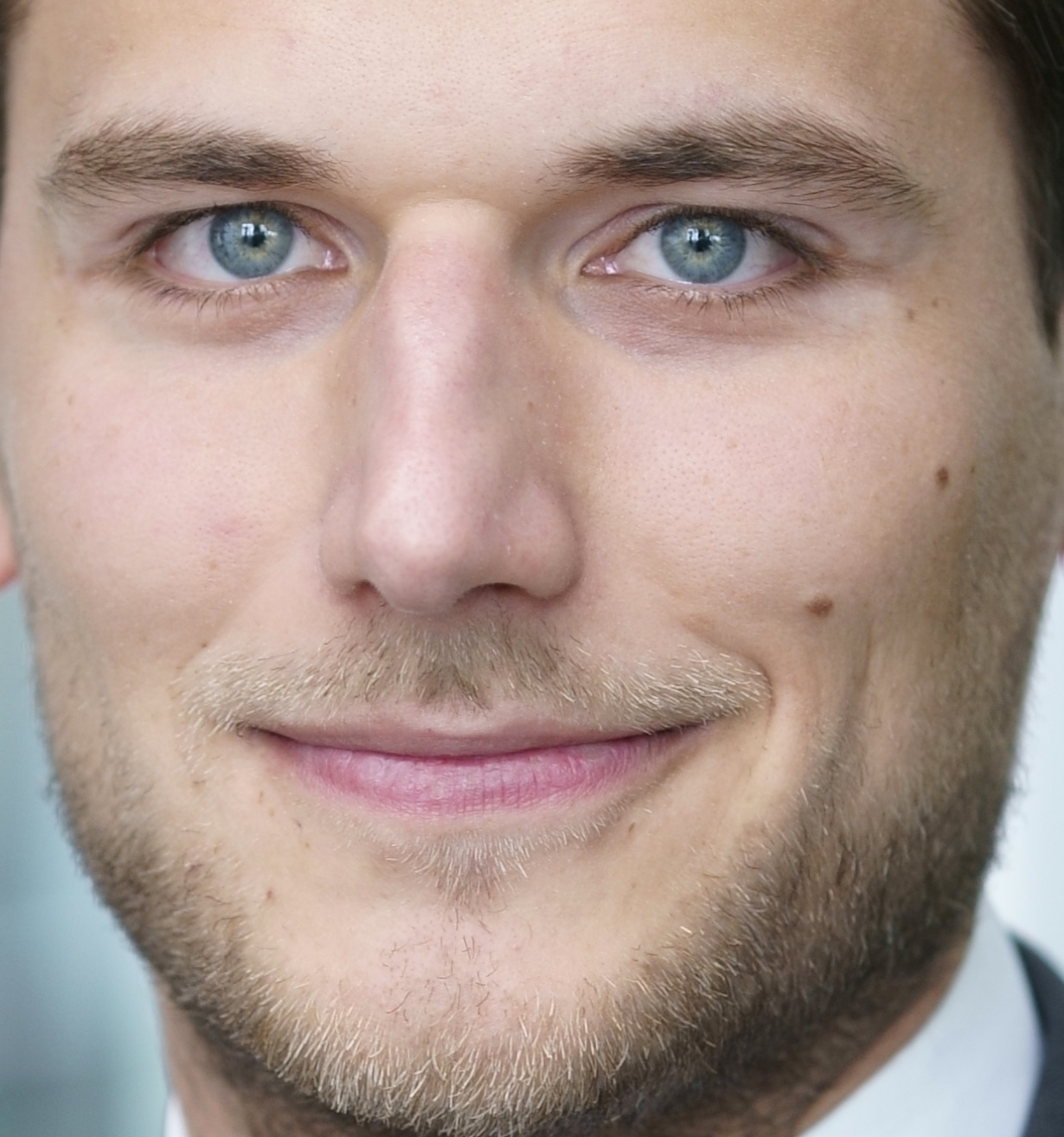}\hspace{.3em}
	\includegraphics[width=0.107\columnwidth]{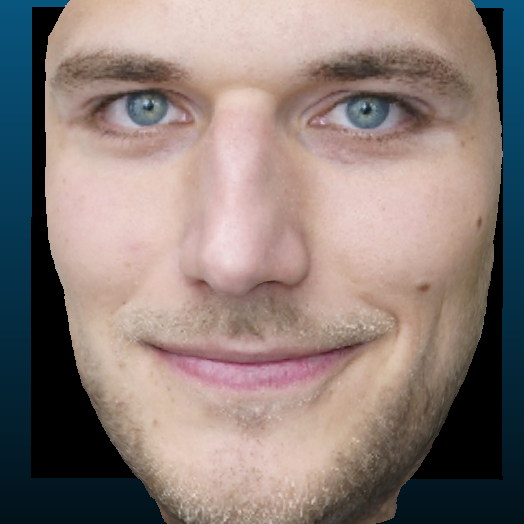}\hfill
	\includegraphics[width=0.107\columnwidth]{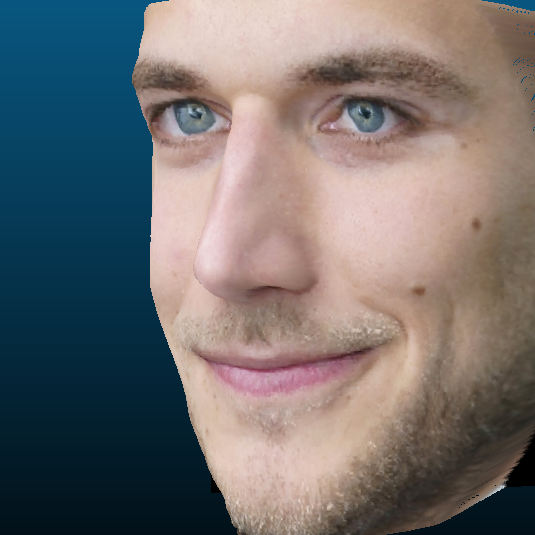}\hfill
	\includegraphics[width=0.107\columnwidth]{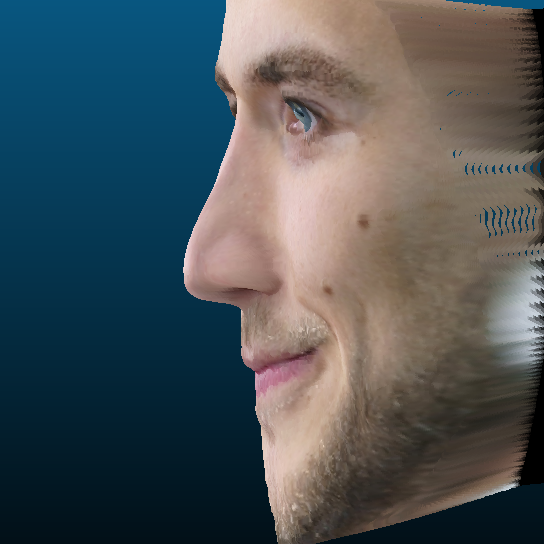}\hfill
	\includegraphics[width=0.107\columnwidth]{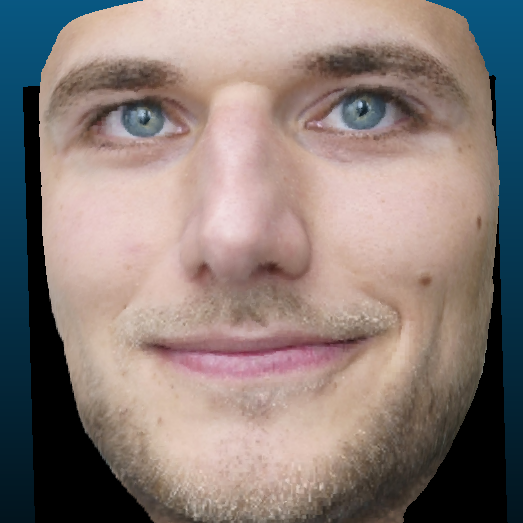}\hspace{.3em}
	\includegraphics[width=0.107\columnwidth]{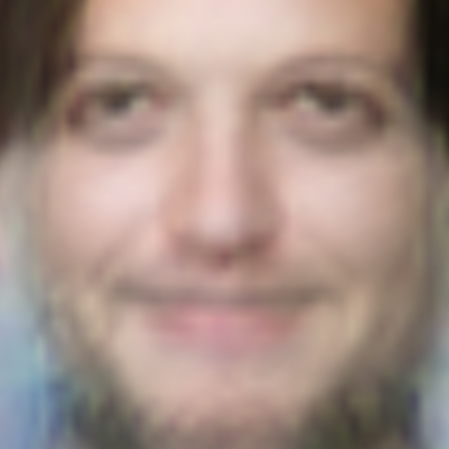}\hfill
	\includegraphics[width=0.107\columnwidth]{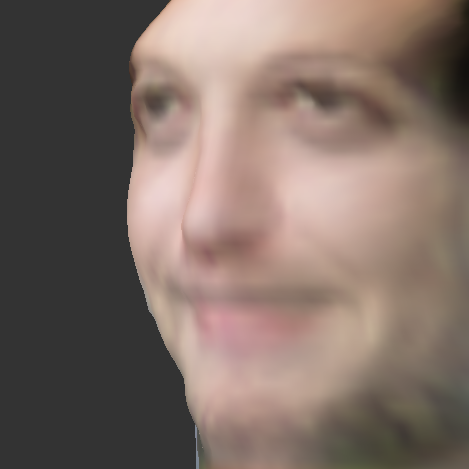}\hfill
	\includegraphics[width=0.107\columnwidth]{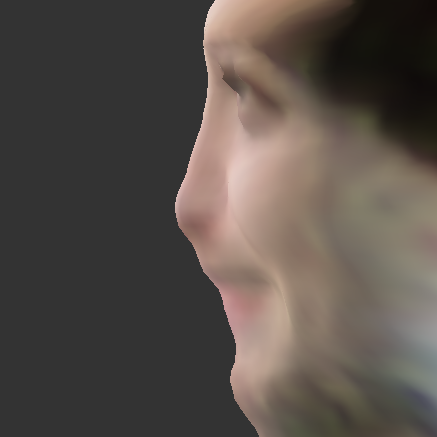}\hfill
	\includegraphics[width=0.107\columnwidth]{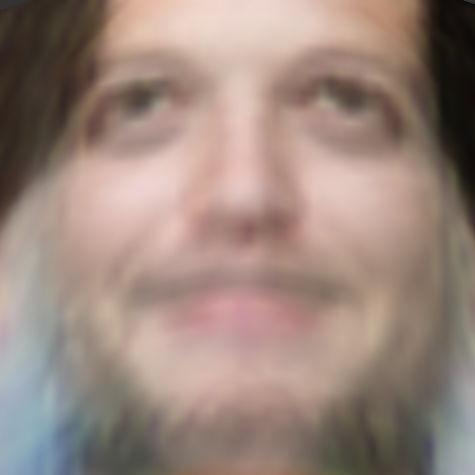}

	\caption{From left to right: RGB input, four snapshots of the synthesized 3D model generated by our method and four snapshots of the synthesized 3D model generated by Wu et al. \cite{wu2020}.}
	\label{fig:ap2}
\end{figure}

We visually evaluate the success of the proposed unsupervised approach and present in Figure \ref{fig:ap2} synthesized 3D models that were created from RGB images of the Bosphorus-3DFA, the CelebAMask-HQ, and images in the wild.

\clearpage
{\small
\bibliographystyle{ieee_fullname}
\bibliography{egbib.bib}
}

\end{document}